\documentclass[10pt,twocolumn]{article}
\usepackage{etex} 
\usepackage[utf8]{inputenc}
\usepackage[T1]{fontenc}
\usepackage{cite}
\usepackage[pdftex]{graphicx}
\usepackage[cmex10]{amsmath}
\usepackage{amssymb}
\usepackage{amstext}
\usepackage{amsthm}
\usepackage{array}
\usepackage{multirow}
\usepackage{multicol}
\usepackage{url}
\usepackage{hyperref}
\usepackage[a4paper, margin=1.5cm]{geometry}
\usepackage{times}
\usepackage[normalem]{ulem} 
\usepackage{xargs} 
\usepackage[pdftex,dvipsnames]{xcolor}
\usepackage{todonotes}
\usepackage{ctable}
\usepackage[]{algorithm2e}

\usepackage{float}

\usepackage{mathtools}
\usepackage{latexsym}
\usepackage{graphicx}
\usepackage{amsfonts}
\usepackage{ctable}

\usepackage{microtype}
\usepackage{graphicx}
\usepackage{subcaption}
\usepackage{soul}

\newcommand{\R}{\mathbb{R}}
\newcommand{\set}[1]{\mathcal{#1}}

\newcommand{\vect}[1]{\boldsymbol{#1}}

\title{A Dataset for Evaluating Blood Detection in Hyperspectral Images}
\author{Michał Romaszewski$^\dag$, Przemysław Głomb, Arkadiusz Sochan, Michał Cholewa}
\date{
	Institute of Theoretical and Applied Informatics, Polish Academy of Sciences\\
	Bałtycka 5, 44-100 Gliwice, Poland\\
	Email: \{mromaszewski, pglomb, asochan, mcholewa\}@iitis.pl\\
	Telephone: +48 32 2317319\\
	\small (\dag\ corresponding author)
}

\begin{document}
\twocolumn[
\begin{@twocolumnfalse}
	\maketitle
	
	\begin{abstract}
	The sensitivity of imaging spectroscopy to haemoglobin derivatives makes it a promising tool for detecting blood. However, due to complexity and high dimensionality of hyperspectral images, the development of hyperspectral blood detection algorithms is challenging. To facilitate their development, we present a new hyperspectral blood detection dataset. This dataset, published in accordance to open access mandate, consist of multiple detection scenarios with varying levels of complexity. It allows to test the performance of Machine Learning methods in relation to different acquisition environments, types of background, age of blood and presence of other blood-like substances. We explored the dataset with blood detection experiments. We used hyperspectral target detection algorithm based on the well-known Matched Filter detector. Our results and their discussion highlight the challenges of blood detection in hyperspectral data and form a reference for further works. 
	\end{abstract}
	
	\vspace*{0.5cm}
\end{@twocolumnfalse}
]

\section{Introduction}

Bloodstains, or remnants of blood present on objects, are among key elements in crime scene evidence gathering and investigation. The age of bloodstains can be deduced from their chemical composition and may provide a clue to the temporal order of events~\cite{zadora2018pursuit}. Digital image processing methods can be used for an objective bloodstain pattern feature extraction~\cite{arthur2017image}, leading to conclusions about positions and movements of persons and objects present in a violent scene~\cite{James2005bpa}. A proper scene analysis procedure, combined with a careful selection of chemical tests, can identify the presence of bloodstains while preserving the DNA information~\cite{Tobe2007sixpresumptive}, allowing for later identification of the participants. Given the volume of information that can be deduced from blood remains, their detection is crucial for an investigation.

Typically blood is identified using chemical tests which change colour~\cite{james2002forensic}, exhibit fluorescence~\cite{budowle2000presumptive} or luminescence~\cite{barni2007forensic} when they come in contact with blood stains. However, since these chemicals must be locally applied in the area where blood stains are present, they can lead to a loss of evidence~\cite{Tobe2007sixpresumptive} or they can interfere with other procedures~\cite{barni2007forensic}. A detection method that is non-contact, applicable to large areas of a scene, and open to automatization is an asset in evidence gathering. Hyperspectral Imaging (HSI), alternatively called Imaging Spectroscopy (IS), is a computer vision approach that can be readily used in this role. Through analysis of the reflectance response across visible and near-infrared (VNIR) electromagnetic spectrum, HSI can be used to infer the presence and properties of blood components, namely the haemoglobin derivatives~\cite{zadora2018pursuit,majda2018hyperspectral}.

The challenges of using HSI for blood detection depend on many factors, such as: the effect of spectral mixing of the bloodstain with the underlying surface~\cite{BioucasDias2012unmixing}, time-related change of bloodstain spectra~\cite{majda2018hyperspectral}, a possible mismatch between the reference and target spectra, differences in the acquisition equipment, complexity of the scene and lighting. In this complex setting, multivariate data analysis in the form of signal processing or Machine Learning methods are needed to extract maximum useful information from the spectra~\cite{Pereira2017nir}. To develop such methods, one must consider both the elements of a scene, that could lead to errors e.g. blood-like substances~\cite{yang2016spectral} and the method behaviour in a simulated or mock-up scene~\cite{edelman2012hyperspectral}, to assess the performance in expected imaging conditions.

To facilitate creation, testing and comparison of blood detection algorithms, in this paper a new dataset for blood detection in hyperspectral images is introduced. It contains a set of hyperspectral images of a mock-up scene. Images in the dataset present sub-scenes of varying complexity. They take into account a number of factors: blood samples of varying sizes, presence of blood-like substances and backgrounds of different composition. Images were captured with two hyperspectral cameras. Selected scenes were repeatedly imaged over several days to capture time-related change in blood spectra. The intention behind this dataset was to portrait the diversity and varying levels of difficulty present in blood detection scenarios. To facilitate the use of the dataset, it was made available on-line under an open licence. Along with the data, the experimental source code and ground truth maps required to replicate results presented in this manuscript were included. To the best authors' knowledge, there is currently no publicly available dataset of a similar nature. To highlight the challenges of this dataset, a reference target detection experiments were performed and their results are included in this paper. The proposed detection algorithm was based on a well known statistical detector -- the Matched Filter (MF) -- following the reasoning presented in~\cite{manolakis2009there} that a reasonable approach is to use detector algorithms with good performance and well-understood theoretical properties. 
 
This paper is organised as follows: Sec.~\ref{sec:scenarios_and_data} describes the new hyperspectral dataset. Sec.~\ref{sec:methods} introduces the problem of hyperspectral target detection, measures of performance used in our experiments, presents an example of target detection and introduces the proposed detection algorithm. Results are presented in Sec.~\ref{sec:experiments_and_results} and discussed in Sec.~\ref{sec:discussion}. Finally, conclusions are presented in Sec.~\ref{sec:conclusion}.
 
\subsection{Related work}
\label{sec:relatedwork}

HSI is an imaging technique that combines properties of a digital camera, which captures a 2D scene image, with a spectroscope, registering incident light as a large number of consecutive, narrowband wavelength responses. This combination of spatial and spectral information provides a rich description of a scene, which can be used to identify substances present or observe their properties, e.g. a time induced degradation. HSI has been extensively used e.g. for crop monitoring and yield modelling in precision farming~\cite{thenkabail2016hyperspectral}, non-contact food quality inspection~\cite{pu2015recent} or pigment identification in an art conservation setting~\cite{grabowski2018automatic}. Examples related to the forensic science include fingerprint and fingermarks trace contamination analysis~\cite{edelman2012hyperspectral}, detection of residues of explosives~\cite{Fernandez2014explosive} or  gunshots~\cite{glomb2018application} and identification of soil alteration for grave localization~\cite{Leblanc2014graves} or forged fragments for document analysis~\cite{Silva2014forgery}.

The spectral characteristics of blood components, notably haemoglobin, make HSI well-suited for bloodstains identification and analysis. Cellular elements, including red and white cells and platelets, form about 45\% of the blood; red blood cells are composed primarily of haemoglobin~\cite{James2005bpa}. In a bloodstream of a healthy individual two species of the latter are present: deoxyhaemoglobin (deoxyHb) and oxyhaemoglobin (oxyHb); without the supporting biological processes, oxyHb oxidises to methaemoglobin (metHb) which then denaturates to hemichrome (HC)~\cite{zadora2018pursuit}. The oxyHb has a number of characteristic spectral features, including dips in reflectance spectra at $\sim414$ nm (Soret band) and $\sim542$ nm and $\sim576$ nm ($\alpha$ and $\beta$ bands); time-induced degradation to metHb is visible in the spectra in particular at those wavelengths~\cite{majda2018hyperspectral}. These data features are mostly consistent for both human and animal blood \cite{zijlstra1997spectrophotometry} and can be applied e.g. for detection of blood and estimation of haemoglobin concentration in fish muscle~\cite{skjelvareid2017detection}. Time induced degradation of oxyHb is not uniform, but has biphasic nature~\cite{Bremmer2011biphasic}; knowledge of environmental factors can increase precision of blood sample age determination. The visibility of oxyHb and metHb in the spectra makes HSI useful in a range of applications related to medical diagnosis~\cite{lu2014medical}. It can be also used to identify and estimate the age of blood~\cite{edelman2012identification}. 

A number of studies have been published, evaluating various approaches for detection, classification and property estimation of the bloodstains. In~\cite{Pereira2017nir} author perform an evaluation of bloodstains detector based on NIR spectrometry data and several classification methods; with a non-porous background (glass and metal) and carefully controlled acquisition conditions many methods achieve or approach close to $100\%$ accuracy. An in depth-study of a number of regression and Machine Learning methods is presented in~\cite{sharma2018trends}; after a detailed examination, more advanced methods (e.g. a SVM) are found to perform better, while stressing the need for more research related to blood age effects. An important research problem lies in distinguishing between blood and other substances, as in the problem studied in~\cite{yang2016spectral}, where authors compare spectral characteristics of blood and nine types of non-blood substances such as tomato sauce or red wine; their conclusion is that blood is significantly different from them and detection is viable. Another interesting application lies in finding and identifying blood stains in the field, which corresponds to detection in an unknown and diverse acquisition environment. For example in \cite{edelman2012hyperspectral} authors created a simulated crime scene in order to estimate the age of blood stains, this research was continued and extended in \cite{edelman2014spectral}. In \cite{yang2016comparison} a simulated scene was used to perform a preliminary verification of two anomaly detection algorithms. Some works such as~\cite{li2014application} or~\cite{zhao2019application} tried to combine these approaches with experiments involving detection of blood in presence of other substances on fabric backgrounds of different colour. Other related works include e.g. detection limit on selected fabric types across variety of infrared ranges~\cite{DeJong2015fabric} or age determination from blood-stained fingerprints~\cite{Cadd2018fingerprints}.

\section{Blood detection dataset}
\label{sec:scenarios_and_data}

Considering a situation where HSI could be used for blood detection at the accident or crime scene, we propose the following assumptions:
\begin{enumerate}
	\item The final goal of the experiment is to detect blood stains in the image, by assigning a positive label to pixels where blood is visible and negative label to other pixels.
	\item We consider the application of supervised algorithms that use a training set of representative blood patterns. In particular, they could come from a different image or an external source e.g. a spectral library.
	\item The detector shouldn't require a background training data, assuming that the only information supplied will be the target spectrum. Additional required parameters (e.g. the covariance matrix) may be estimated from the image.
	\item Blood traces in the image are typically visible as groups of pixels. However, these target pixels do not have to be pure i.e. target spectrum may be mixed with background spectra.
\end{enumerate}

In accordance with these assumptions, we've formulated a number of target detection scenarios. A scenario is associated with a fragment of the prepared mock-up scene containing background materials of different colour, shape and composition.  Blood traces of varying size and shape have been placed in the scene along with a set of visually similar substances such as artificial blood, tomato juice or red paints. As a result, the expected difficulty of detection task varies between scenarios: from a relatively simple environment where the majority of targets is located on a uniform, white background to complex environments with varied backgrounds and a number blood-like substances present. Available images represent multiple  acquisitions over a period of three weeks, with a particular focus on the first two days of acquisition which allows to observe the impact of time-inducted changes in the scene on detection performance. 

\subsection{Description of the dataset}
The dataset presented in this paper consists of hyperspectral images of six fragments of the scene. Individual scenes are presented in the Fig.~\ref{fig:scene:whole} and assigned the letters \emph{A-F}. Their description and motivation are as follows:

\begin{figure}
	\centering
	\begin{subfigure}[b]{1\linewidth}
		\includegraphics[width=1.0\linewidth]{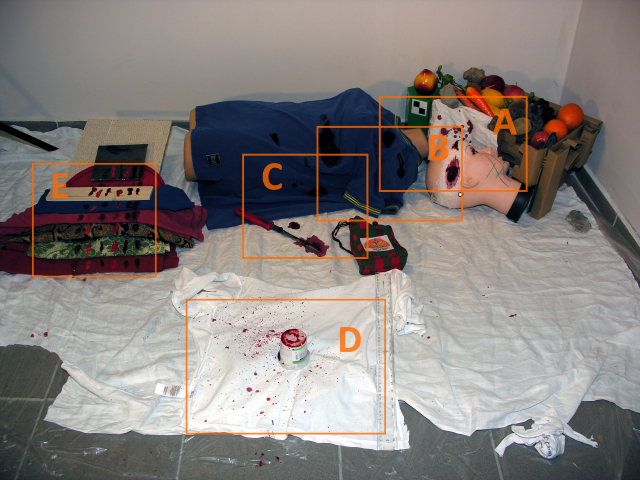}
		\caption{The mock-up of a forensic scene with locations of images \emph{A-E}.}
		\label{fig:scene:whole}
	\end{subfigure}
	\begin{subfigure}[b]{0.42\linewidth}
		\includegraphics[width=1\linewidth]{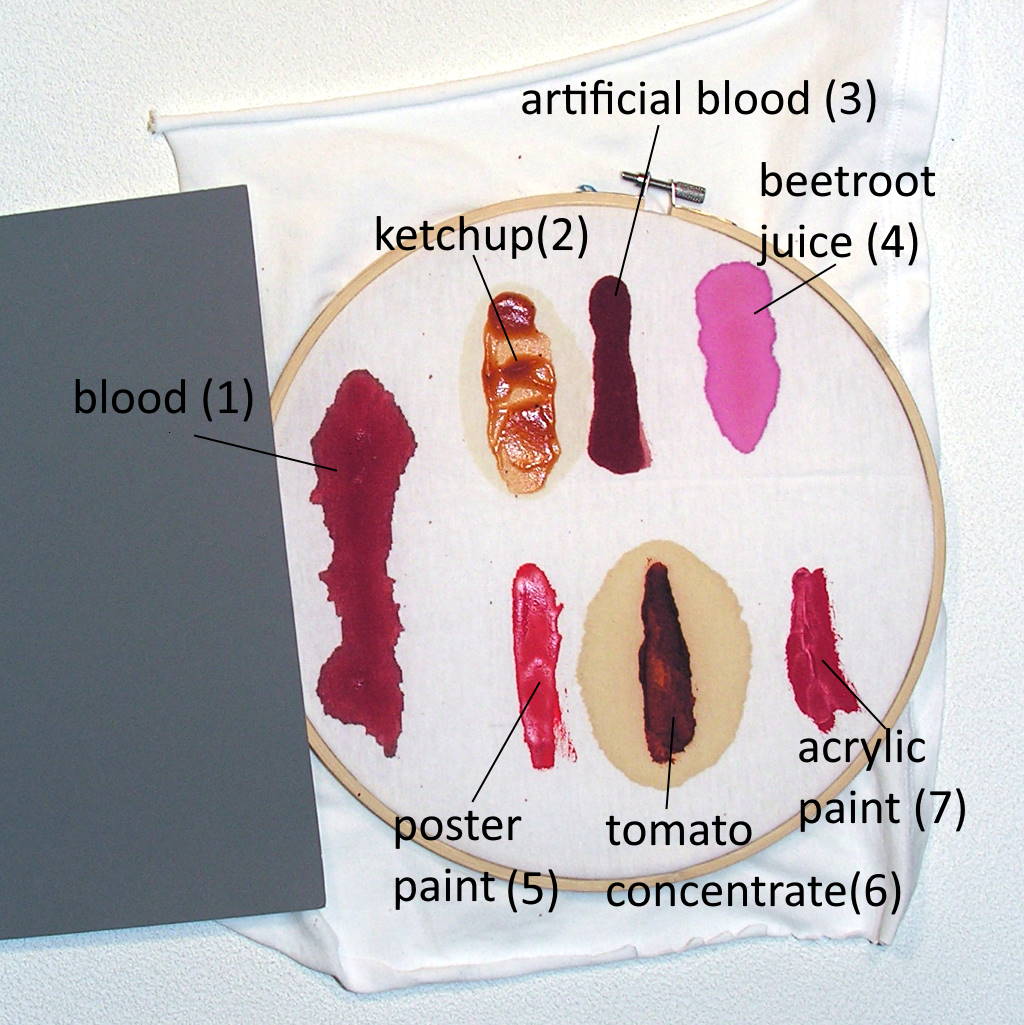}
		\caption{Overview of the \emph{F `Frame'} scene.}
		\label{fig:scene:fannot}
	\end{subfigure}
	\begin{subfigure}[b]{0.56\linewidth}
		\includegraphics[width=1\linewidth]{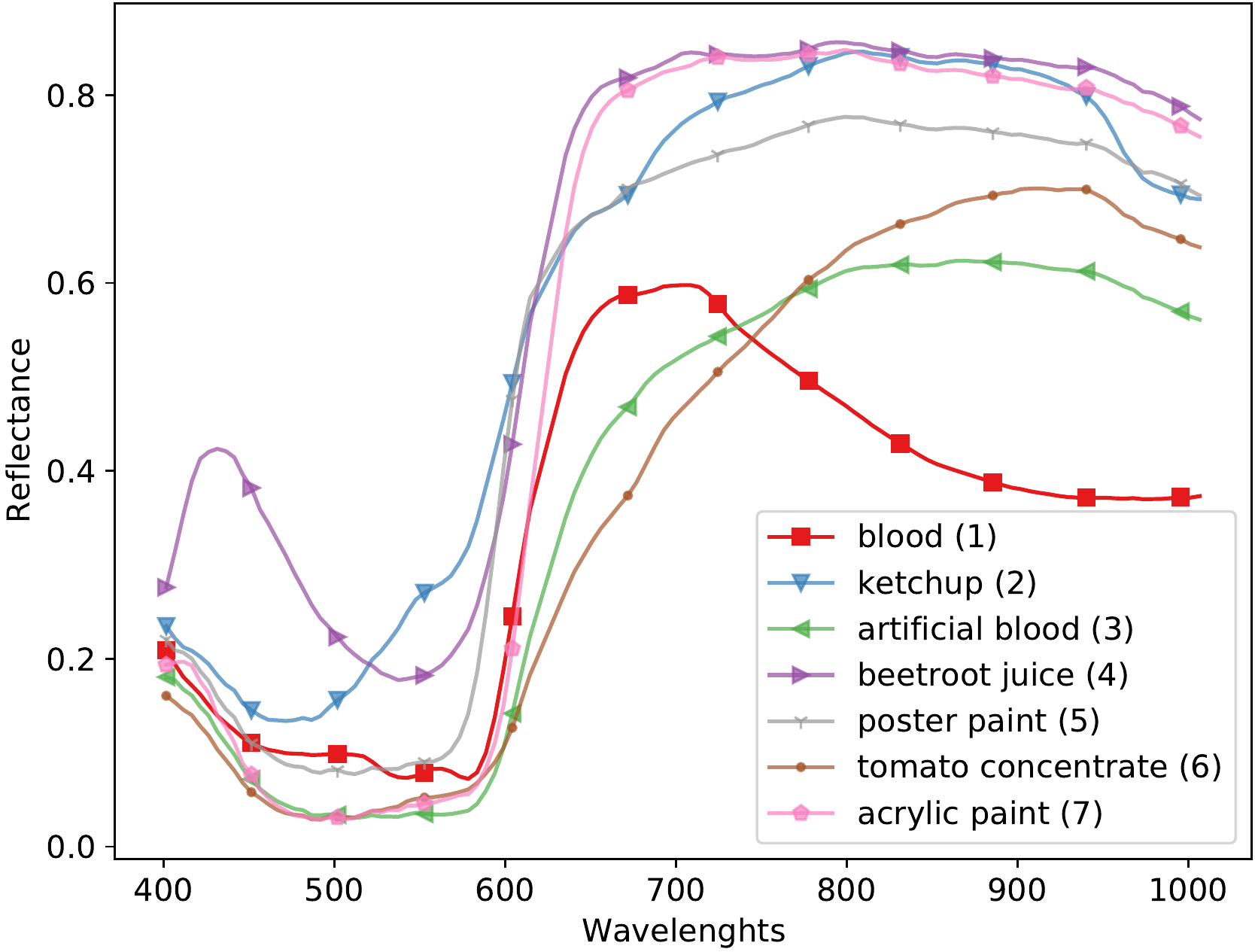}
		\caption{Spectra of blood and blood-like substances from one of the \emph{F} images.}
		\label{fig:scene:spectra}
	\end{subfigure}
	\caption{Illustration of the dataset.} 
	\label{fig:scene}
\end{figure}

\subsubsection{Scene \emph{A `Head'}}
Traces of real and artificial blood are located on white materials -- bandage, a fragment of a white sheet -- and the fragment of the blue shirt on the left side. The area around the blood traces is fairly uniform, but in the far background of the image there is a mixture of plastics and organic substances -- fruits and vegetables -- of different colour.

The scene is intended as a relatively easy detection scenario, with the majority of targets located on a white background. The main challenge lies in the presence of two different backgrounds and an artificial blood. 

\subsubsection{Scene \emph{B `Shirt'}}
On this scene, traces of real and artificial blood are located on the blue material of the thick cotton shirt put on a mannequin. In addition, part of the \emph{`Head'} scene is also visible. This scenario is more complex that scenario \emph{A}, allowing to examine the effectiveness of blood detection on a dark background in presence of cloth fabric bends and resulting shadows.

\subsubsection{Scene \emph{C `Knife'}}
On this scene, traces of real and artificial blood are present on the blade and handle of the knife and in its vicinity. 
The scene allows to test detection in presence of substances of significantly different type and colour i.e. metal blade, plastic handle, mixture of fabrics.

\subsubsection{Scene \emph{D `Splash'}}
This scene contains a blood splash pattern on a white cotton shirt. The splash was prepared by placing the blood sample on a plastic container located in the centre of the scene, and hitting it with a hard object. The traces form a circular pattern while its geometrical centre stays mostly clean. 

Apart from detection problem, the method of application allows to experiment with spatial properties of the blood splashes e.g. determining the location of the geometric centre of the splash or estimating bloodstain pattern features.

\subsubsection{Scene \emph{E `Comparison'}}
This scene contains blood and five blood-like substances on eight different backgrounds, which are a mixture of fabrics, wood, plastic and metal, some of them of red hue. They were arranged in uniform vertical stripes while blood along with five blood-like substances were placed on them in horizontal stripes. 

The purpose of the scenario is to test the detector in a challenging and diverse environment with many materials optically similar to detected targets.

\subsubsection{Scene \emph{F `Frame'}}
This scene contains blood and six blood-like substances on a piece of a white fabric. The fabric was stretched on a wooden frame to minimise the occurrence of creases. This scene is intended as a source of blood spectra for experiments and was acquired multiple times in the space of a month and also with different hyperspectral cameras.

\subsection{Preparation of the scene}
First, the the mock-up scene, containing the blood-like substances was prepared. Then, three blood samples obtained from authors in a sample collection facility were applied. Each sample had a volume of 15~ml; in total 45~ml were used. Blood samples did not contain anticoagulant, and were applied to the scene in about 10 minutes after being collected. The approximate age of samples in images in presented in the Tab.~\ref{tab:target_age}.

Acquisitions were carried out over a period of multiple weeks. With one exception,  all acquisitions were performed in the same room without windows, with stable humidity and temperature of about $15$ degrees Celsius. 
Elements of scenes \emph{A}-\emph{E} were not moved in this time, although the camera and lights were adjusted in course of the acquisitions. 

The frame denoted as scene \textit{F} was moved outside of the laboratory in the second day to be used for reference acquisitions with different hyperspectral camera (a SPECIM hyperspectral system) and was returned later. This reference acquisition is denoted as $F(2k)$ in Tab.~\ref{tab:target_age}. 
Since that image had different, greater wavelength resolution ($d=1000+$ vs $d=128$ bands) it was downsampled to match other images in the dataset. Linear interpolation was used in the downsample process to achieve accurate wavelengths values. In addition, in the first day the frame was placed in the mock-up scene in order to acquire its image in similar lighting conditions to other images in the dataset. This image is denoted as $F(1s)$. 

\ctable[
cap     = Approximate age of blood spectra in the images,
caption = Names and day of acquisition for images in the paper. Codes are unique tags used in the paper. Target source denotes another image in the dataset that was used as a source of blood spectrum in reference experiment. Library index and library age denote the index and age of the blood spectrum from spectral library described in~\cite{majda2018hyperspectral}. Image age is the approximate age of target blood in the image.,
label   = tab:target_age,
star
]{lllllll}{	
}{     \FL
	Image name&Day&Code&Target source&Library index&Library age&Image age\ML
	Frame&1&\textit{F(1)}&\textit{F(1a)}&24&1h15&1h20m\NN
	Frame&1 (in scene)&\textit{F(1s)}&\textit{F(1a)}&21&3h45m&3h50m\NN
	Frame&1 (later)&\textit{F(1a)}&\textit{F(1)}&20&4h50m&5h50m\NN
	Frame&2&\textit{F(2)}&\textit{F(1a)}&15&4d&2d\NN
	Frame&2 (different camera)&\textit{F(2k)}&\textit{F(2)}&15&4d&2d\NN
	Frame&7&\textit{F(7)}&\textit{F(2)}&12&7d&7d\NN
	Frame&21&\textit{F(21)}&\textit{F(7)}&5&22d&21d\NN
	Splash&1&\textit{D(1)}&\textit{F(1)}&22&2h45m&3h\NN
	Head&1&\textit{A(1)}&\textit{F(1)}&21&3h45m&3h30m\NN
	Shirt&1&\textit{B(1)}&\textit{F(1s)}&20&4h50m&4h20m\NN
	Knife&1&\textit{C(1)}&\textit{F(1)}&21&3h45m&4h\NN
	Comparison&1&\textit{E(1)}&\textit{F(1a)}&20&4h50m&4h40m\NN
	Comparison&7&\textit{E(7)}&\textit{F(7)}&12&7d&7d\NN
	Comparison&21&\textit{E(21)}&\textit{F(21)}&5&22d&21d\NN
}

\subsection{Hyperspectral acquisition}
Hyperspectral data acquisition was performed with Surface Optics SOC710 camera. This camera records spectra at VNIR range $377-1046$~nm; the output image has dimensions $696 \times 520$ with $128$ bands and $12$ bit dynamic range. The camera is equipped with sensor line translation unit and can be used from static stand as a conventional camera i.e. it does not require a mechanical translation of the observed sample or rotary stand, as in traditional `push broom' hyperspectral cameras. 

The lighting was provided with four halogen lamps and adjusted for each scenario separately, so that most of the dynamic range of the camera was used and image saturation was avoided. Captured hyperspectral images were subject to a standard calibration procedure, including: the removal of a dark frame, spectral and radiometric calibration as well as reflectance normalization using the Munsel Color grey calibration panel. \emph{B} and \emph{C} images do not contain calibration panel, therefore reflectance calibration was performed based on a separate acquisition and verified by comparing spectra with other images.

\subsubsection{Reflectance correction}
\begin{figure}
	\centering
	\begin{subfigure}[b]{0.49\linewidth}
		\includegraphics[width=1.0\linewidth]{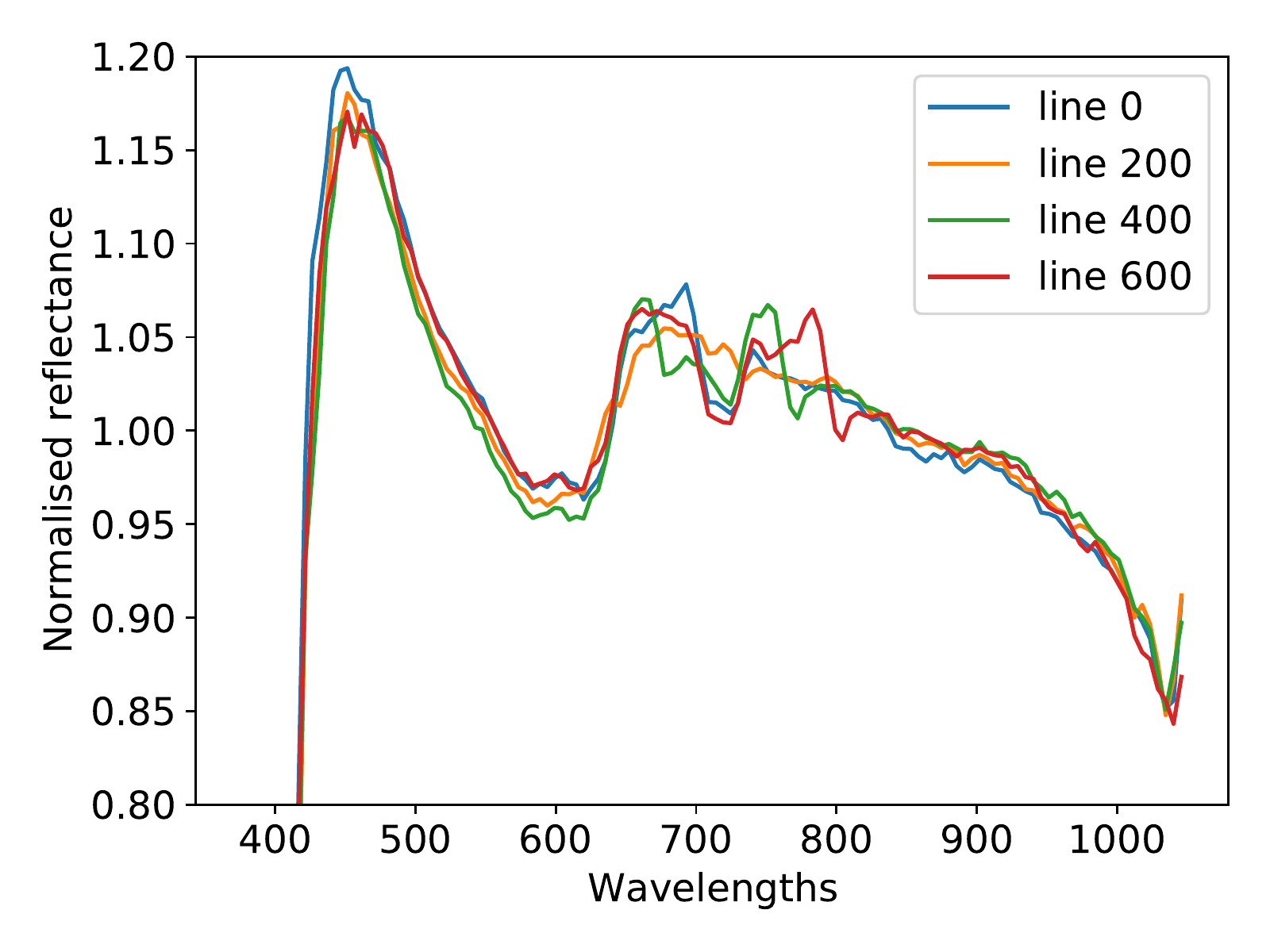}
		\caption{Before reflectance correction}
	\end{subfigure}
	\begin{subfigure}[b]{0.49\linewidth}
		\includegraphics[width=1.0\linewidth]{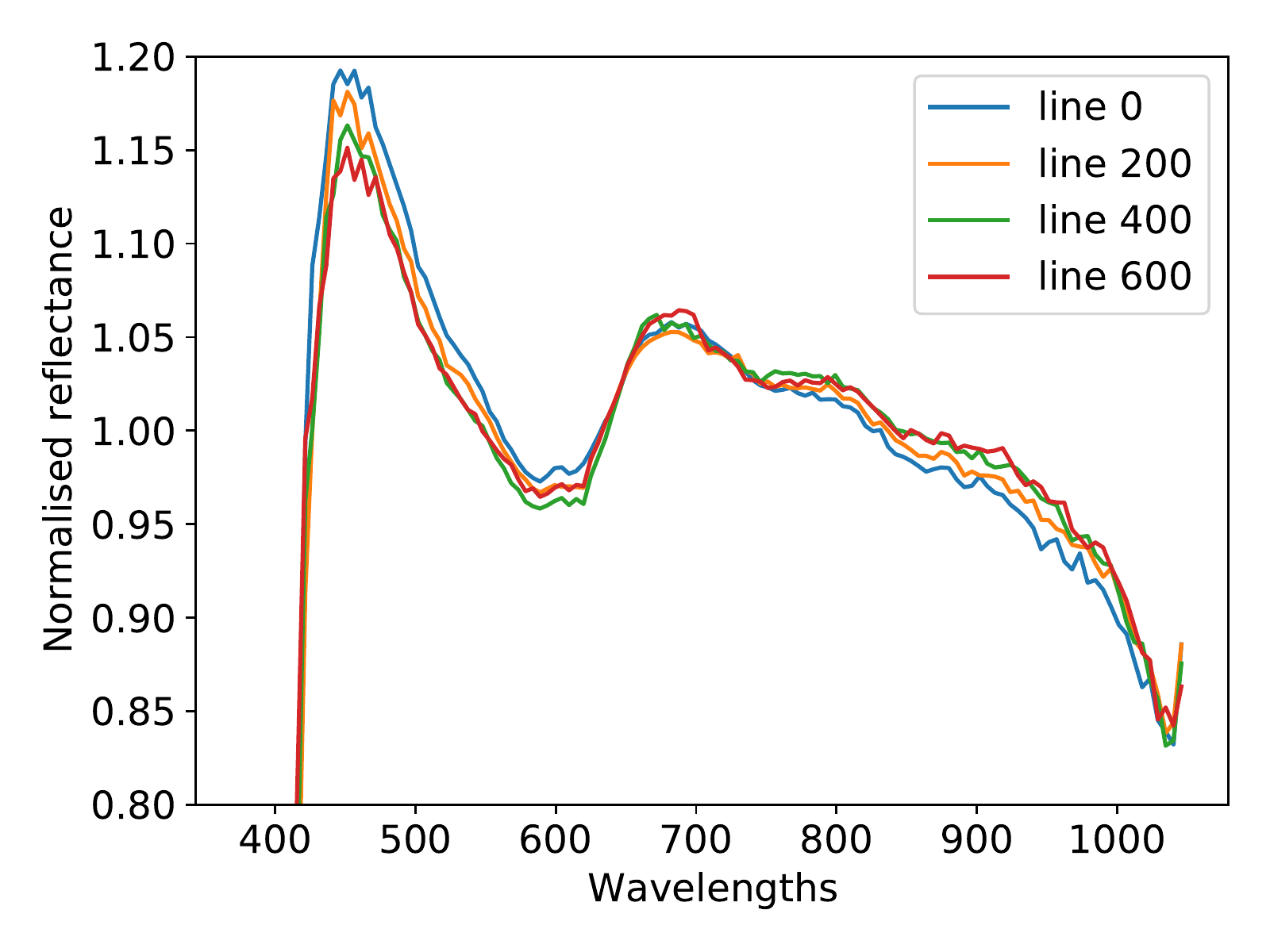}
		\caption{After reflectance correction}
	\end{subfigure}
	\caption{Visualisation of the impact of reflectance correction on reference image spectra. The imaged area was a white paper sheet. Each spectrum represents the mean spectrum of 100 pixels from one horizontal line in the image, normalised by dividing this spectrum by its median. In panel~(a), before correction distortions in range 600-850~nm are visible. In panel~(b), after correction these distortions were removed. The plot has been zoomed in to the y-axis (reflectance) range of $\langle0.8,1.2\rangle$.}
	\label{fig:correction}	
\end{figure}

We have observed that the spectra from the camera contain visible distortions that correlate with pixel horizontal position. These distortions result from equipment error and introduce artefacts with a constant shape, amplitude dependent on pixel luminosity and spectral position dependent on pixel horizontal coordinate. To correct these distortions, we use data from reference hyperspectral image of grey Munsell Color panel (Munsell Notation: N5, 18\% reflectance), for which the spectral characteristics are known.

To correct an image $P$, an image of the grey panel $G$ captured with the camera  was used along with the panel reference spectrum $\vect{h}$, registered with a different instrument. For every pixel $\vect{p}\in P$, we estimate its corrected version $\vect{p}'$ based on corresponding pixel from the panel $\vect{g}_p\in G$ and reference $\vect{h}$ with the formula
\begin{equation}
\vect{p}' = \vect{p} - (\vect{g}_p - \vect{h})\frac{\delta_p}{\delta_g}
\end{equation}
where $\delta_p=\mathrm{median}(\vect{p})$, $\delta_g=\mathrm{median}(\vect{g}_p)$ and $\vect{p}, \vect{g}_p, \vect{h}\in R^d$. 

Visualisation of the correction effect is presented in Fig.~\ref{fig:correction}. This correction has the effect of estimating the artefact based on differences between measured and reference panel data $(\vect{g}_p - \vect{h})$. Spatial correspondence with the pixel being corrected takes care of accuracy of artefact shape, as it varies with pixel position. The scaling coefficient $\frac{\delta_p}{\delta_g}$ accounts for the luminosity factor. We have performed numerous experiments to verify that this correction improves the quality of the spectral information and does not introduce additional noise, including verification with substances of known spectra and statistical tests for presence of artefact-related information within the data.

\subsection{Dataset annotation}
Annotation of blood and blood-like substances was performed by authors. The criteria of assigning a pixel to a class was the prior knowledge about where the substance was applied, visibility of the substance in the image and responses of multiple detectors searching for spectra selected in the image. The correctness of annotation was verified by comparing a subset of images annotated by multiple experts.

In some cases such as the tomato concentrate -- class~$6$ in Fig.~\ref{fig:scene:spectra} -- the visible stain can be divided into two clearly different areas of intensity. For substances other than blood, only the more pronounced area was designated. For the blood, pixels where it is clearly visible were annotated as class~$1$ \emph{`blood'}. Pixels where blood occurrence is uncertain or indistinct were annotated with a separate class denoted as class~$8$  \emph{`uncertain blood'}.

\subsection{Spectral properties of blood}
\label{subsec:blood_properties}
\begin{figure}
	\centering
	\begin{subfigure}[b]{0.49\linewidth}
		\includegraphics[width=1.0\linewidth]{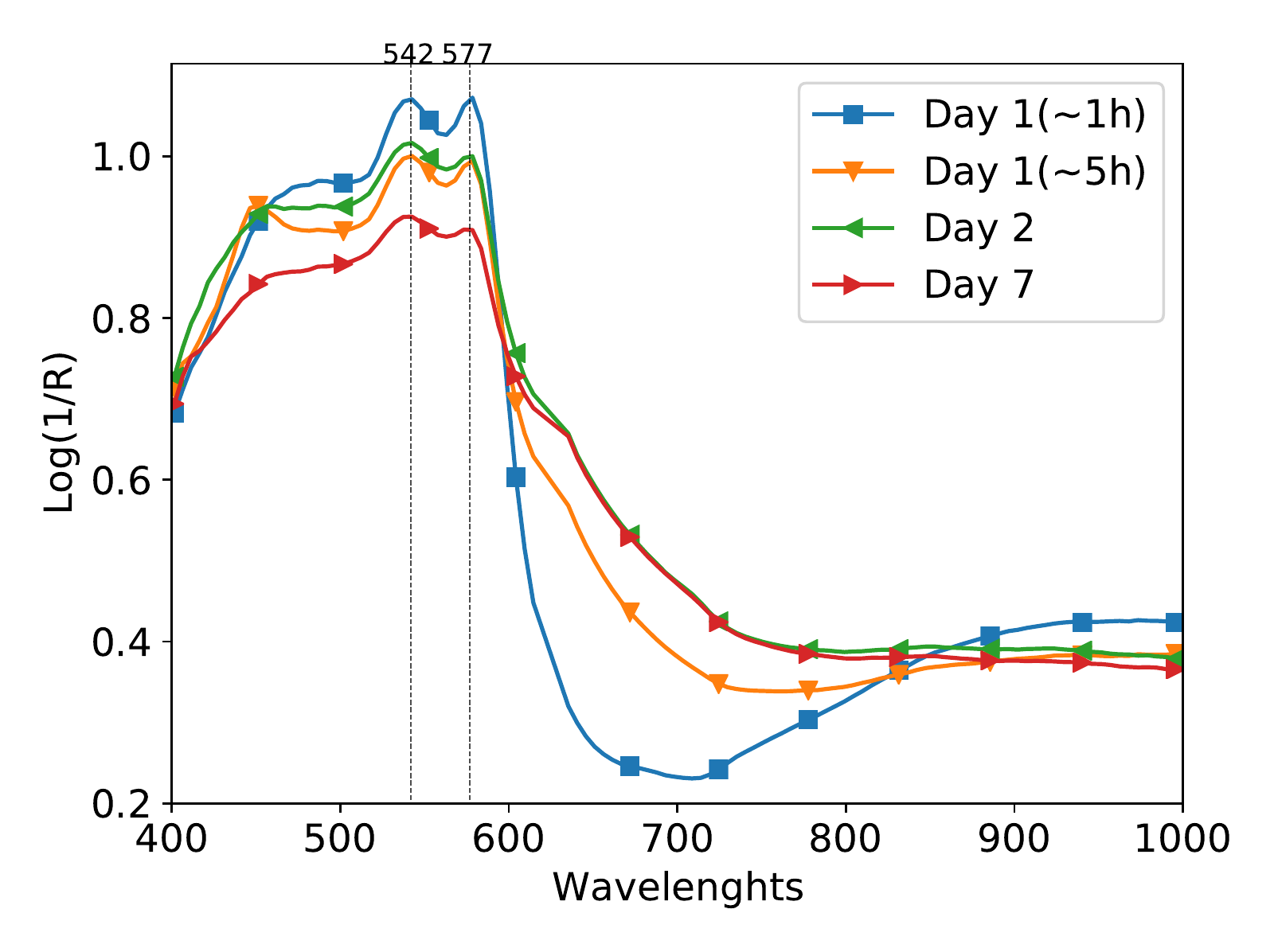}
		\caption{}
		\label{fig:blood_properties:blood}
	\end{subfigure}
	\begin{subfigure}[b]{0.49\linewidth}
		\includegraphics[width=1.0\linewidth]{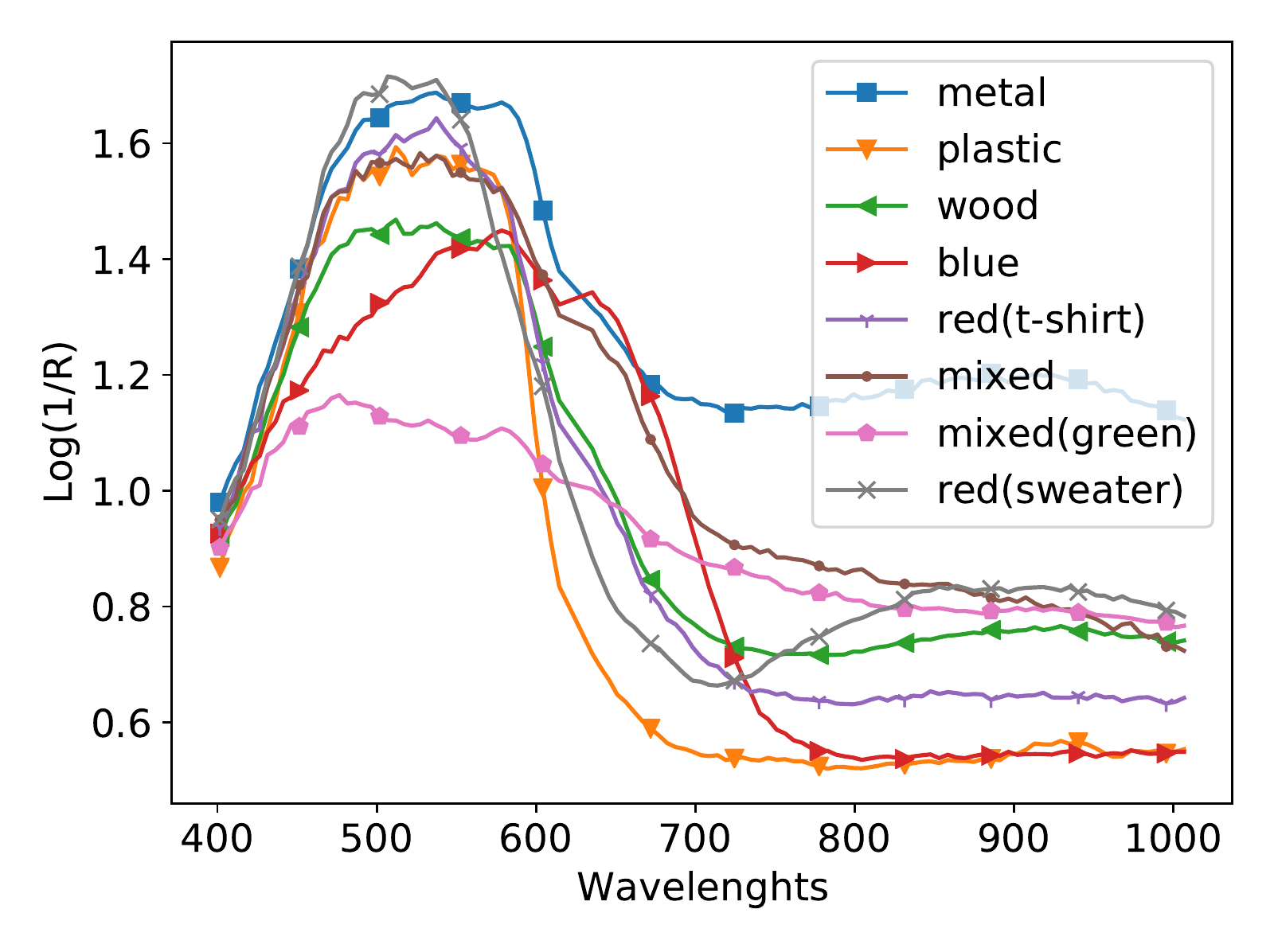}
		\caption{}
		\label{fig:blood_properties:background}
	\end{subfigure}
	\begin{subfigure}[b]{0.49\linewidth}
		\includegraphics[width=1.0\linewidth]{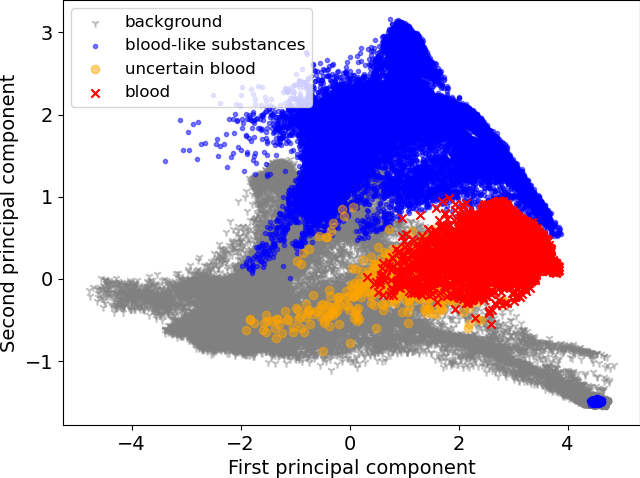}
		\caption{}
	\end{subfigure}
	\begin{subfigure}[b]{0.49\linewidth}
		\includegraphics[width=1.0\linewidth]{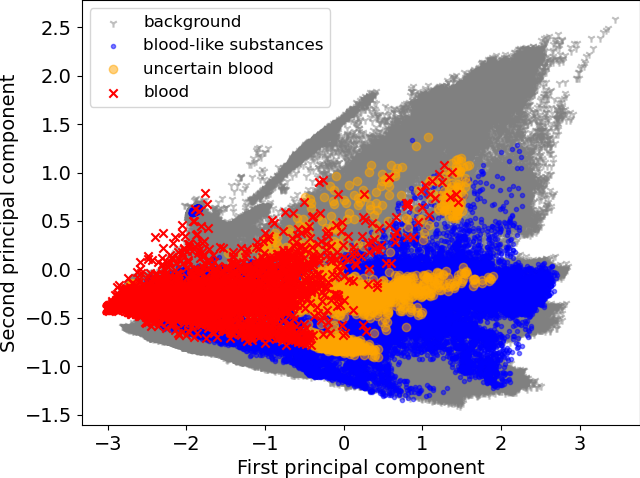}
		\caption{}
	\end{subfigure}
	\caption{Panel~(a): differences between averaged blood spectra captured over the week. The marked frequencies, corresponding to the so-called $\alpha-$ and $\beta-$bands, are associated with the spectral response of haemoglobin. Panel~(b): Differences between blood spectra on different backgrounds in the scenario \textit{E(1)}. Spectra plotted in (a) and (b) were averaged over the image. Bottom: PCA plots where data is projected on first two principal components in scenarios \textit{F(1)} on panel~(c)  and \textit{E(1)} on panel~(d).} 
	\label{fig:blood_properties}
\end{figure}

The impact of the day of acquisition on blood spectra is presented in Fig.~\ref{fig:blood_properties:blood}. The time-related change of spectra is visible, especially at the point of $\alpha$ and $\beta$ bands. Those changes are consistent with those reported in the literature~\cite{majda2018hyperspectral}. This is a result of the oxyHb-metHb-HC degradation~\cite{zadora2018pursuit}; during the first day spectra change significantly due to oxidation of haemoglobin while changes in the following days are minor. This change is most pronounced within the first hours, as this phase is more dynamic~\cite{Bremmer2011biphasic}; this suggests that detection of fresh blood is a more complex problem due to additional spectra variability.  The Fig.~\ref{fig:blood_properties:background} also presents differences in blood spectra resulting from different background materials. The influence of the background is considerable, which makes the detection problem more challenging. However, it is known that application of multivariate data analysis algorithms can provide reliable results in the presence of confusing spectra~\cite{yang2016spectral} or when used on coloured, even dark coloured backgrounds~\cite{edelman2012identification}.

\subsection{Dataset and code availability}
The dataset consist of 14 hyperspectral images in ENVI format, each $\sim180$~MB in size, along with documentation and target annotations. In accordance with open-access mandate, the dataset is publicly available on-line\footnote{Dataset DOI: 10.5281/zenodo.3984905}
under an open license. In addition, code for experiments performed in this paper are also available under  GNU General Public License \footnote{Source Code: \url{https://github.com/iitis/HSI_blood_detection.git}}

\section{Methods}

\label{sec:methods}
\label{sec:target_detection}
We formulate the task of detecting blood in the imaged scene as a hyperspectral target detection problem. The aim of the detector is to find hyperspectral pixels where the target spectrum is present. The choice of the detection method is based on the premise that we are looking for full-pixel targets.

\subsection{Hyperspectral target detection with likelihood ratio detector}
\label{sec:lr_detectors}

We treat a hyperspectral image as a set of pixels $\vect{x}\in\set{X}\subset\R^{N\times d}$, where $N$ is the number of pixels in the image and $d$ is the number of bands. We view the task of hyperspectral target detection, following~\cite{manolakis2003hyperspectral,manolakis2009there}, as the choice between two competing hypotheses,
\begin{equation*}
\begin{aligned}
&H_0: \text{(target absent)}\\
&H_1: \text{(target present)}.
\end{aligned}
\end{equation*}
Let $\mathcal{L}(H_0|\vect{x})$ and $\mathcal{L}(H_1|\vect{x})$ be likelihoods of $H_0$ and $H_1$ given some observed $\vect{x}$, called the pixel under test (PUT). Their quotient, the likelihood ratio $\Lambda(\vect{x})=\frac{\mathcal{L}(H_1|\vect{x})}{\mathcal{L}(H_0|\vect{x})}$, can be used to make a decision about the $\vect{x}$. Given some threshold $\eta$, if $\Lambda(\vect{x})>\eta$ tests true, we reject the $H_0$ in favour of $H_1$. By the Neyman–Pearson lemma, the test based on the likelihood ratio is the most powerful statistical test. Because of its role, the $\Lambda(\vect{x})$ is also called the detection statistics.

Since detectors based on Gaussian probability models often lead to good performance, as concluded in~\cite{manolakis2003hyperspectral,manolakis2009there}, a reasonable starting point could be the assumption that both the target and the background spectra can be described by multivariate Gaussian distributions. In that case, $\mathcal{L}(H_0|\vect{x}) = P_{\vect{\mu},\vect{\Gamma}}(\vect{x})$, or the probability of $\vect{x}$ under a multivariate normal distribution with mean $\vect{\mu}$ and a covariance matrix $\vect{\Gamma}$. Depending on how parameters are defined for background and target probability density functions, this leads to a family of Gaussian-based likelihood ratio detectors, among them the Quadratic Detector and the Matched Filter. 

\subsection{Quadratic detector}

A detection problem with Gaussian distributions can be represented in the form of the following hypotheses~\cite{manolakis2003hyperspectral}:
\begin{equation}
\begin{aligned}
&H_0: \vect{x}\sim \mathcal{N}(\vect{\mu}_0,\vect{\Gamma}_0)\\
&H_1: \vect{x}\sim \mathcal{N}(\vect{\mu}_1,\vect{\Gamma}_1),
\end{aligned}
\end{equation}
where $\mathcal{N}(\cdot,\cdot)$ denotes a multivariate Gaussian distribution, vector $\vect{\mu}_0$ and matrix $\vect{\Gamma}_0$ are the mean and covariance matrix of the background, vector $\vect{\mu}_1$ and matrix $\vect{\Gamma}_1$ are the mean and covariance matrix of targets.
Provided that the parameters are all known in advance, we can use the Quadratic Detector~\cite{manolakis2003hyperspectral} (QD):
\begin{equation}
\begin{aligned}
\label{eq:qd}
y = D_{\text{QD}}(\vect{x},\vect{\mu}_0,\vect{\mu}_1,\vect{\Gamma}_0,\vect{\Gamma}_1)\\
=
(\vect{x}-\vect{\mu}_0)^T\vect{\Gamma}_0^{-1}(\vect{x} 
-
\vect{\mu}_0)-(\vect{x}-\vect{\mu}_1)^T\vect{\Gamma}_1^{-1}(\vect{x}-\vect{\mu}_1),
\end{aligned}
\end{equation}
which compares the Mahalanobis distance of the pixel under test to means of target and background distribution.

\subsection{Matched filter}

\label{sec:mf}
The use of QD is subject to assumptions in the form of knowledge about the distribution of the target and the background in the image. This assumption can be viewed as too restricting; we can limit our knowledge of the target to spectral signature $\vect{\mu}_t$ alone and assume that the target and background classes have the same covariance matrix $\vect{\Gamma}_0=\vect{\Gamma}_1\equiv\vect{\Gamma}$. The use of Neyman–Pearson lemma leads us to the Matched Filter (MF) detector: 
\begin{equation}
\label{eq:MF}
y = D_{\text{MF}}(\vect{x},\vect{\mu},\vect{\mu}_t,\vect{\Gamma})
=
\frac{(\vect{x}-\vect{\mu})^T\vect{\Gamma}^{-1}(\vect{\mu}_t-\vect{\mu})}{(\vect{\mu}_t-\vect{\mu})^T\vect{\Gamma}^{-1}(\vect{\mu}_t-\vect{\mu})}
\end{equation}
which measures the distance along a projection onto the line connecting target $\vect{\mu}_t$ and data mean $\vect{\mu}$, scaled by the data covariance matrix. Without the need to provide separate parameters for target and background distributions, parameters estimated from the data i.e. the sample mean $\vect{\mu}$ and sample covariance matrix $\vect{\Gamma}$ are commonly used, simplifying the application of this detector.

\subsection{Comparison of QD and MF}
\label{sec:qdmf}

\begin{figure*}
	\centering
	\begin{subfigure}[b]{0.45\linewidth}
		\includegraphics[width=1.0\linewidth]{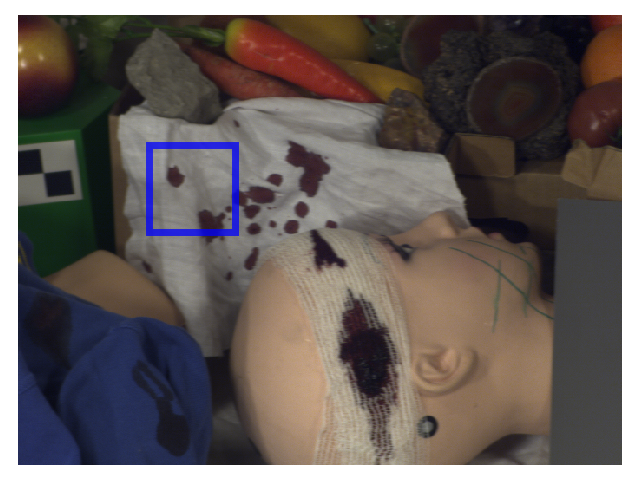}
		\caption{RGB rendering of image \emph{A(1)}; blue box marks the pixels selected for the filter visualization.}
		\label{fig:filters:image}
	\end{subfigure}\hspace{0.05\linewidth}
	\begin{subfigure}[b]{0.45\linewidth}
		\includegraphics[width=1.0\linewidth]{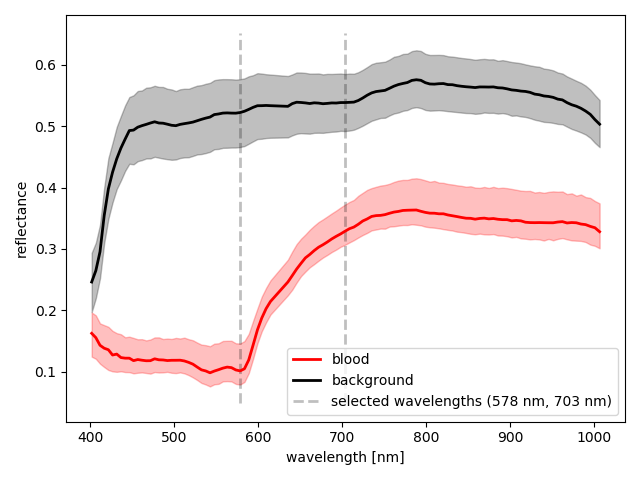}
		\caption{Averaged spectra of blood and background pixels from the selected area. Two wavelengths, one close to the $\beta$ band (see~\ref{sec:relatedwork}) and one distant, are used for the visualization.}
		\label{fig:filters:spectra}
	\end{subfigure}
	\begin{subfigure}[b]{0.45\linewidth}
		\includegraphics[width=1.0\linewidth]{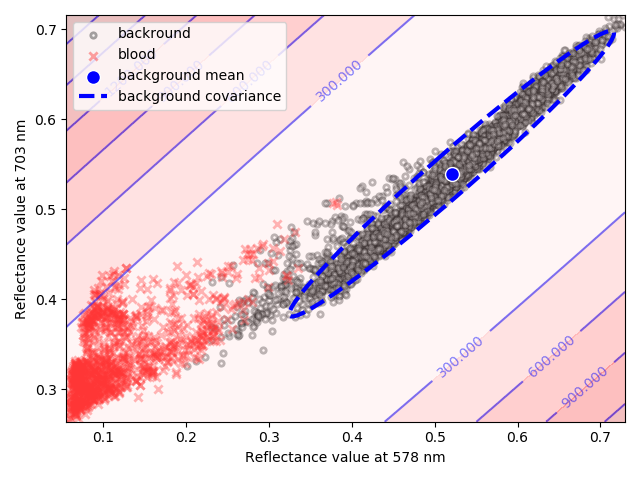}
		\caption{Background component of the QD -- first part of the Equation~\ref{eq:qd} -- measures Mahalanobis distance to the non-blood pixels.}
		\label{fig:filters:qdb}
	\end{subfigure}\hspace{0.05\linewidth}
	\begin{subfigure}[b]{0.45\linewidth}
		\includegraphics[width=1.0\linewidth]{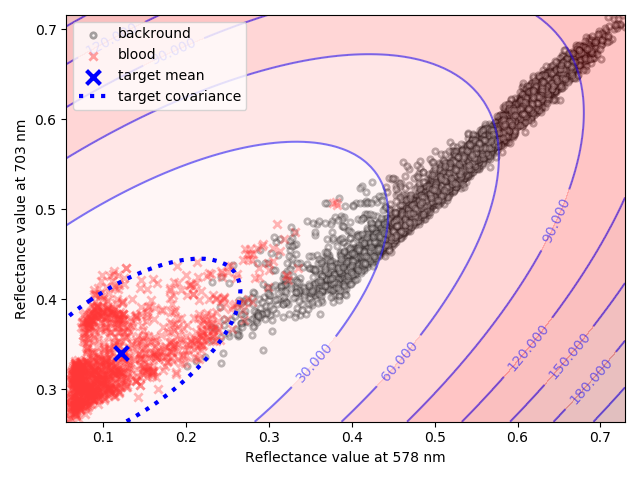}
		\caption{Target component of the QD -- second part of the Equation~\ref{eq:qd} -- measures Mahalanobis distance to the blood pixels.}
		\label{fig:filters:qdt}
	\end{subfigure}
	\begin{subfigure}[b]{0.45\linewidth}
		\includegraphics[width=1.0\linewidth]{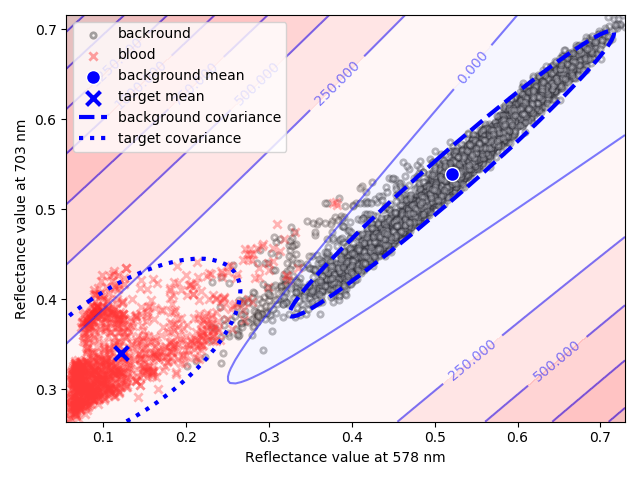}
		\caption{Full QD -- whole Equation~\ref{eq:qd} -- difference of background and target distances, or how far the points are from the background and how close to target at the same time.}
		\label{fig:filters:qd}
	\end{subfigure}\hspace{0.05\linewidth}
	\begin{subfigure}[b]{0.45\linewidth}
		\includegraphics[width=1.0\linewidth]{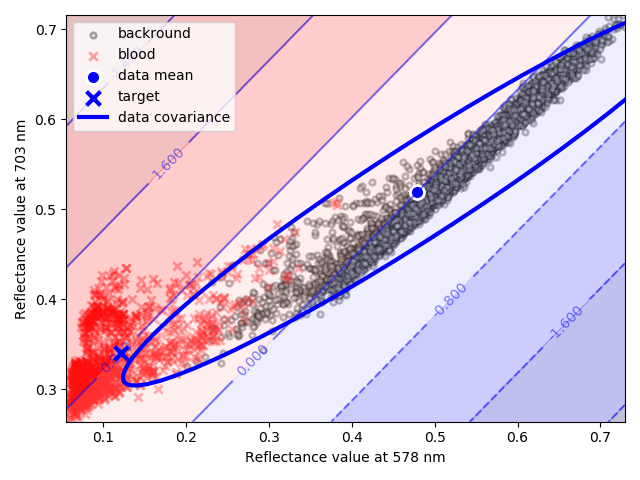}
		\caption{MF -- Equation~\ref{eq:MF} -- projects the data onto a line connecting data mean and the target signature, distance measures scaled by the inverse of the data covariance matrix.}
		\label{fig:filters:mf}
	\end{subfigure}
	\caption{Visualization of a simple case of Quadratic Detector and Matched Filter. Each detector produces, for each pixel -- here reduced to two selected wavelengths -- a scalar value corresponding to the probability of it being the target substance, presented as annotated contour plot. The differences are in target and background models used. Both detectors assume Gaussian data model; note that while real-life pixel distribution deviates from the Gaussian significantly, it is still a reasonable model, in particular considering the covariance of the wavelengths values. All covariance ellipses are plotted at $95\%$ confidence intervals.} 
	\label{fig:filters}
\end{figure*}

A case study of an application of QD and MF is presented in Fig.~\ref{fig:filters}. The QD formula has two components: background distance (see Fig.~\ref{fig:filters:qdb}) and target distance (see Fig.~\ref{fig:filters:qdt}). Mahalanobis distance is used, as Euclidean distance would be affected by covariances and variance differences. The essence of QD is the addition of two estimates i.e. how far is the PUT from the background and how close to the target, which are combined into one value (see Fig.~\ref{fig:filters:qd}). This can be done when either the means and covariance matrices are known in advance, or some set of labelled points exist that allows for respective sample estimates. MF, on the other hand, uses all data covariance for distance scaling, and projects the data along the line where the background-target difference is expected to be the most pronounced. This produces a detector which can be expected to be less effective, but much easier to apply. 

Both QD and MF assume that the data distribution is Gaussian. While the actual data distribution will often deviate from that model, nevertheless a Gaussian may provide a well-working approximation, in particular capturing the covariance present in the data set. 

\subsection{Evaluation of a detector effectiveness}
\label{subsec:detector effectiveness}

Given a detector with known detection statistics for some value of threshold $\eta$, the decision made by the detector can be represented as a Confusion Matrix (CM), whose diagram is presented as the Tab.~\ref{tab:conf_mat}. CM divides detection results between four types of outputs: True Positives (TP) are pixels where target was correctly detected, False Positives (FP) refer to pixels where target was detected incorrectly, True Negatives (TN) are pixels where target is correctly not detected and False Negatives (FN) refer to pixels where target was present but not detected.
\begin{table}
	\centering
	\begin{tabular}{|c|c|c|}
		\hline
		&  actual $H_1$&actual $H_0$\\
		\hline
		predicted $H_1$&TP&FP\\
		\hline
		predicted $H_0$&FN&TN\\
		\hline
	\end{tabular}
	\caption{A Confusion Matrix}
	\label{tab:conf_mat}
\end{table}
The confusion matrix is the basis to define common metrics used to evaluate detector's performance: True Positive Rate, $\text{TPR}=\frac{\text{TP}}{TP+FN}$, False Positive Rate, $\text{FPR}=\frac{\text{FP}}{FP+TN}$, $\text{Recall}=\frac{\text{TP}}{TP+FN}$ and $\text{Precision}=\frac{\text{TP}}{TP+FP}$.

However, it is not obvious how to set the threshold $\eta$, as increasing its value increases the probability of correct detections $P_{TP}$ but at the same time increases the probability of false positive detection (false alarms) $P_{FP}$. To evaluate a detector across a range of $\eta$ values, the receiver operating characteristics (ROC) curve are used~\cite{davis2006relationship}. ROC curve plots the probability $P_{TPR}(\eta)$ versus the probability $P_{FPR}(\eta)$ for possible values of the threshold $\eta$. The performance of the detector can be expressed as a numerical value in the form of the Area Under Curve (AUC) statistics, which assumes values in the range $\langle0, 1\rangle$, with $AUC=1.0$ corresponding to an `oracle' detector that assigns highest values of detection statistics to target pixels and $AUC=0.5$ corresponding to a detector that assigns target labels at random.

Imbalance in target and background class sizes, in particular when the number of correct detections is small in comparison to the number of background pixels, leads to a serious bias towards the majority (background) class. This can make ROC curves less informative for comparing detectors, as values of $AUC_{ROC}$ for evaluated detectors become similar.

For such skewed datasets, Precision-Recall (PR) curves, that plot values of $\text{Precision}(\eta)$ versus $\text{Recall}(\eta)$, can give a more informative picture of an algorithm's performance~\cite{davis2006relationship}. However, when comparing two detectors, both AUC and PR measures give similar results~\cite{davis2006relationship}. 

\subsection{Detection scenarios}
\label{sec:detection_scenarios}
Regarding likelihood ratio detectors presented in the Sec.~\ref{sec:lr_detectors}, we define the following scenarios of their use:

\begin{enumerate}
	\item \emph{Ideal QD}. Target and background means and covariance matrices ($\vect{\mu}_0$, $\vect{\Gamma}_0$, $\vect{\mu}_1$, $\vect{\Gamma}_1$) from the image under test are supplied to the Quadratic Detector, given by the Eq.~\eqref{eq:qd}.
	\item \emph{Ideal MF}. Sample mean and sample covariance matrix of the image under test ($\vect{\mu}$, $\vect{\Gamma}$) are supplied to the MF detector given by the Eq.~\eqref{eq:MF}. Mean target spectrum $\vect{\mu}_t$ is also taken from the image under test, as a mean of the \emph{`blood'} class. 
	\item \emph{Inductive MF}. Sample mean and sample covariance matrix of the image under test ($\vect{\mu}$, $\vect{\Gamma}$) are supplied to the MF detector given by the Eq.~\eqref{eq:MF}. Mean target spectrum $\vect{\mu}_t$ is a sample mean spectrum of the \emph{`blood'} class from another image in the dataset. This approach is similar to the problem of inductive learning in ML, where a model trained on one set of representative samples is used to make predictions on a different set.
	\item \emph{MF}$_{lib}$. Sample mean and sample covariance matrix of the image under test ($\vect{\mu}$, $\vect{\Gamma}$) are supplied to the MF detector given by the Eq.~\eqref{eq:MF}. Mean target spectrum $\vect{\mu}_t$ is taken from an external spectral library.
	\item Two-stage estimation. Since some variations between spectra of image under test and external reference are inevitable, e.g. due to background mixing or equipment differences, we propose to use a two-stage approach. First, a sample of blood spectra is detected in the image under test based on an external pattern. Second, the final label assignment is made using the sample of blood spectra detected in the first stage. This approach can be viewed as a combination of \emph{Inductive MF} or \emph{MF}$_{lib}$, followed by the \emph{Ideal MF}. We present the algorithm in Sec.~\ref{sec:detection_algorithm}. 
\end{enumerate}

\begin{figure*}
	\centering
	\begin{subfigure}[b]{0.29\textwidth}
		\includegraphics[width=1.0\linewidth]{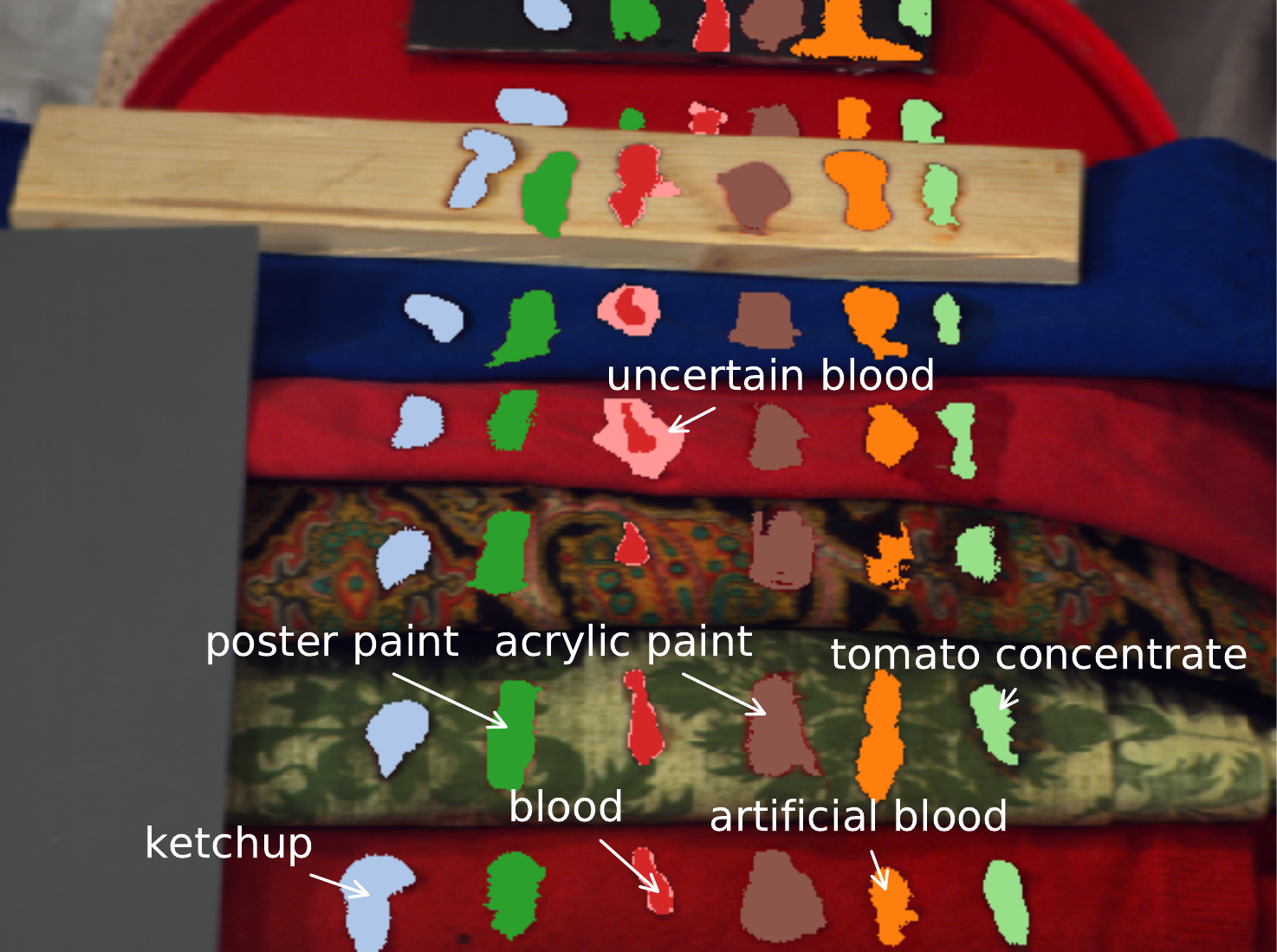}
		\caption{Ground truth}
	\end{subfigure}
	\begin{subfigure}[b]{0.29\textwidth}
		\includegraphics[width=1.0\linewidth]{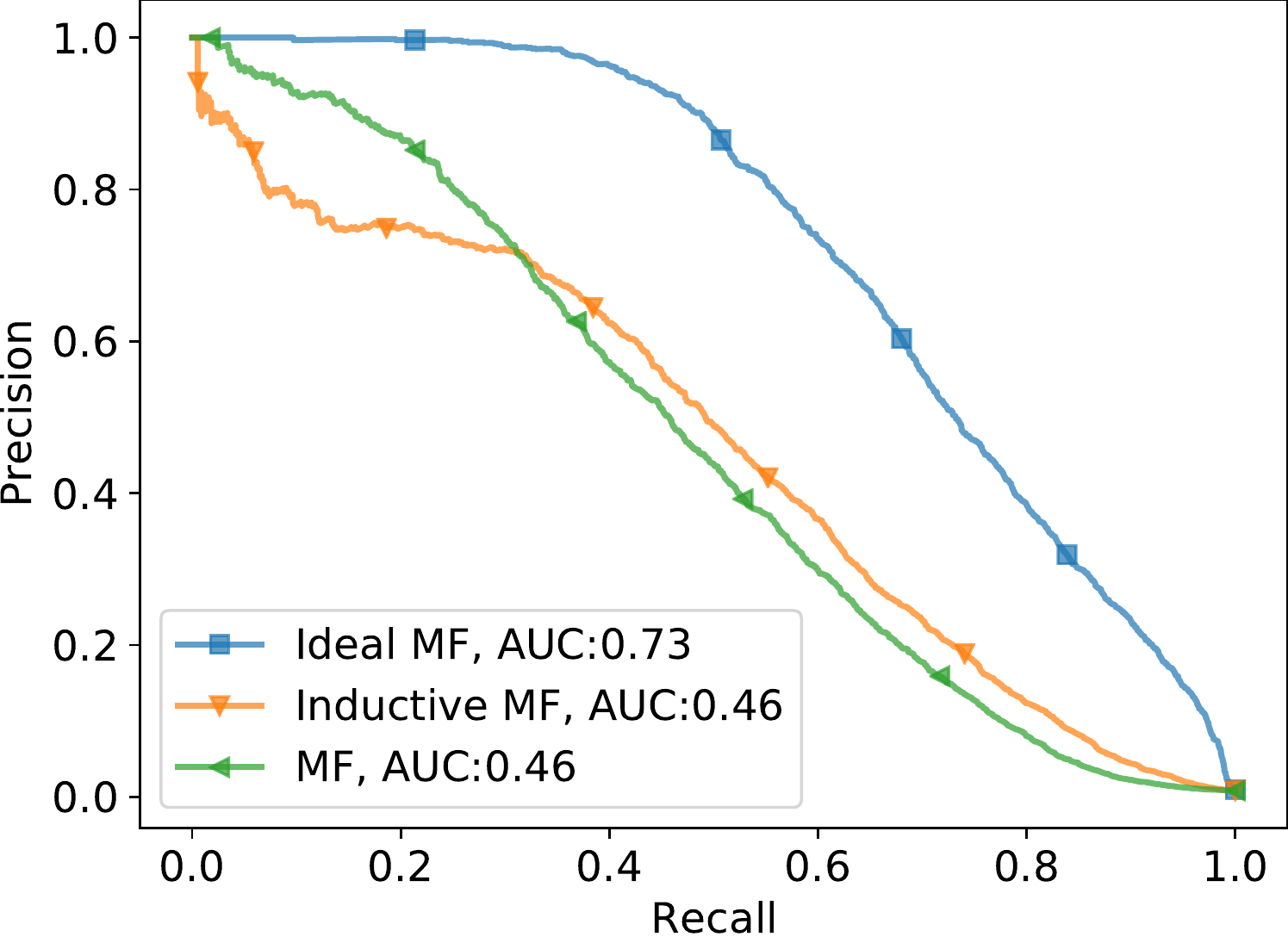}
		\caption{PR curves}
	\end{subfigure}
	\begin{subfigure}[b]{0.29\textwidth}
		\includegraphics[width=1.0\linewidth]{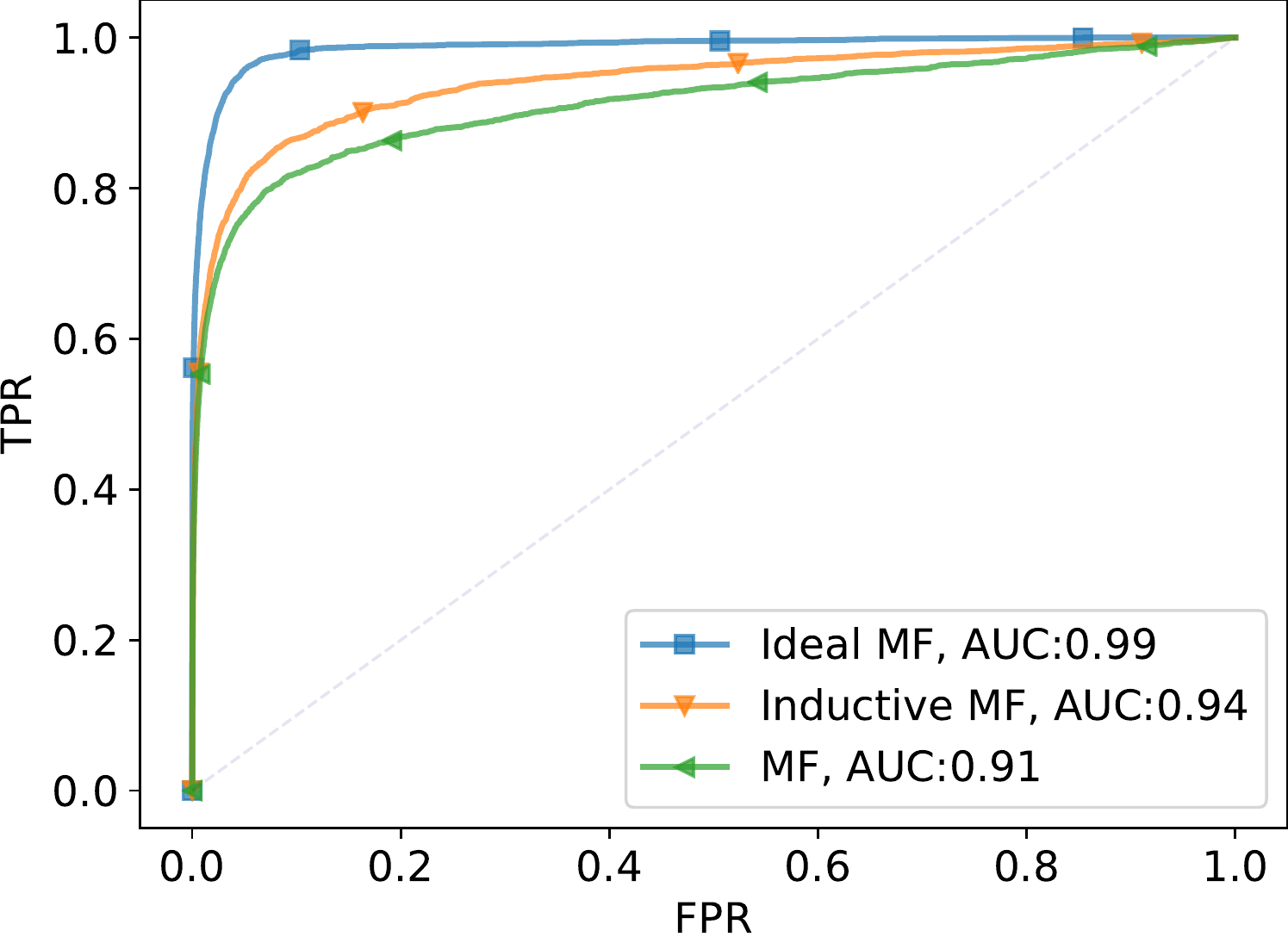}
		\caption{ROC curves}
	\end{subfigure}
	\begin{subfigure}[b]{0.29\textwidth}
		\includegraphics[width=1.0\linewidth]{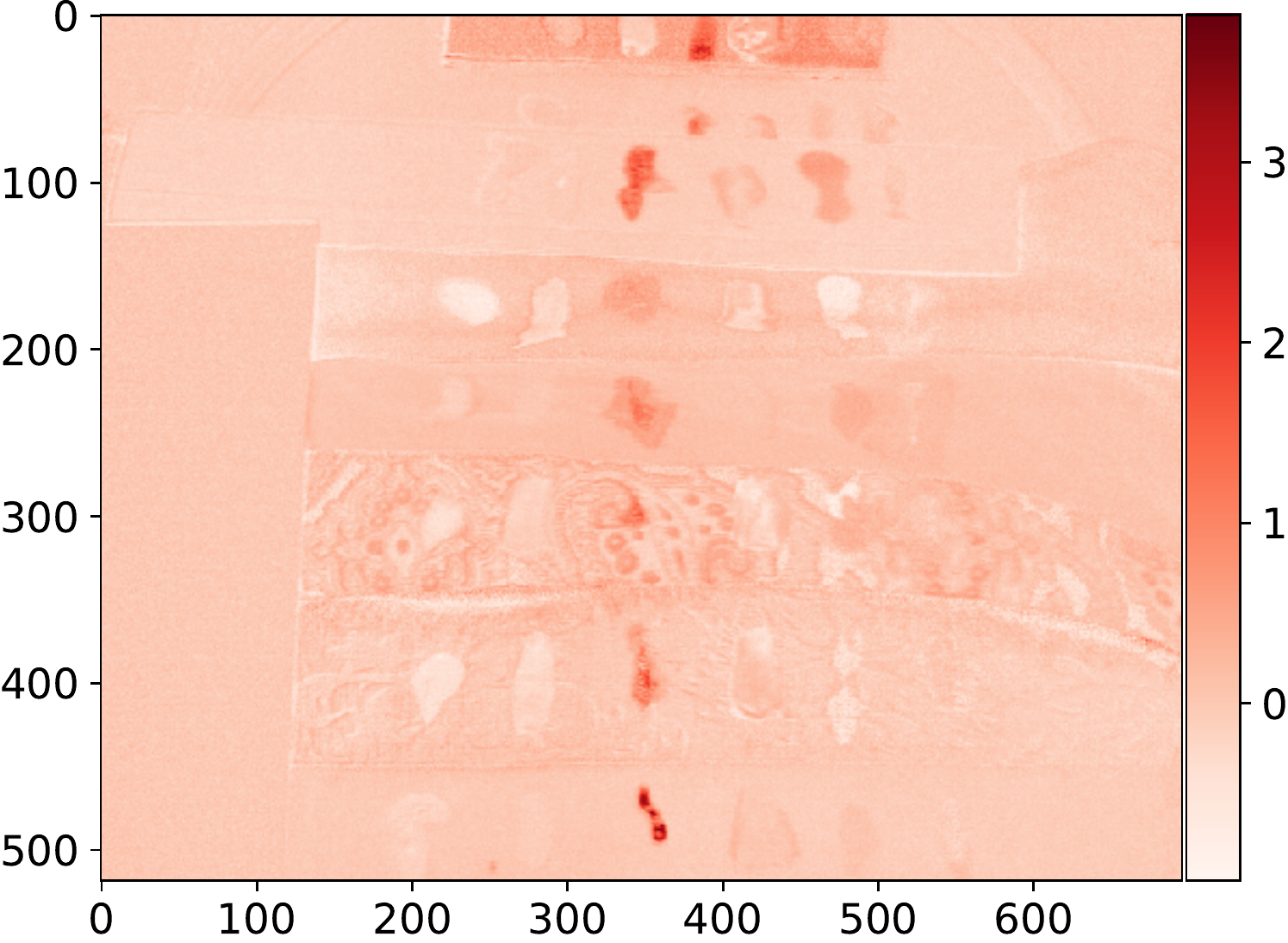}	
		\caption{Output of the \textit{Ideal MF}}
	\end{subfigure}
	\begin{subfigure}[b]{0.29\textwidth}
		\includegraphics[width=1.0\linewidth]{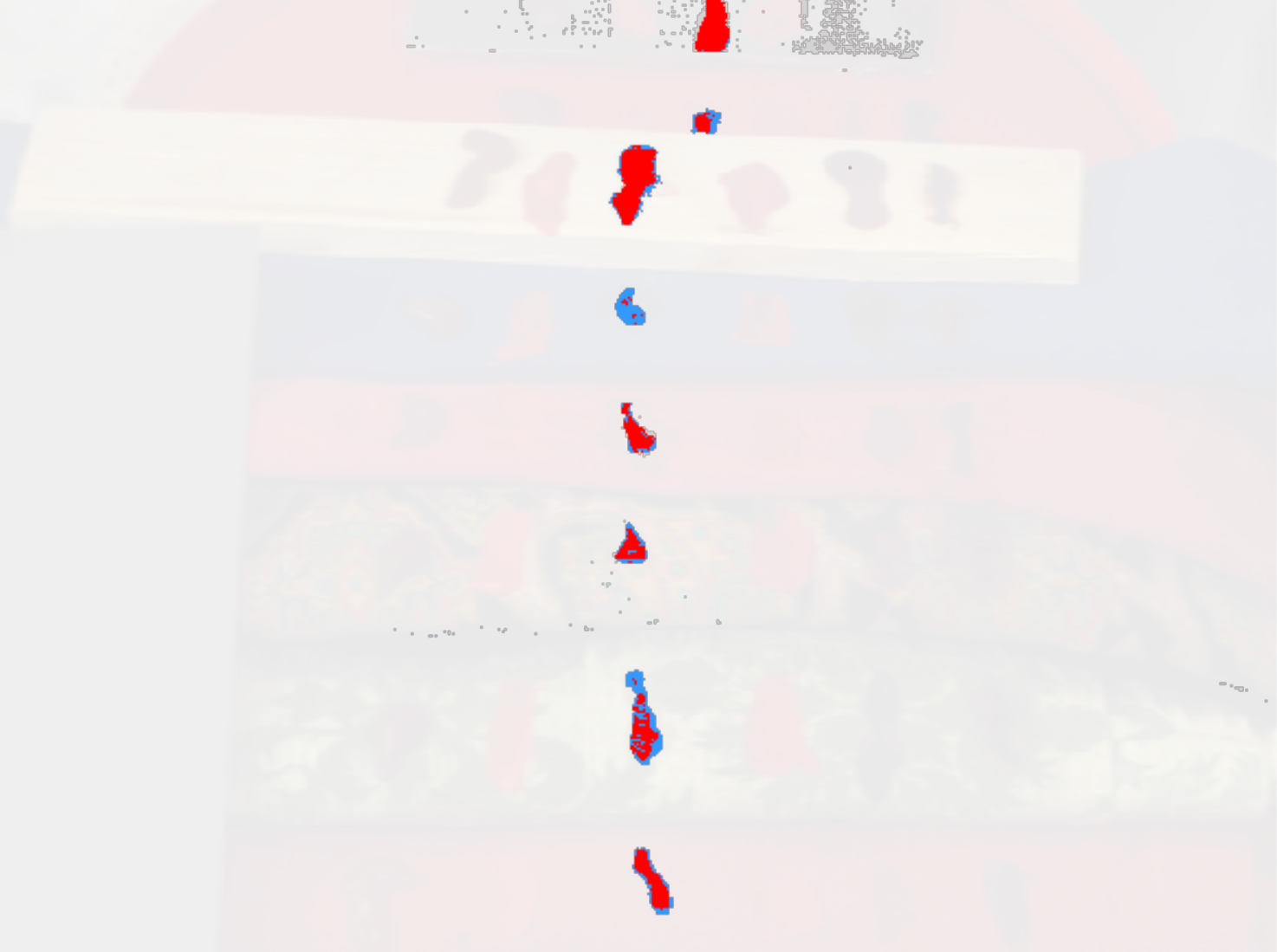}	
		\caption{Detection map for \textit{Ideal MF}}
	\end{subfigure}
	\begin{subfigure}[b]{0.29\textwidth}
		\includegraphics[width=1.0\linewidth]{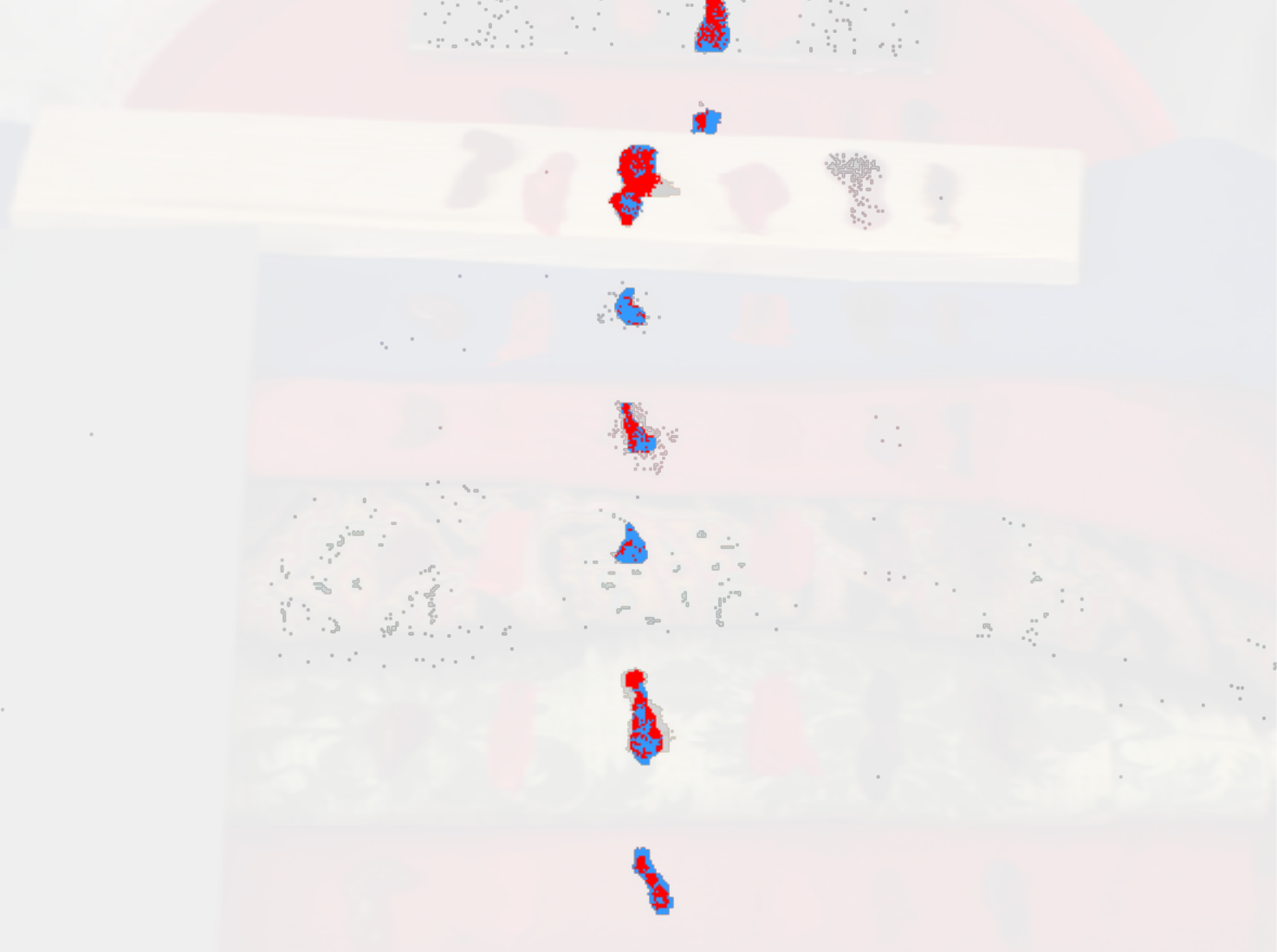}			
		\caption{Detection map for \textit{Inductive MF}}
	\end{subfigure}
	
	\caption{Example: detection performance of three Matched Filter detectors, applied to the scene \textit{E(1)}. The \textit{Ideal MF} is trained on blood spectrum from the image itself, the \textit{Inductive MF} on blood from the image \textit{F(1)} and the MF on spectrum from the spectral library. Blood class is denoted with red colour in the panel~(a). Panels~(b) and~(c) present PR and ROC curves of the detectors. Panel~(d) presents detection score of the \textit{Ideal MF}.  The last two panels are detection maps as overlays on the RGB image, where red colour denotes True Positives (TP), blue colour denotes False negatives (FN) and grey colour denotes False Positives (FP).}
	\label{fig:example}
\end{figure*}

\subsection{Two stage detection algorithm}
\label{sec:detection_algorithm}
The motivation for introducing and extended detector is based on the discussion in~\cite{manolakis2009there} i.e. for complex scenes it is advisable to perform two runs of the detector, with the first used to select `target-free' pixels that accurately characterize the background. The effectiveness of using the first run of the detector to perform a ranking of pixels into `like-blood' and `like-background' has been confirmed in our initial experiments. However, we've noticed that in parallel to selection of `target-free' pixels, we can select `target-rich' ones and use them as a better target template. This way we can possibly approach the high performance of the \emph{Ideal MF} detector trained on the mean spectra of blood in the image and take advantage of the fact that many pixels with high values of MF detector score are True Positives~(TP). An example of this is given in the Sec.~\ref{sec:example}.

We propose to use a two-stage detector. In the first stage an MF will be used to locate pixels in the image that are likely from the target class. A mean spectrum of these pixels will be used to create a new, `local' estimation of a target spectral signature. This new target signature will be used in the second stage, to train a final MF detector that will output pixel detection scores. A description of this procedure is provided as Algorithm~\ref{algorithm:tsmf}.

\begin{algorithm}[h]
	\KwData{Target spectral signature $\vect{\mu}_t\in\R^d$, Array of pixels $\vect{X}\subset\R^{K\times d}$, parameter $\text{N}$}
	\KwResult{Pixel scores $\vect{Y}\in\R^K$}
	\tcc{Stage~1}
	$\vect{\mu}\longleftarrow$ \text{sample mean from} $\vect{X}$\\
	$\vect{\Gamma}\longleftarrow$ \text{sample covariance matrix from} $\vect{X}$\\
	$\vect{Y}\longleftarrow \emptyset$\\
	\For{pixel $\vect{x}$ in $\vect{X}$}
	{
		$\vect{Y}\longleftarrow D_{\text{MF}}(\vect{x},\vect{\mu},\vect{\mu}_t,\vect{\Gamma})$
	}
	\tcc{Stage~2}
	$a\longleftarrow \mathrm{argsort}(\vect{Y})$\\
	\tcc{$\vect{X}[a]$ denotes sorting $\vect{X}$ by $a$}
	$\vect{X}'\longleftarrow \text{first}\ \text{N}\ \text{values of}\ \vect{X}[a]$\\
	$\vect{\mu}_t'\longleftarrow \text{mean}(\vect{X}')$\\
	$\vect{Y'}\longleftarrow \emptyset$\\
	\For{pixel $\vect{x}$ in $\vect{X}$}
	{
		$\vect{Y'}\longleftarrow D_{\text{MF}}(\vect{x},\vect{\mu},\vect{\mu}'_t,\vect{\Gamma})$
	}	
	\KwRet{$\vect{Y'}$}
	\caption{A two-stage detection algorithm based on the Matched Filter, described in Sec.~\ref{sec:mf}. In the first stage, Eq.~\eqref{eq:MF} is used to compute detection statistics for pixels. In the second stage a new target spectral signature $\vect{\mu}_t'$ is computed as an average of $N$ highest scoring pixels from Stage~1, and the MF detector is reapplied.}
	\label{algorithm:tsmf}
\end{algorithm}

\subsection{Experimental details}

The experiment applies scenarios 2-5, including Algorithm~\ref{algorithm:tsmf}, to images from the dataset described in Sec.~\ref{sec:scenarios_and_data}. 

\paragraph{Data processing} The noisy bands [0-4], [122-128] and [48, 49, 50] were removed prior to the experiments, leaving 113 bands. In order to compensate for uneven illumination of the imaged scene, hyperspectral cubes were normalised by dividing every hyperspectral pixel by its median. 

\paragraph{External spectra} are taken from the spectral blood library provided by authors of~\cite{majda2018hyperspectral}. Target spectra in the library were selected to be similar in age to the blood in the image. The approximate age of target spectra is presented in the Tab.~\ref{tab:target_age}. To align spectral ranges, wavelengths and the number of bands of the external library to our dataset, we have used linear interpolation on library spectra to obtain values corresponding to our camera.

\paragraph{Parameters} The parameter of the Algorithm~\ref{algorithm:tsmf} i.e. the number of pixels $N$ used to obtain the new target signature in the second stage was set in our experiments to $N=1000$. Sensitivity analysis of this parameter is further discussed in the Sec.~\ref{sec:sensitivity_analysis}.

\paragraph{Implementation} Experiments were implemented in Python 3.6.9 using libraries: numpy 1.16.4, scipy 1.3.1, scikit-learn 0.22.1, spectral 0.19, matplotlib 3.2.2. Experiments were conducted using a computer with Intel(R) Core~i7-5820K~CPU~@~330GHz with 64GB of RAM and with the Windows 10 Pro system. The running time of detection algorithms did not exceed tens of seconds.

\section{Results}
\label{sec:experiments_and_results}
\label{sec:example}
Results of applications of scenarios 2-4 to one of the images in the dataset, namely the scene \textit{E(1)} are presented in Fig.~\ref{fig:example}. The panel~(a) presents the ground truth; the \emph{`blood'} class is denoted with a red colour. Panels~(b) and~(c) present the detection performance in the form of respectively PR and ROC curves, with the aggregated performance expressed as Area Under Curve (AUC). The AUC value is higher in the \emph{Ideal MF} than in the remaining two scenarios.

Panel~(d) presents detection statistics, or pixel scores, of the \textit{Ideal MF} as a heat map. Areas corresponding to the blood location are visibly darker, which corresponds with higher value of the score for a given pixel. The two remaining plots present a coloured detection map for \emph{Ideal MF} and \textit{Inductive MF} scenarios. This map is obtained by setting the threshold $\eta$ (see Sec.~\ref{sec:lr_detectors}) to the actual ratio of blood pixels in the image i.e. $\eta=\frac{\text{no. blood pixels}}{\text{no. all pixels}}$. Colours in the map correspond to True Positive detections as well as two types of errors, False Negatives and False Positives. Notable errors, visible in the detection map presented in the panel~(e), include FP detection of class~$3$ \emph{`artificial blood'}, visible in the top area of the image. Errors for the \emph{Inductive MF}, presented in the last panel are more numerous, which is consistent with a lower AUC than the \emph{Ideal MF} observed in panels~(b) and~(c).

\begin{figure}
	\centering
	\includegraphics[width=0.9\linewidth]{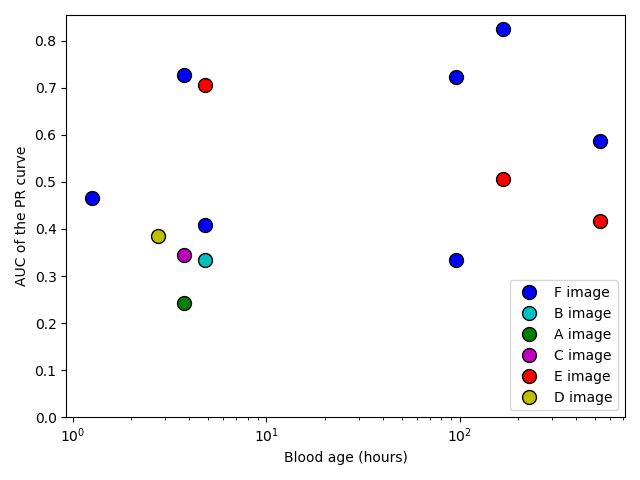}
	\caption{The scatterplot of performance (measured as AUC of the PR curve) of Algorithm 1 in function of age of the samples. For the two most common classes of images patterns can be identified. The comparison E scene has a noticeable downward trend, as the difficulty of the problem in the complex scene increases. For the frame F scene, no apparent change is visible.}
	\label{fig:fromtime}
\end{figure}

Results of application of scenarios 2-5 to all images are presented in Tab.~\ref{tab:new_results} in the form of Area Under Curve (AUC) of Precision-Recall (PR) curves. Results of the \textit{Ideal MF} are high, especially for images from the first days. \emph{E} images form an exception, possibly due to higher complexity of the scene background. In addition, results of the proposed algorithm and the \textit{Inductive MF} tend to be higher than \textit{MF}$_{lib}$. The higher performance of the Algorithm~\ref{algorithm:tsmf} supports the thesis that the proposed second stage tends to improve the initial detection performance. A plot of performance in function of samples age is presented in Fig.~\ref{fig:fromtime}. The \emph{E} images have a visible downward trend, as the degradation of blood samples makes them more difficult to detect in the presence of distracting scene elements. The simple frame image \emph{F} has no noticeable dependency on time.

\ctable[
cap = Aggregated detector performance,
caption     =  Aggregated detection performance in experiments expressed as Area Under Curve (AUC) of the Precision-Recall (PR) curve.,
label   = tab:new_results,
star
]{lcccc}{	
}{     \FL
	Code&\textit{Ideal MF}&\textit{Inductive MF}&\emph{MF}$_{lib}$&Algorithm~\ref{algorithm:tsmf}\ML
	\textit{F(1)}&1.00&0.41&0.39&0.47\NN
	\textit{F(1s)}&1.00&0.92&0.46&0.73\NN
	\textit{F(1a)}&1.00&0.17&0.24&0.41\NN
	\textit{F(2)}&0.99&0.61&0.37&0.72\NN
	\textit{F(2k)}&1.00&0.33&0.41&0.33\NN
	\textit{F(7)}&1.00&0.98&0.50&0.83\NN
	\textit{F(21)}&1.00&0.98&0.37&0.59\NN
	\textit{D(1)}&0.98&0.36&0.28&0.38\NN
	\textit{A(1)}&0.98&0.61&0.30&0.24\NN
	\textit{B(1)}&0.95&0.31&0.24&0.33\NN
	\textit{C(1)}&0.96&0.33&0.34&0.34\NN
	\textit{E(1)}&0.73&0.46&0.46&0.71\NN
	\textit{E(7)}&0.62&0.26&0.32&0.51\NN
	\textit{\textit{E(21)}}&0.59&0.33&0.29&0.42\LL
}

\begin{figure*}
	\centering
	\begin{subfigure}[b]{1.0\textwidth}
		\includegraphics[width=0.32\linewidth]{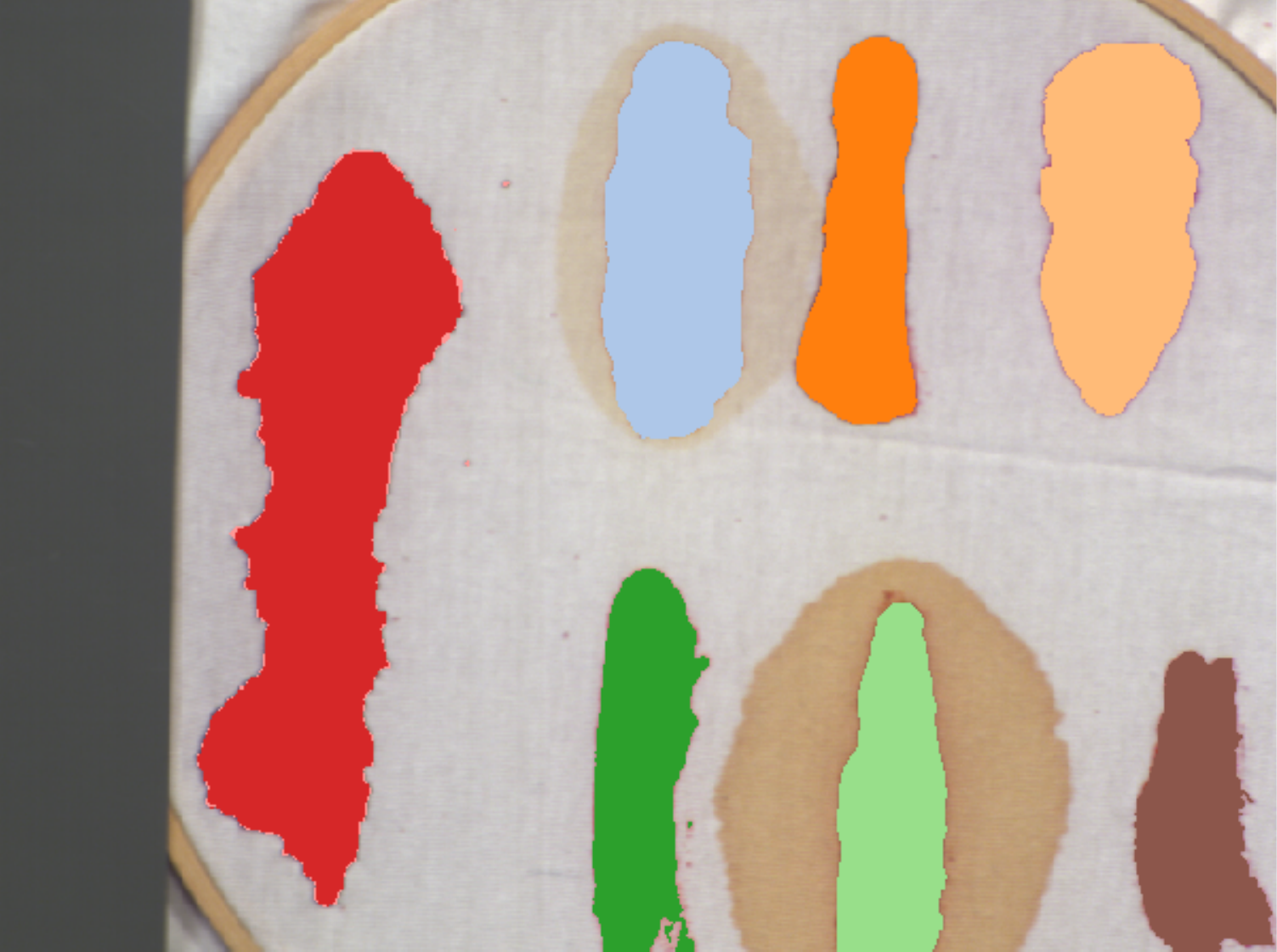}
		\includegraphics[width=0.32\linewidth]{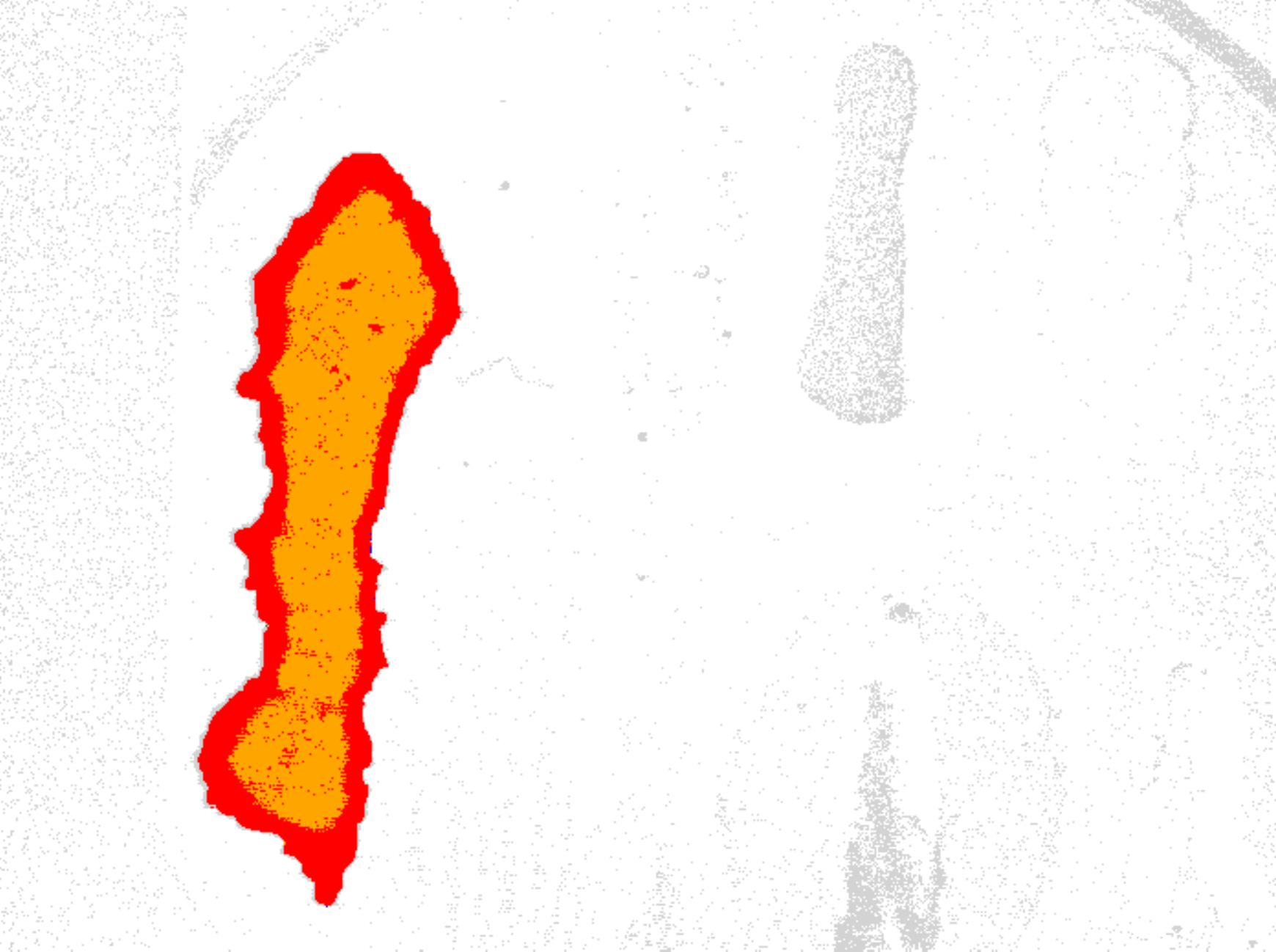}
		\includegraphics[width=0.32\linewidth]{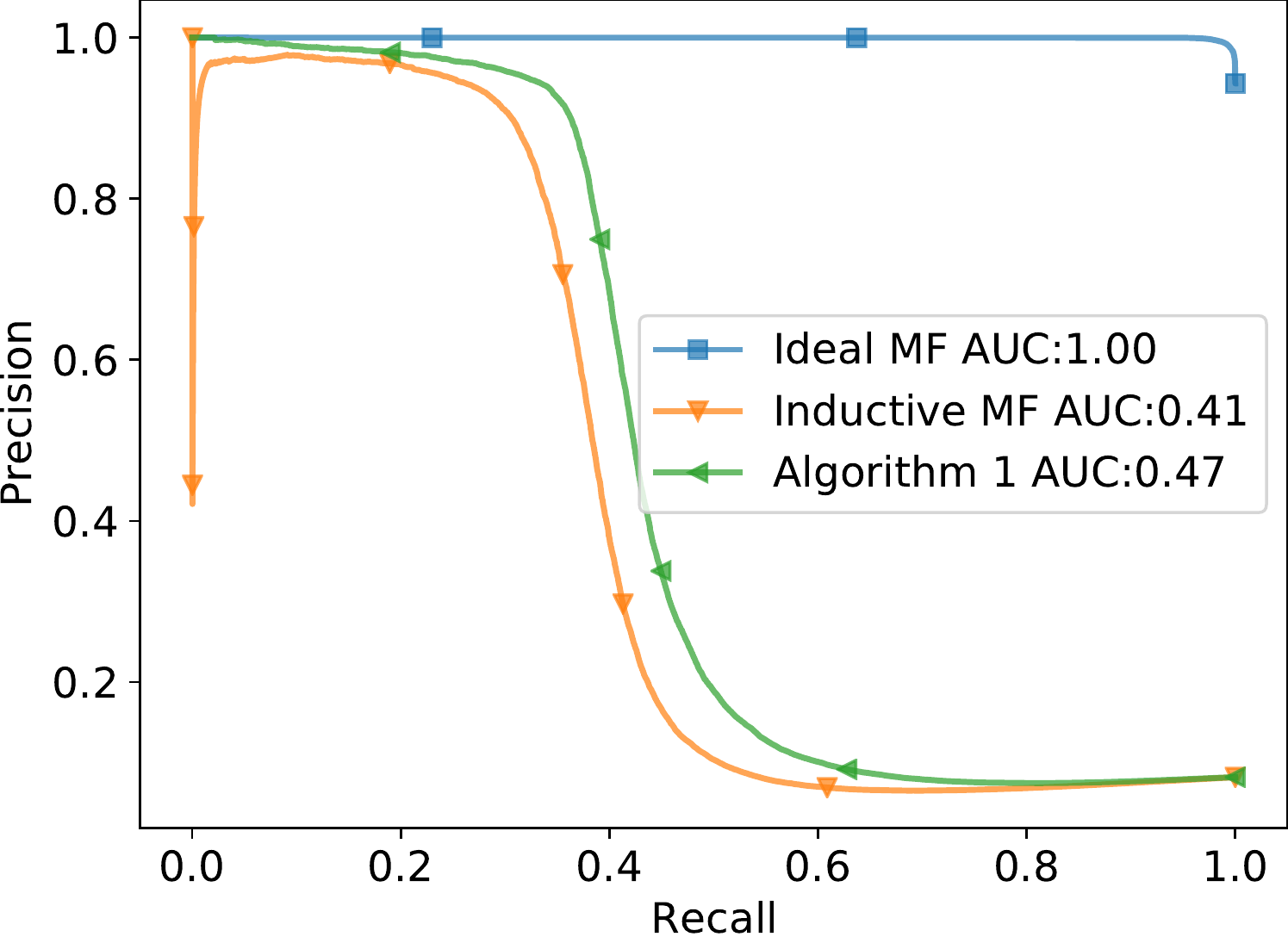}
		\caption{Image \textit{F(1)}}
		\label{fig:res_all_best:f1}
	\end{subfigure}
	\begin{subfigure}[b]{1.0\textwidth}
		\includegraphics[width=0.32\linewidth]{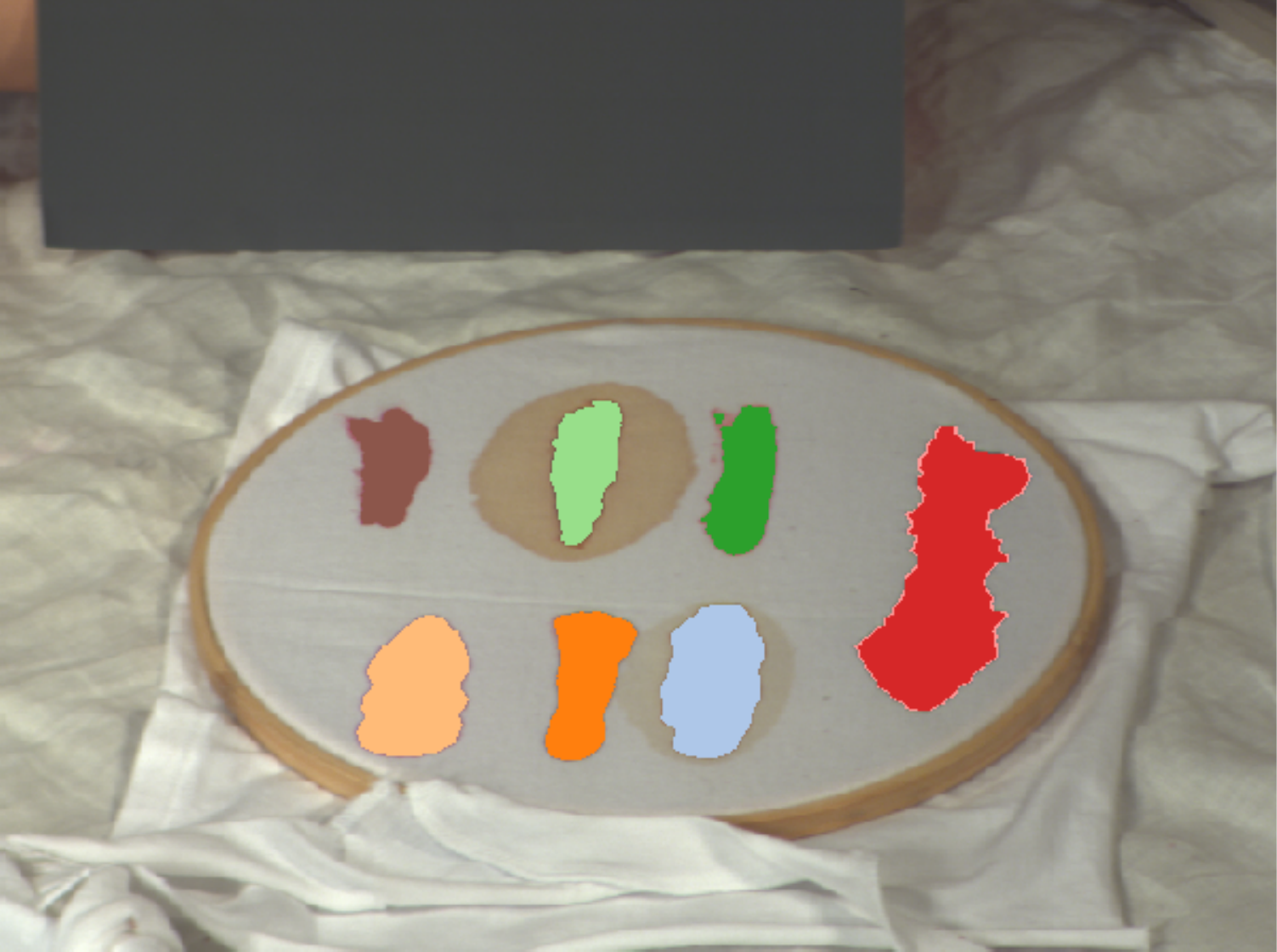}
		\includegraphics[width=0.32\linewidth]{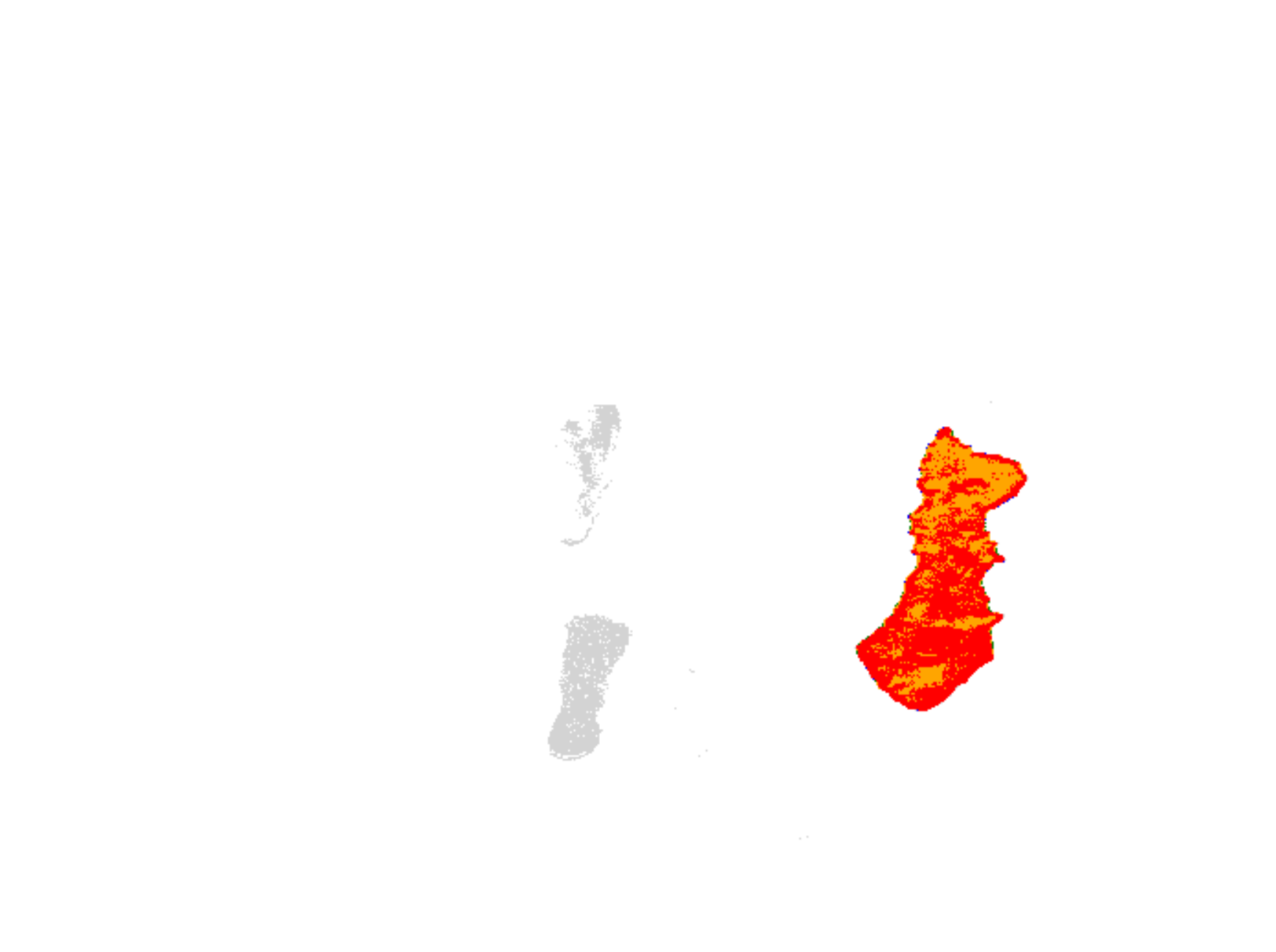}
		\includegraphics[width=0.32\linewidth]{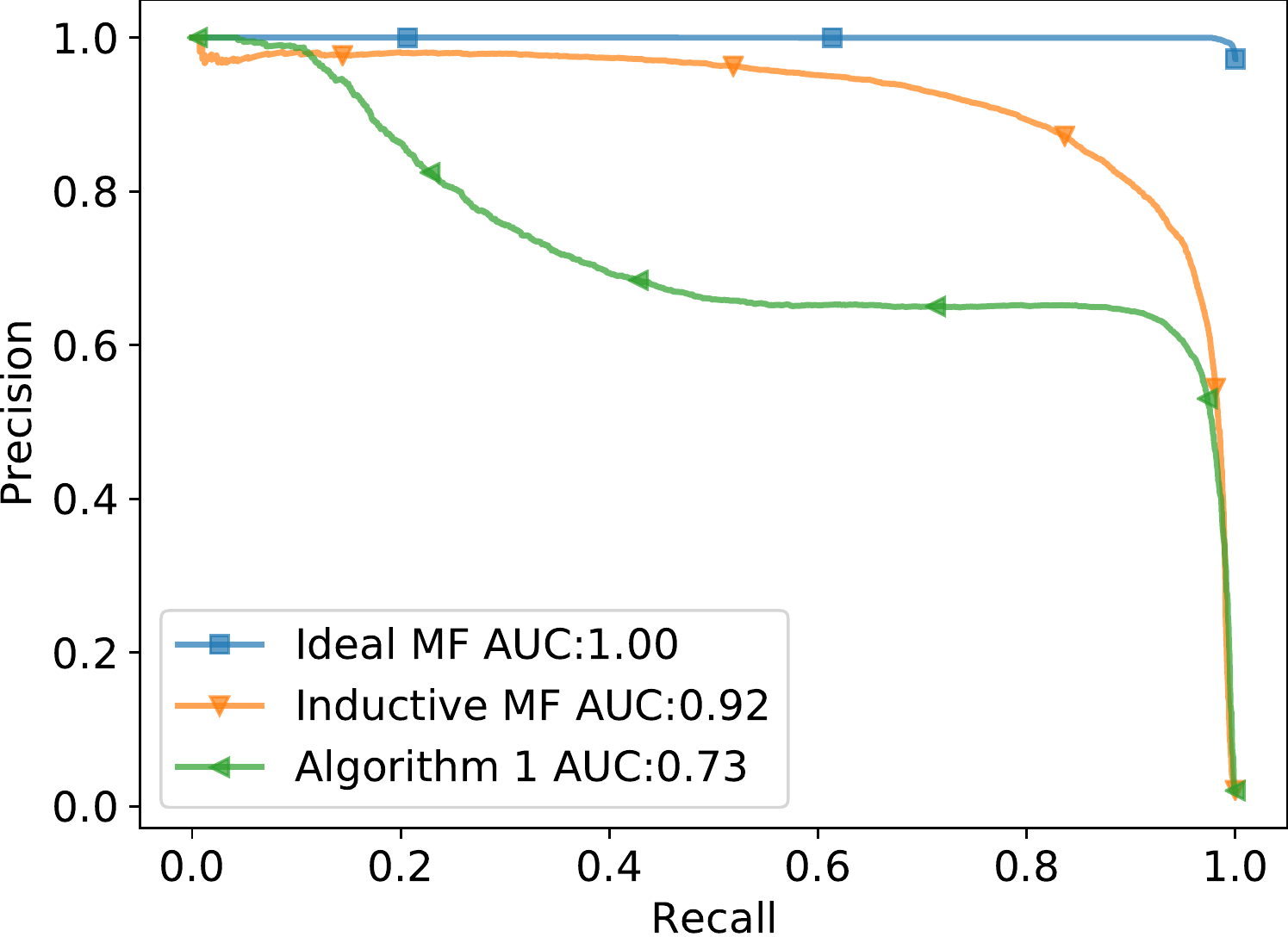}
		\caption{Image \textit{F(1s)}}
	\end{subfigure}
	\begin{subfigure}[b]{1.0\textwidth}
		\includegraphics[width=0.32\linewidth]{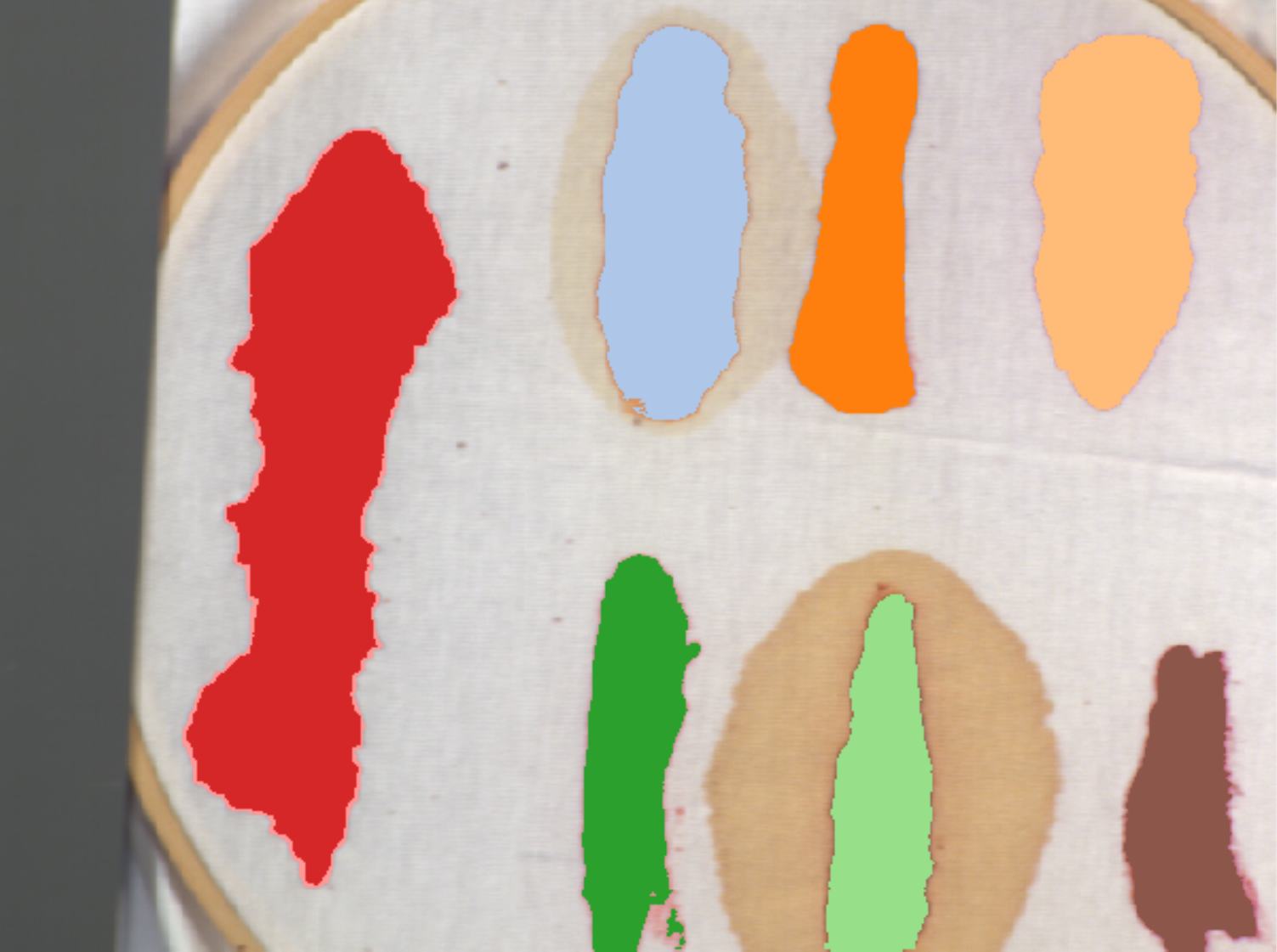}
		\includegraphics[width=0.32\linewidth]{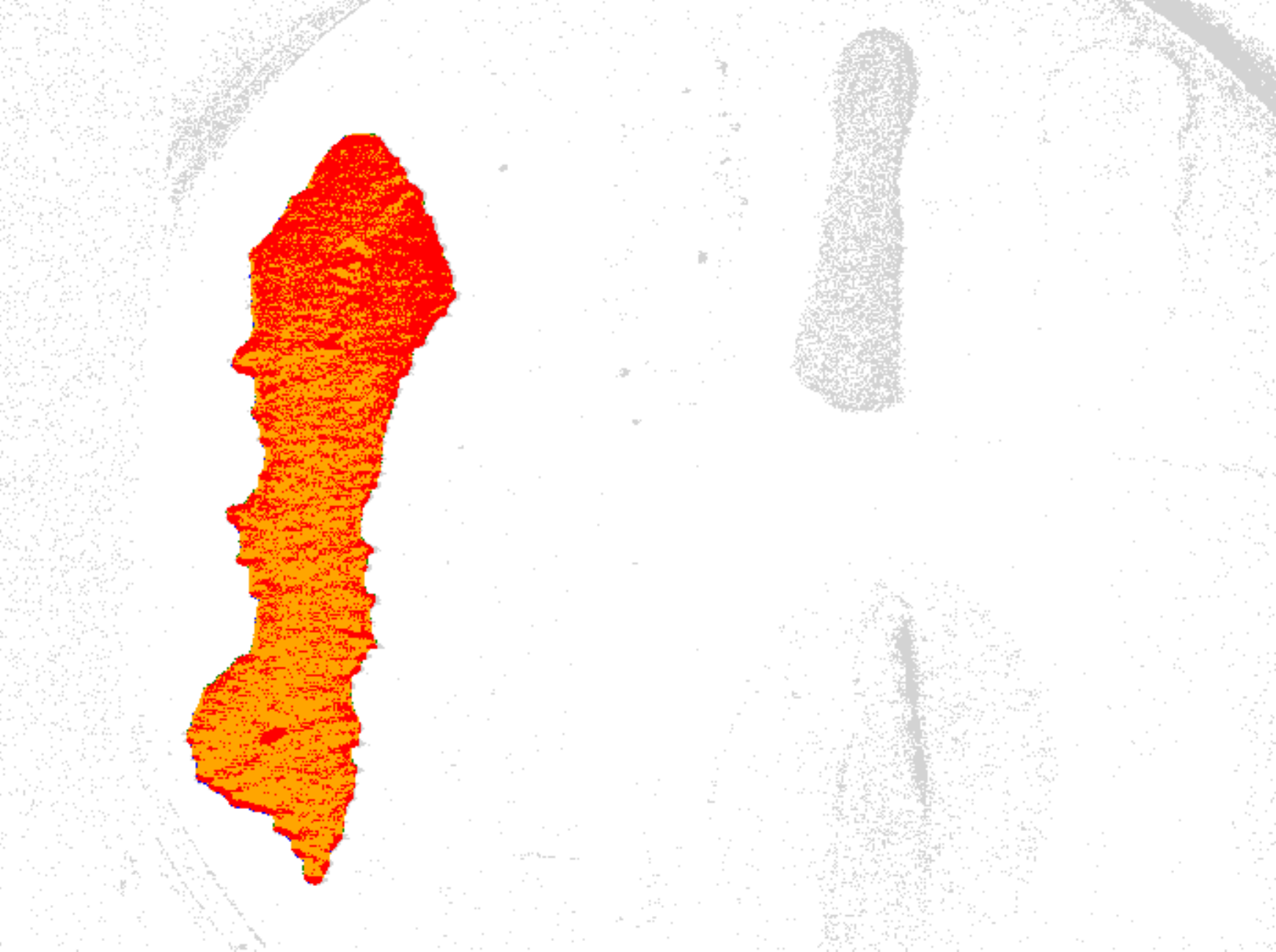}
		\includegraphics[width=0.32\linewidth]{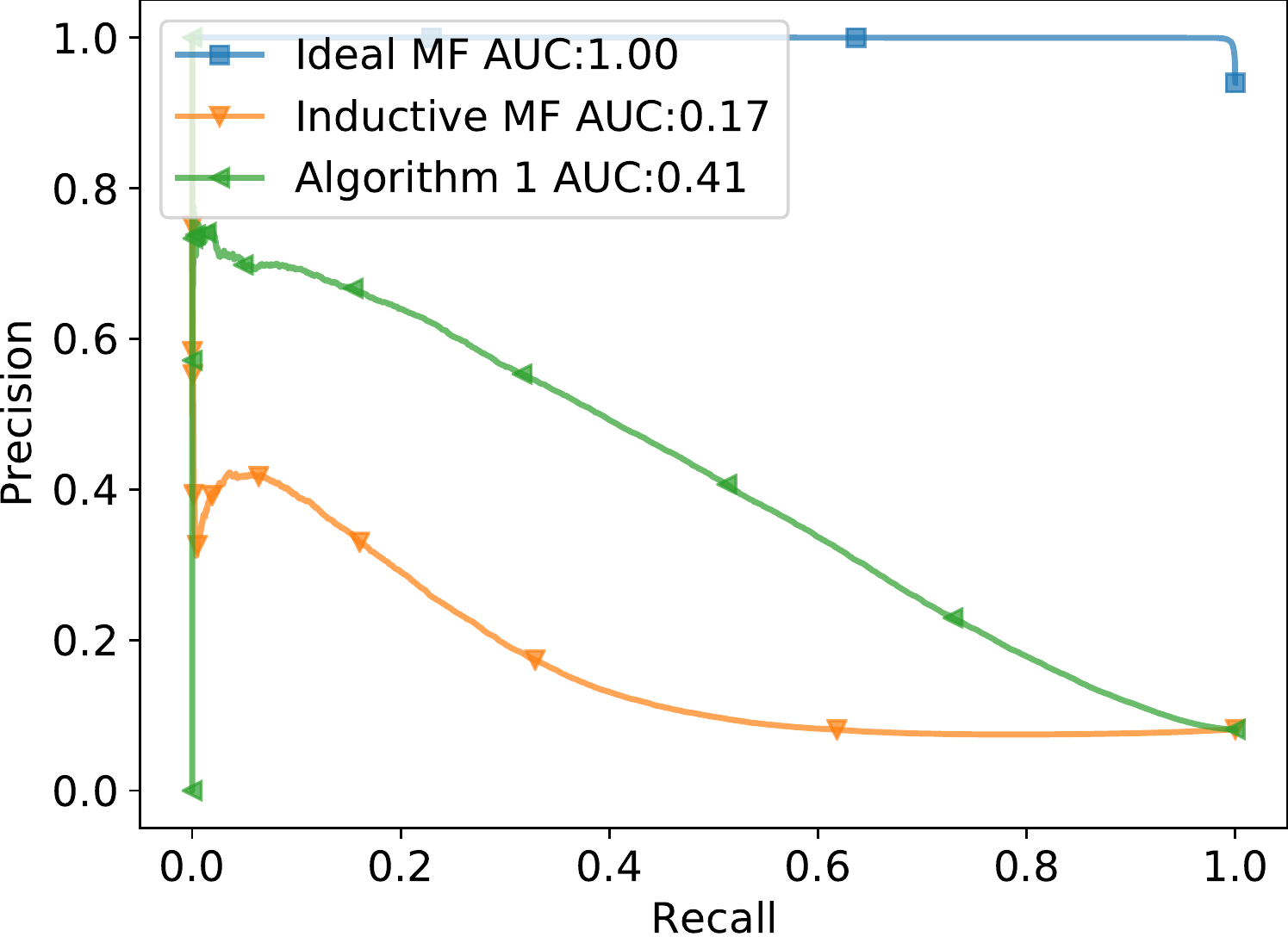}
		\caption{Image \textit{F(1a)}}
	\end{subfigure}
	\begin{subfigure}[b]{1.0\textwidth}
		\includegraphics[width=0.32\linewidth]{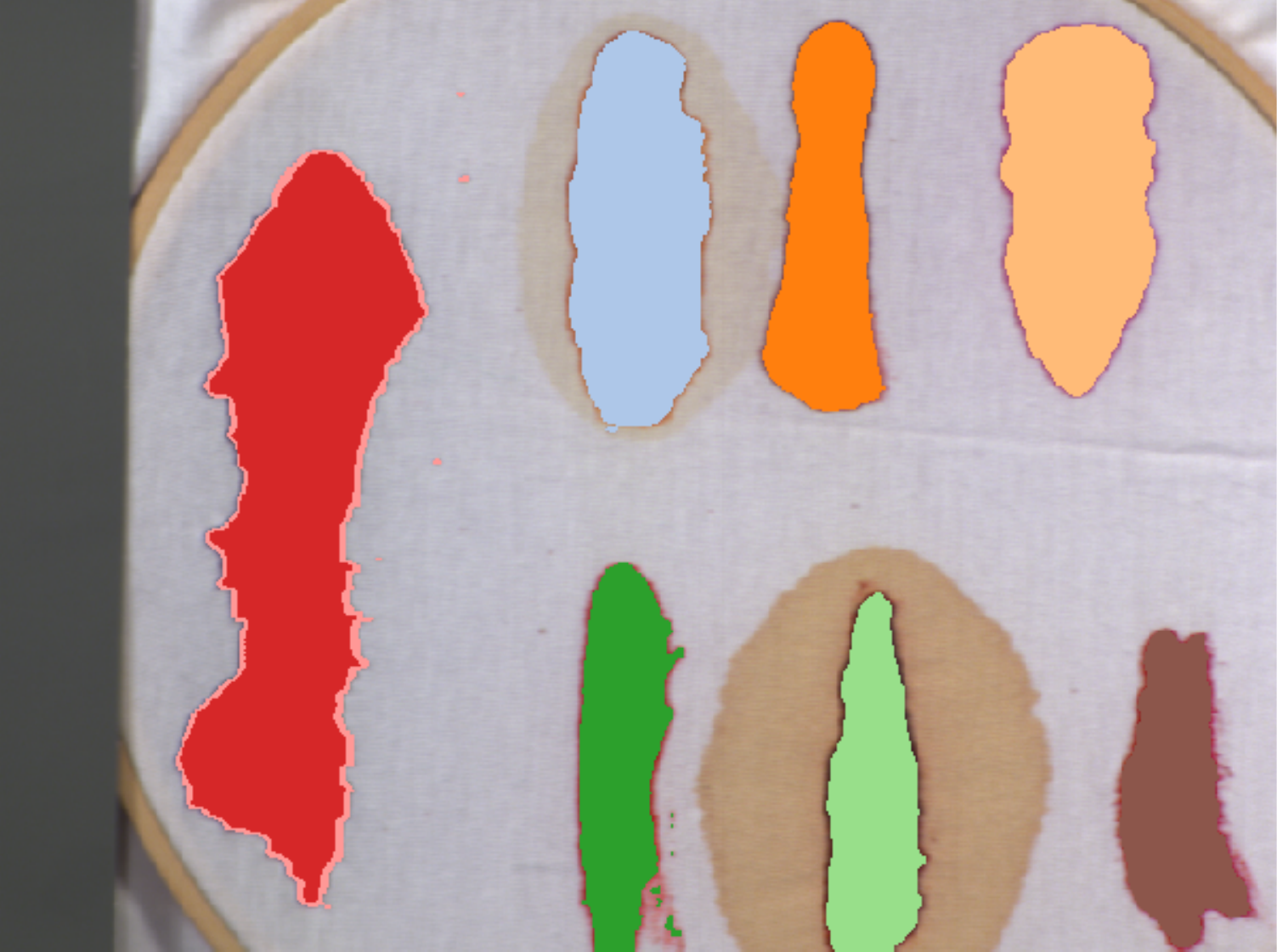}
		\includegraphics[width=0.32\linewidth]{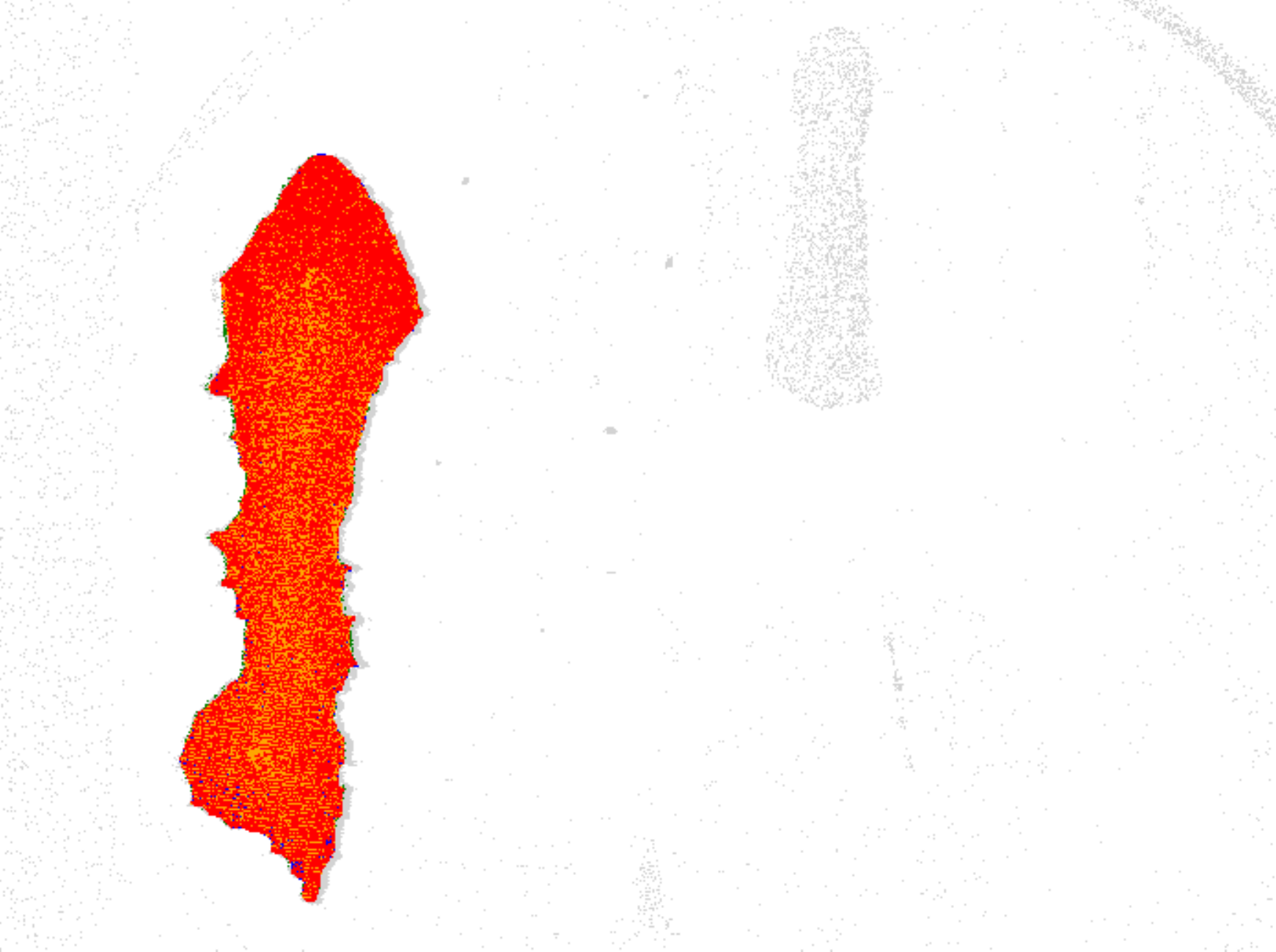}
		\includegraphics[width=0.32\linewidth]{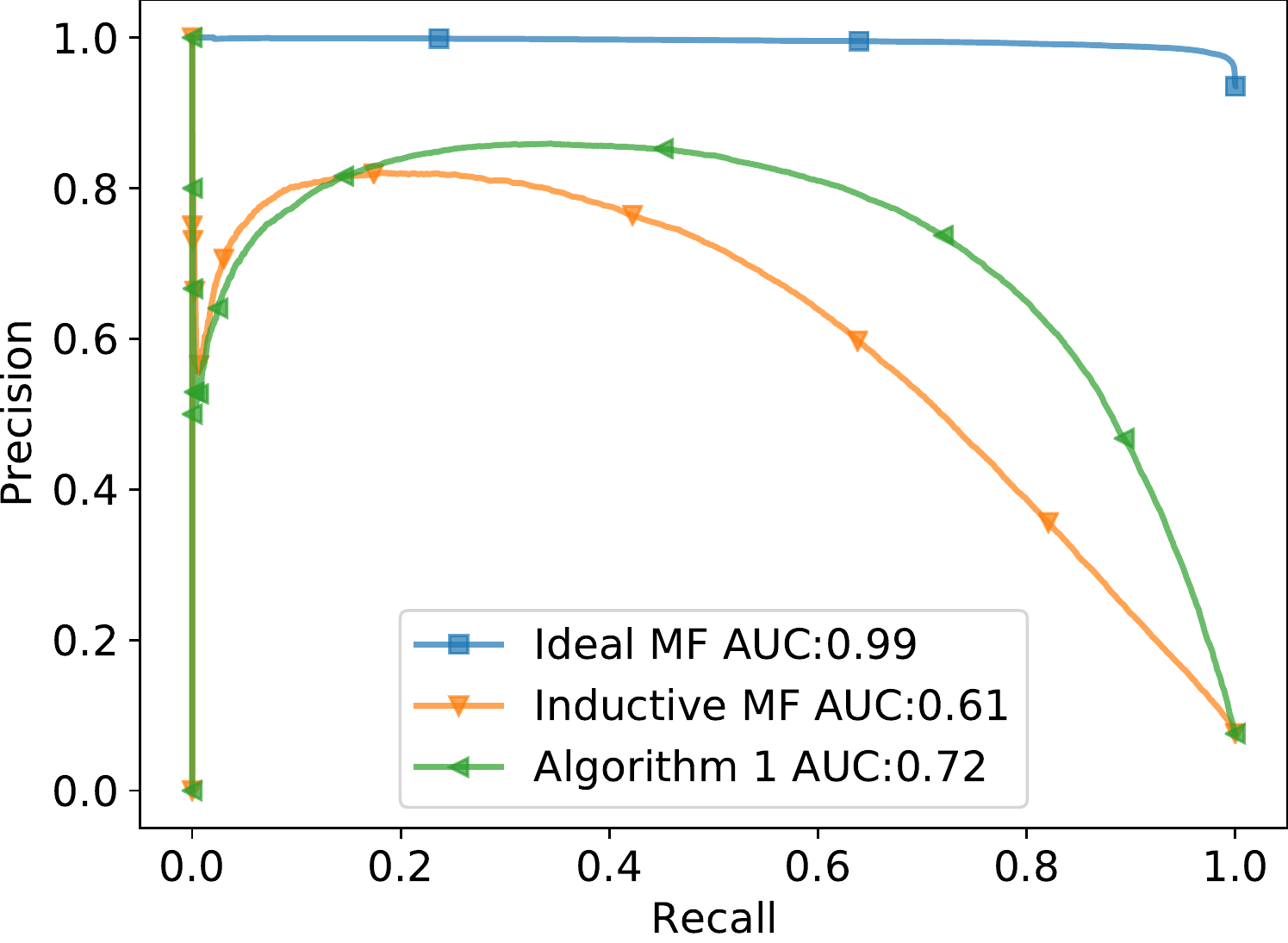}
		\caption{Image \textit{F(2)}}
	\end{subfigure}
	\begin{subfigure}[b]{1.0\textwidth}
		\includegraphics[width=0.32\linewidth]{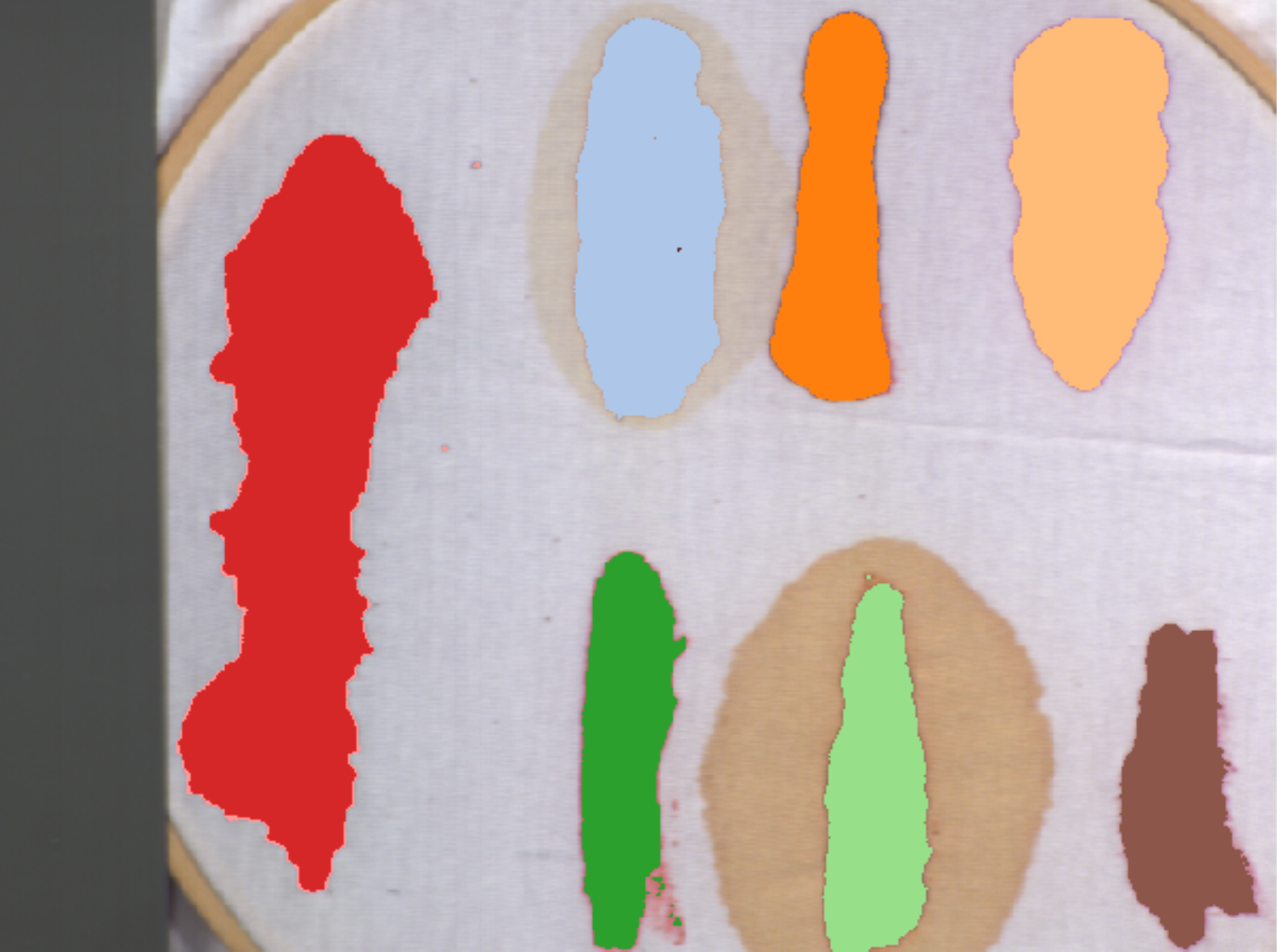}
		\includegraphics[width=0.32\linewidth]{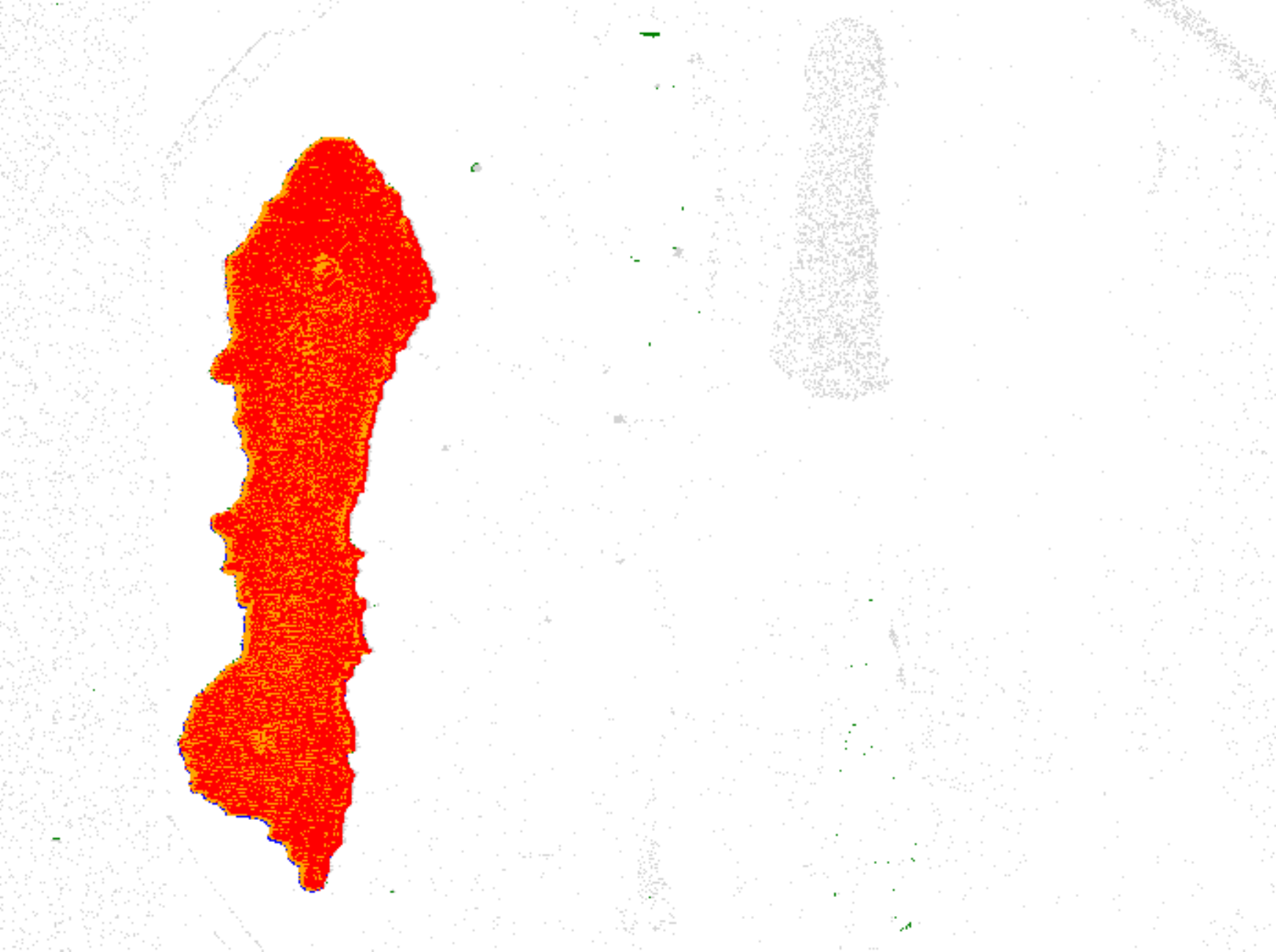}
		\includegraphics[width=0.32\linewidth]{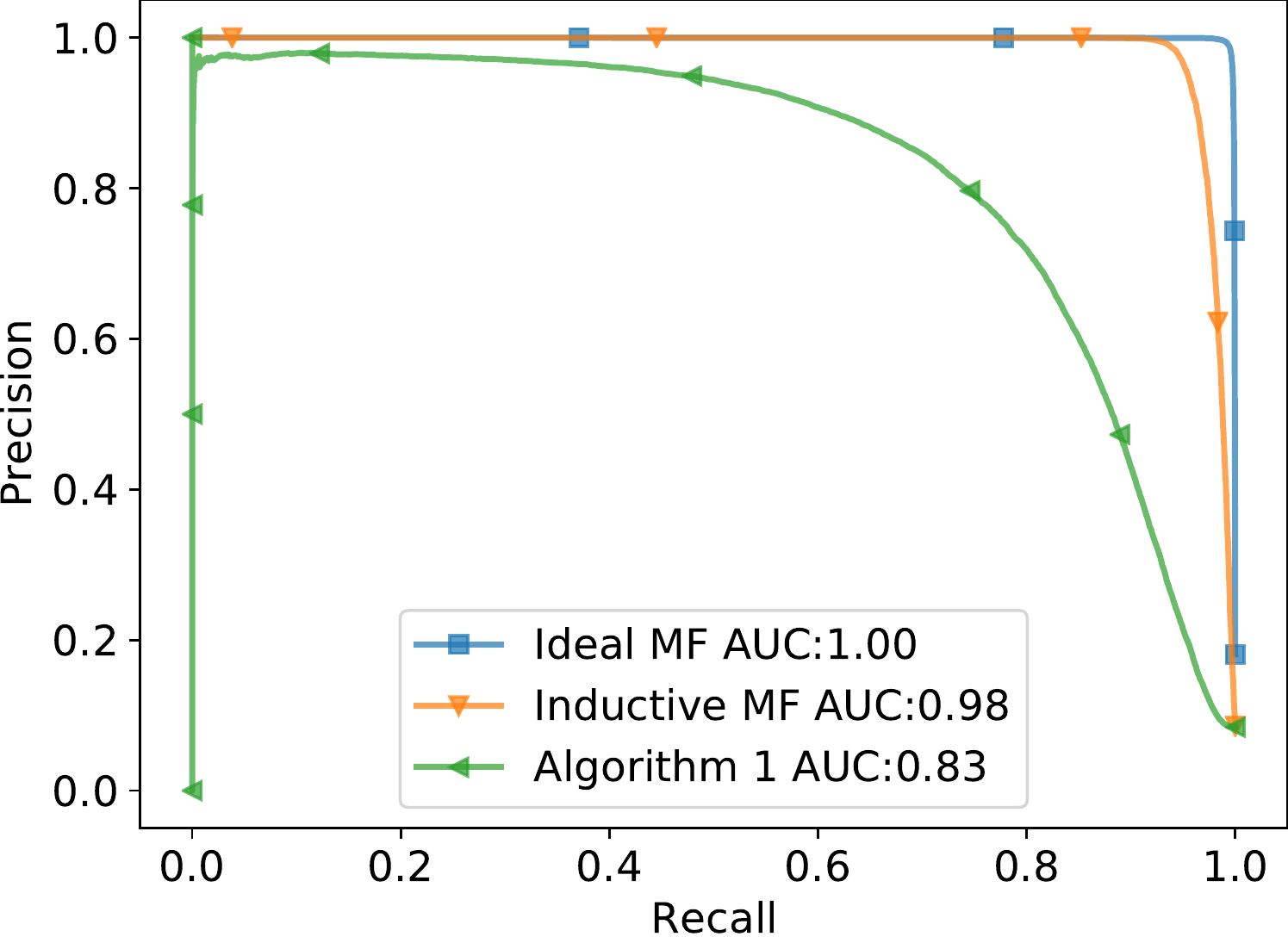}
		\caption{Image \textit{F(7)}}
	\end{subfigure}
	\caption{Performance of the Algorithm~\ref{algorithm:tsmf} for dataset images. Left column presents the ground truth--blood is denoted with red colour. Middle column presents the coloured example output of the detector--correct detections (TP) are coloured red, potential detections (FN of the Algorithm~\ref{algorithm:tsmf} but TP of the \textit{Ideal MF}) are coloured orange, incorrect detections (FP of the Algorithm~\ref{algorithm:tsmf}) are coloured grey, incorrect detections (FP) of the  \textit{Ideal MF} not detected (TN) by the Algorithm~\ref{algorithm:tsmf} are coloured green, detections missed (FN) by both algorithms are coloured blue. The last column presents PR curves for the Algorithm~\ref{algorithm:tsmf}, the \textit{Inductive MF} and the \textit{Ideal MF}.}
	\label{fig:res_all_best}
\end{figure*}

\begin{figure*}
	\centering
	\begin{subfigure}[b]{1.0\textwidth}
		\includegraphics[width=0.32\linewidth]{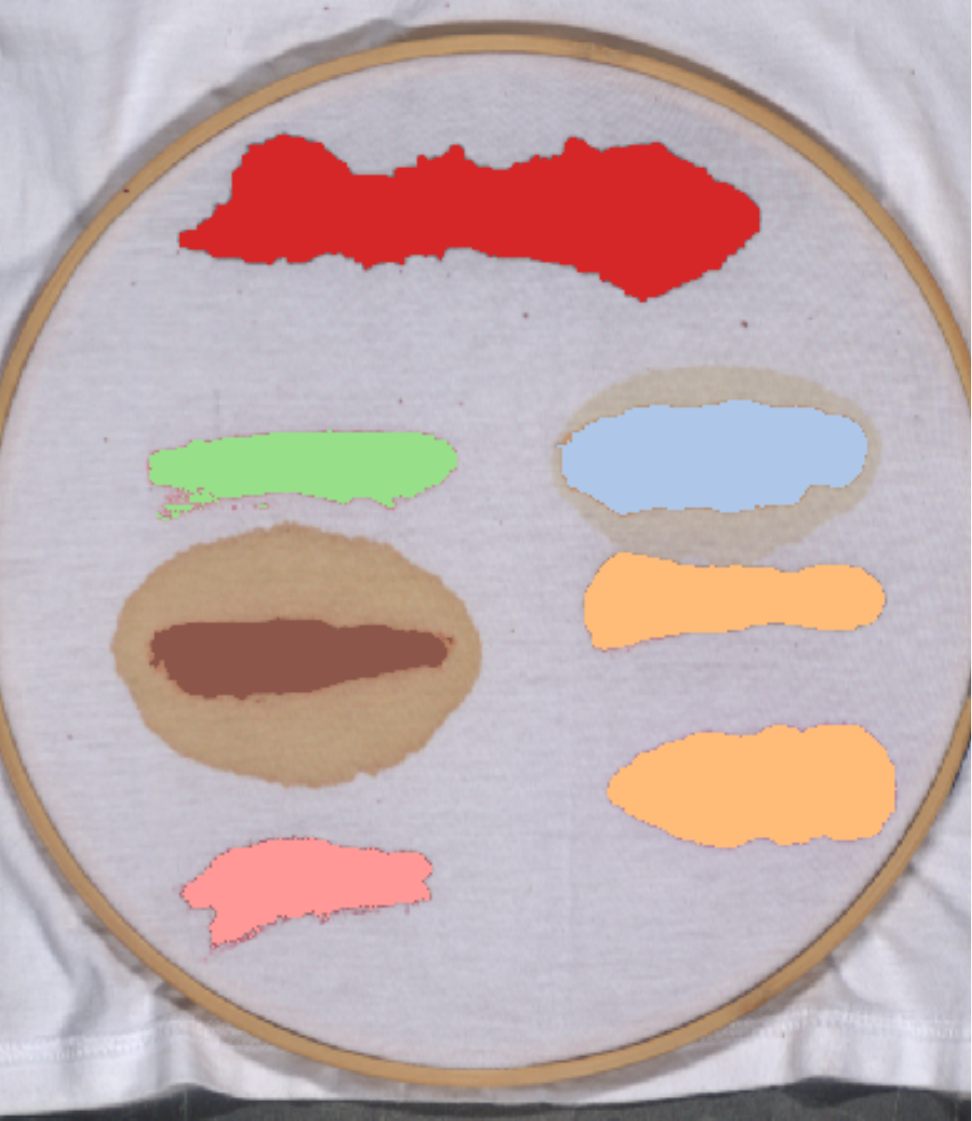}
		\includegraphics[width=0.32\linewidth]{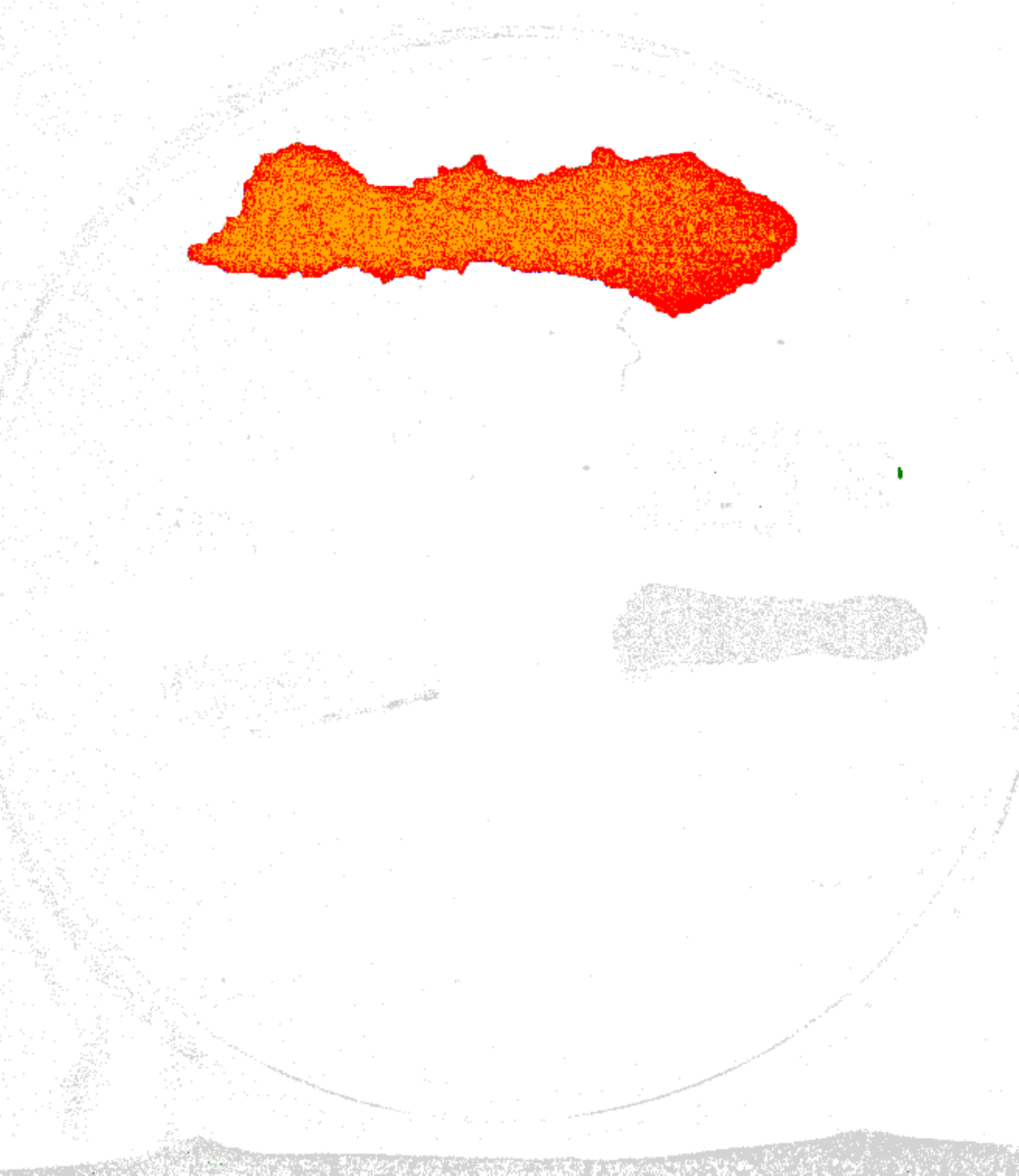}
		\includegraphics[width=0.32\linewidth]{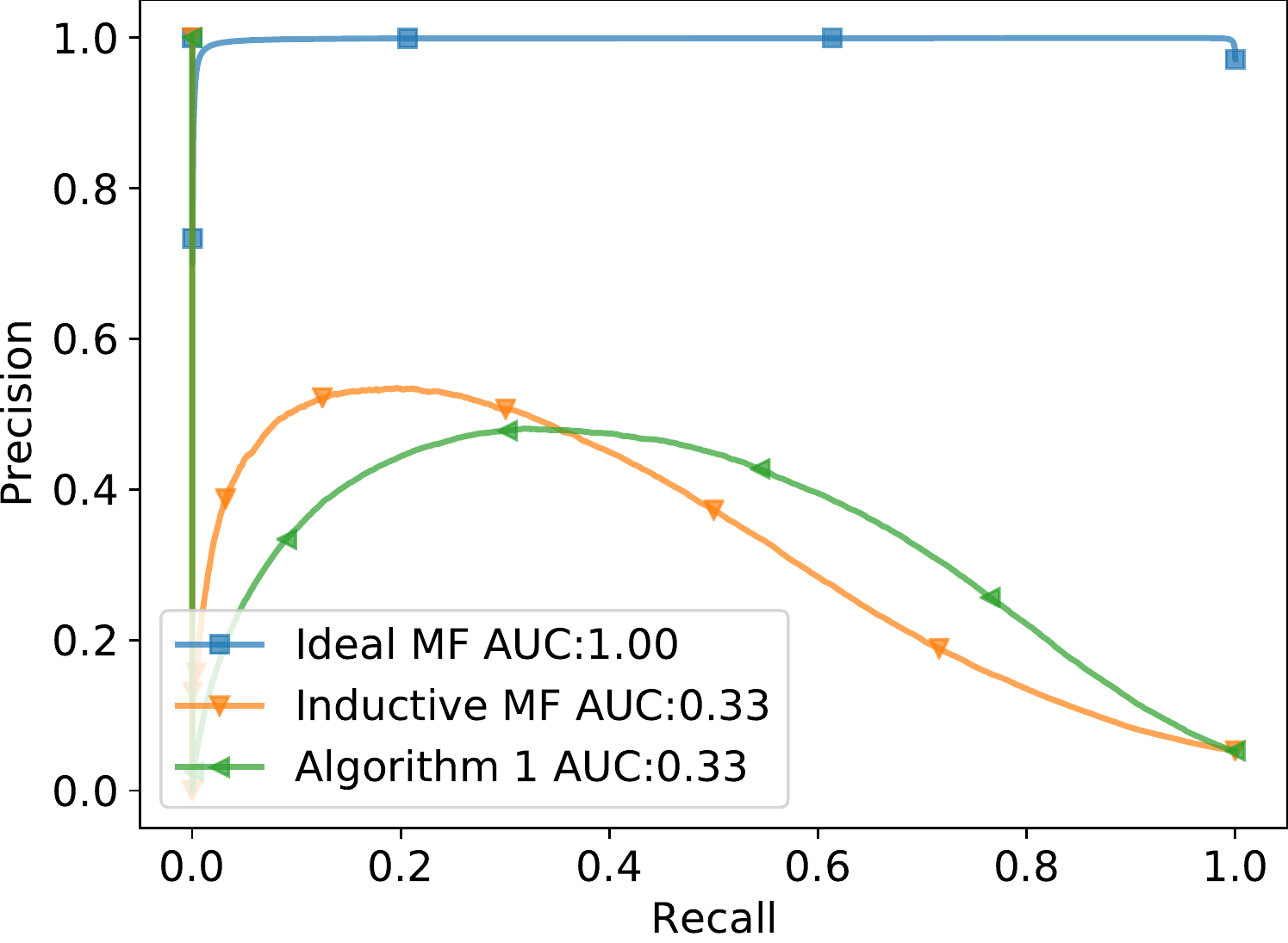}
		\caption{Image \textit{F(2k)}}
	\end{subfigure}
	\begin{subfigure}[b]{1.0\textwidth}
		\includegraphics[width=0.32\linewidth]{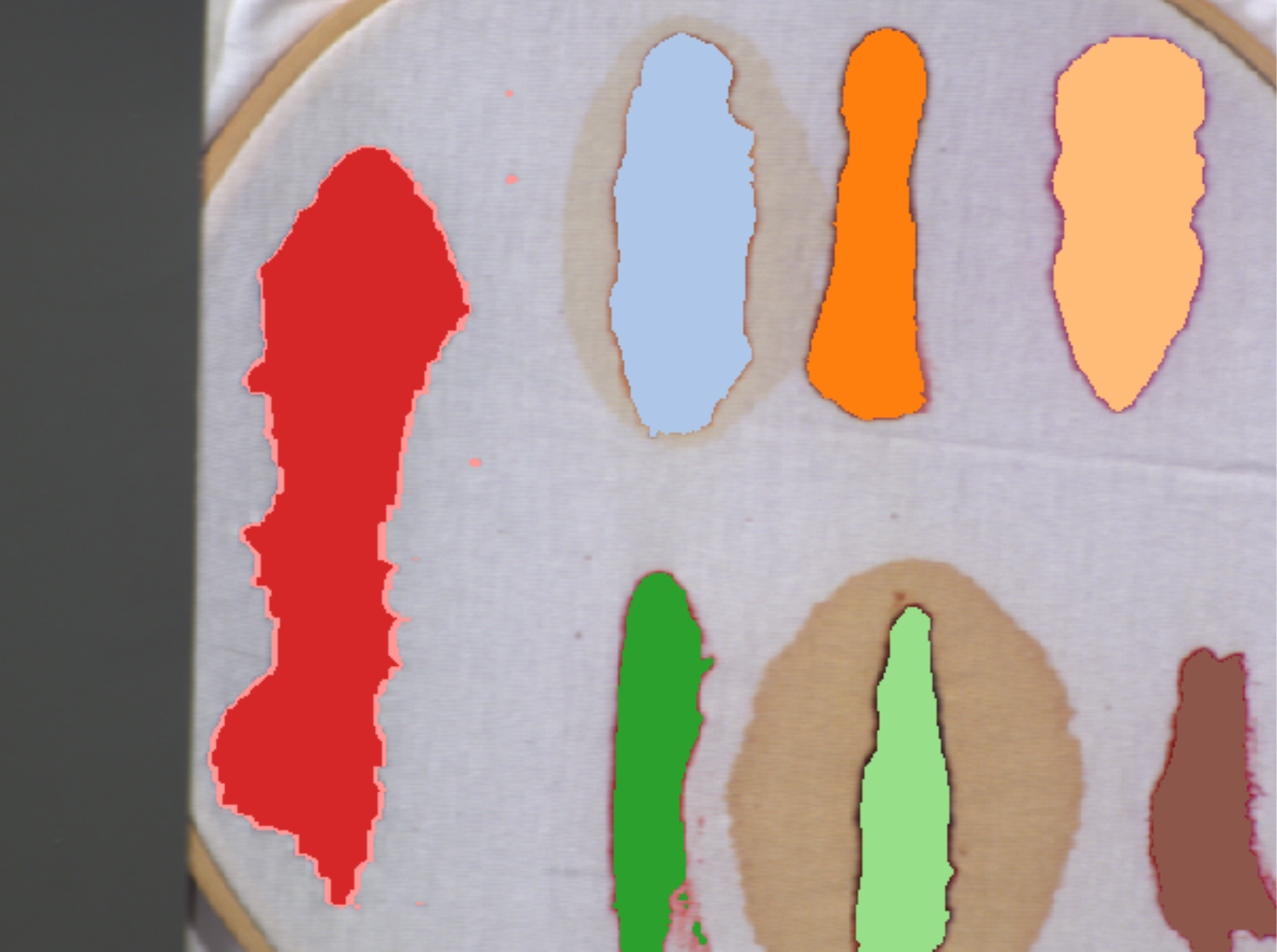}
		\includegraphics[width=0.32\linewidth]{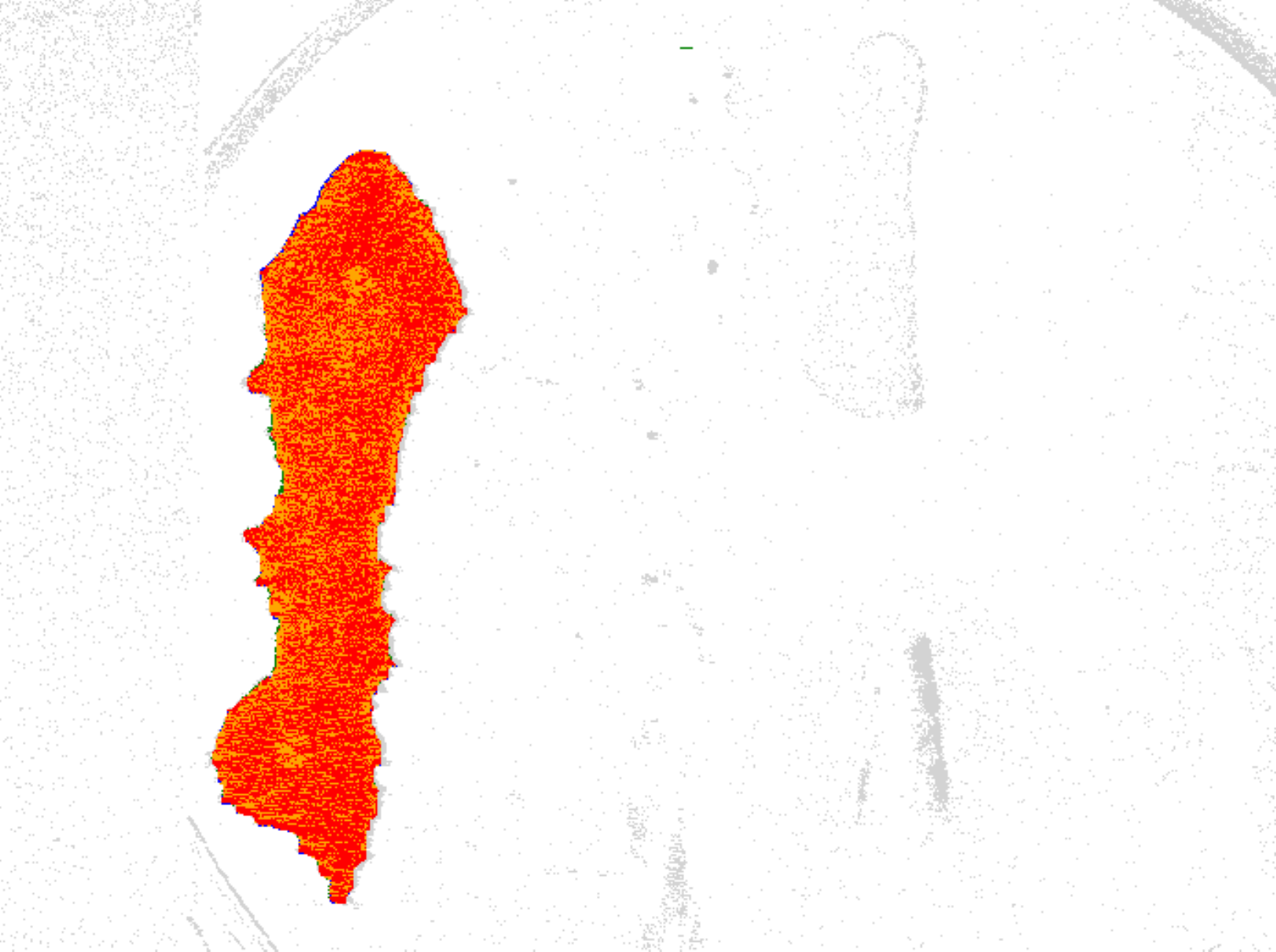}
		\includegraphics[width=0.32\linewidth]{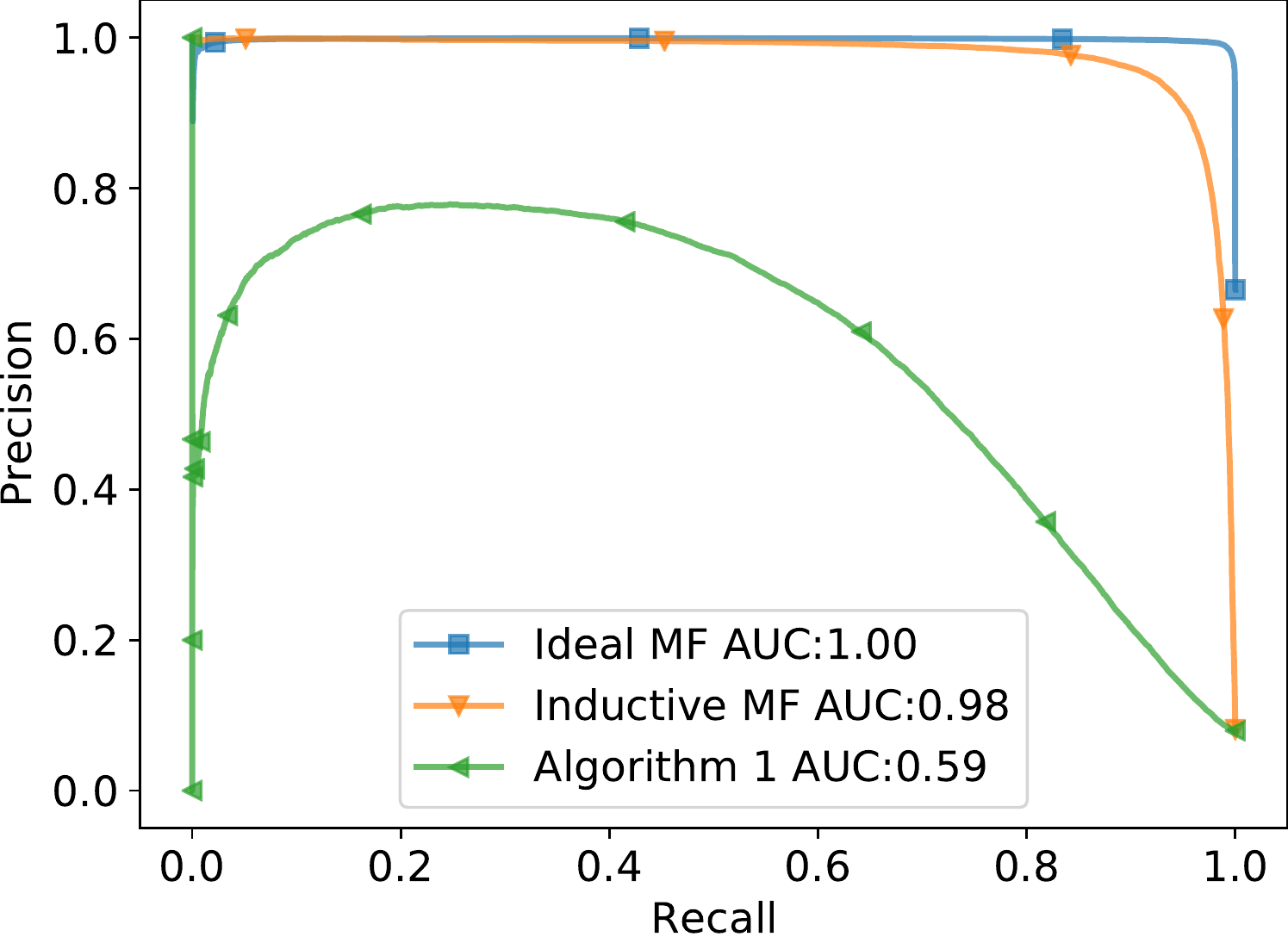}
		\caption{Image \textit{F(21)}}
	\end{subfigure}
	\begin{subfigure}[b]{1.0\textwidth}
		\includegraphics[width=0.32\linewidth]{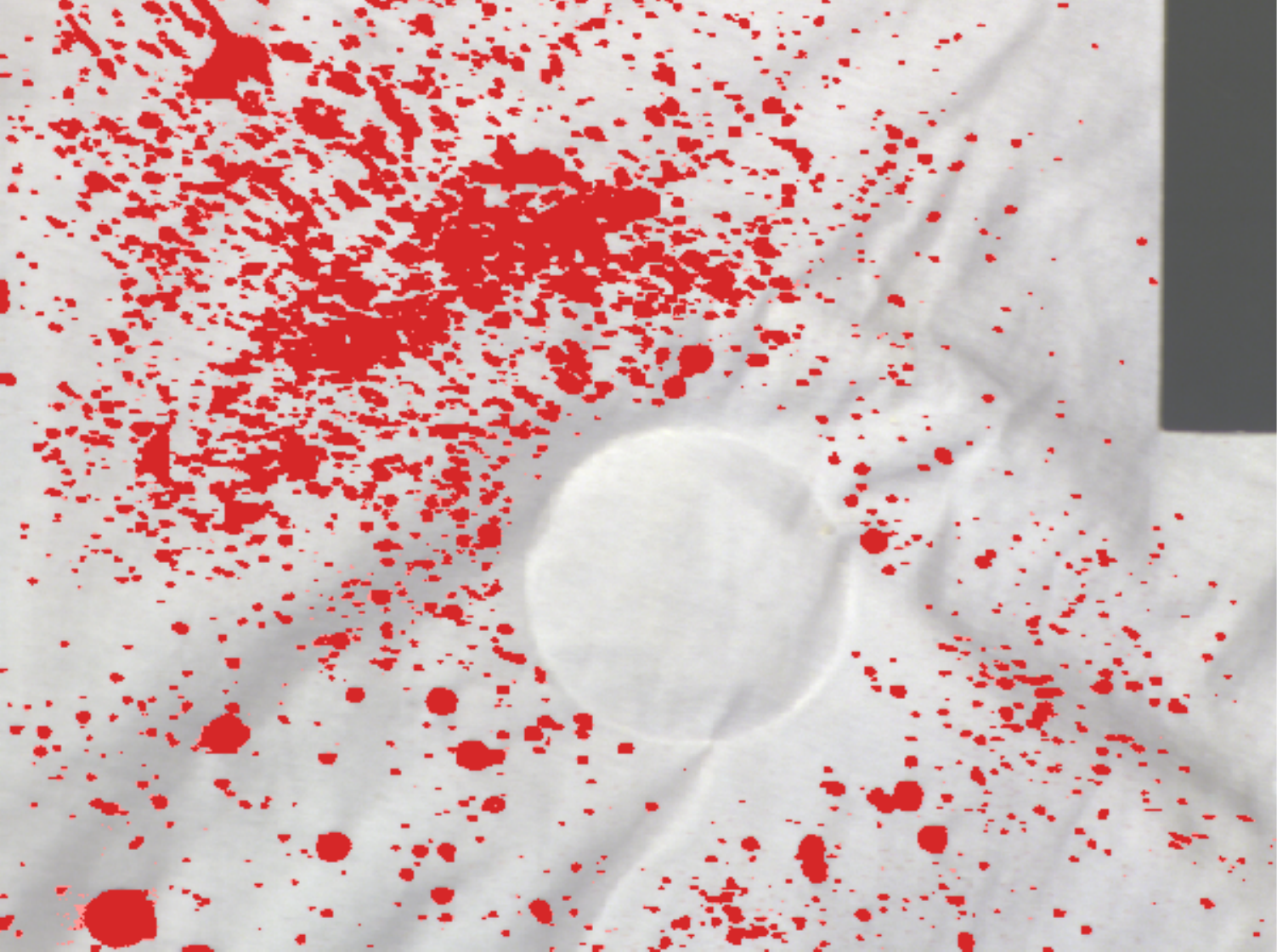}
		\includegraphics[width=0.32\linewidth]{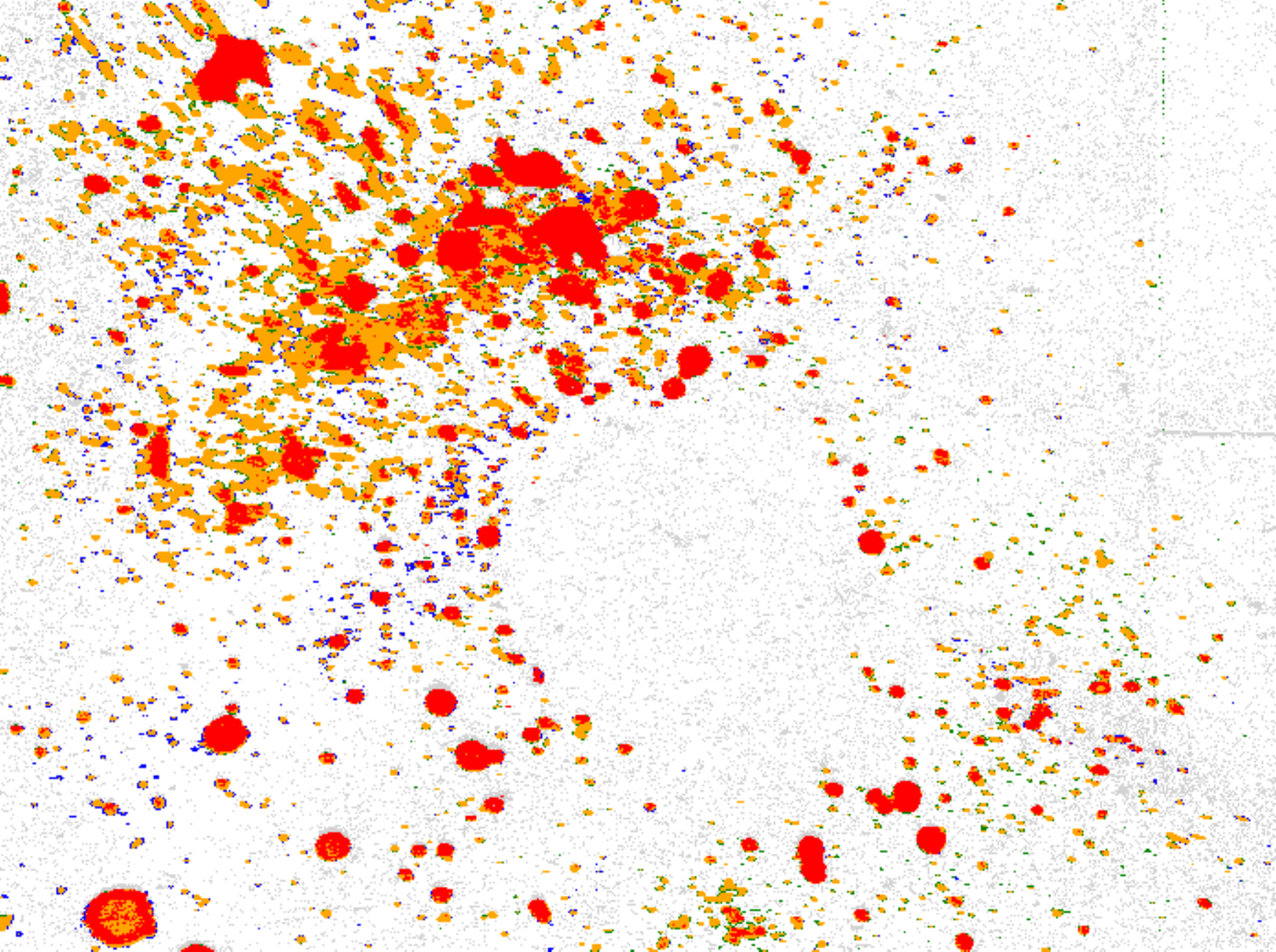}
		\includegraphics[width=0.32\linewidth]{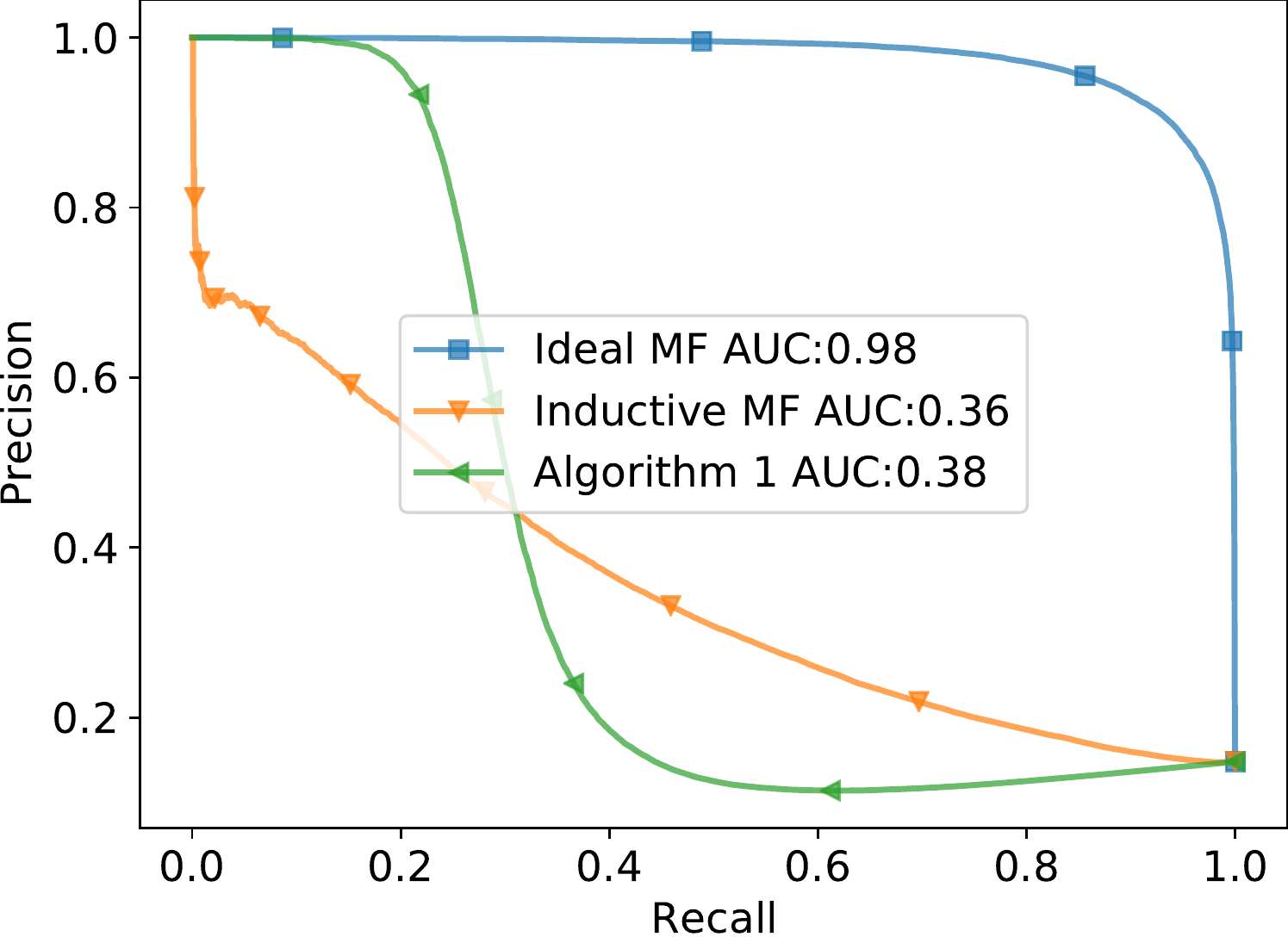}
		\caption{Image \textit{D(1)}}
	\end{subfigure}
	\begin{subfigure}[b]{1.0\textwidth}
		\includegraphics[width=0.32\linewidth]{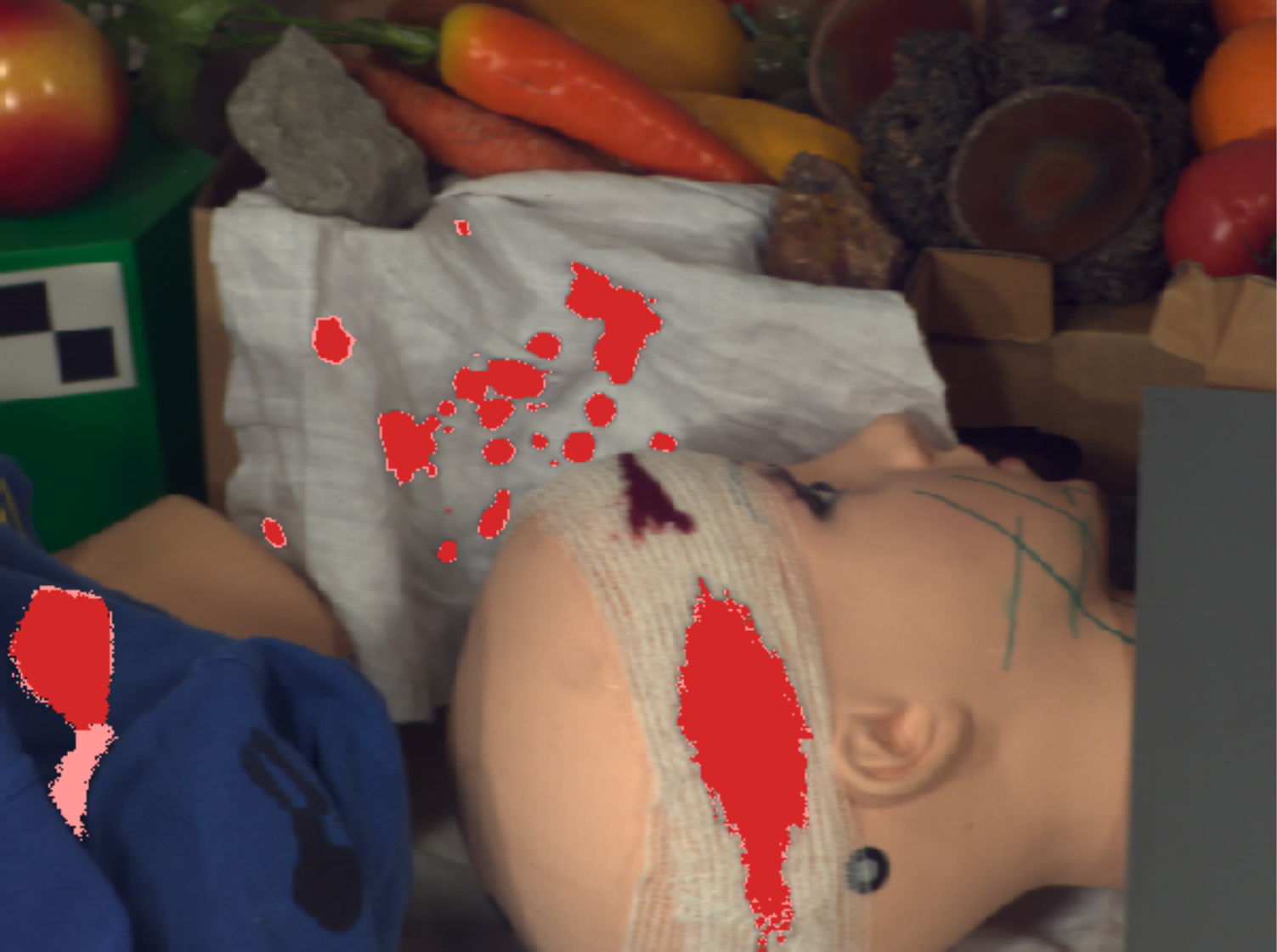}
		\includegraphics[width=0.32\linewidth]{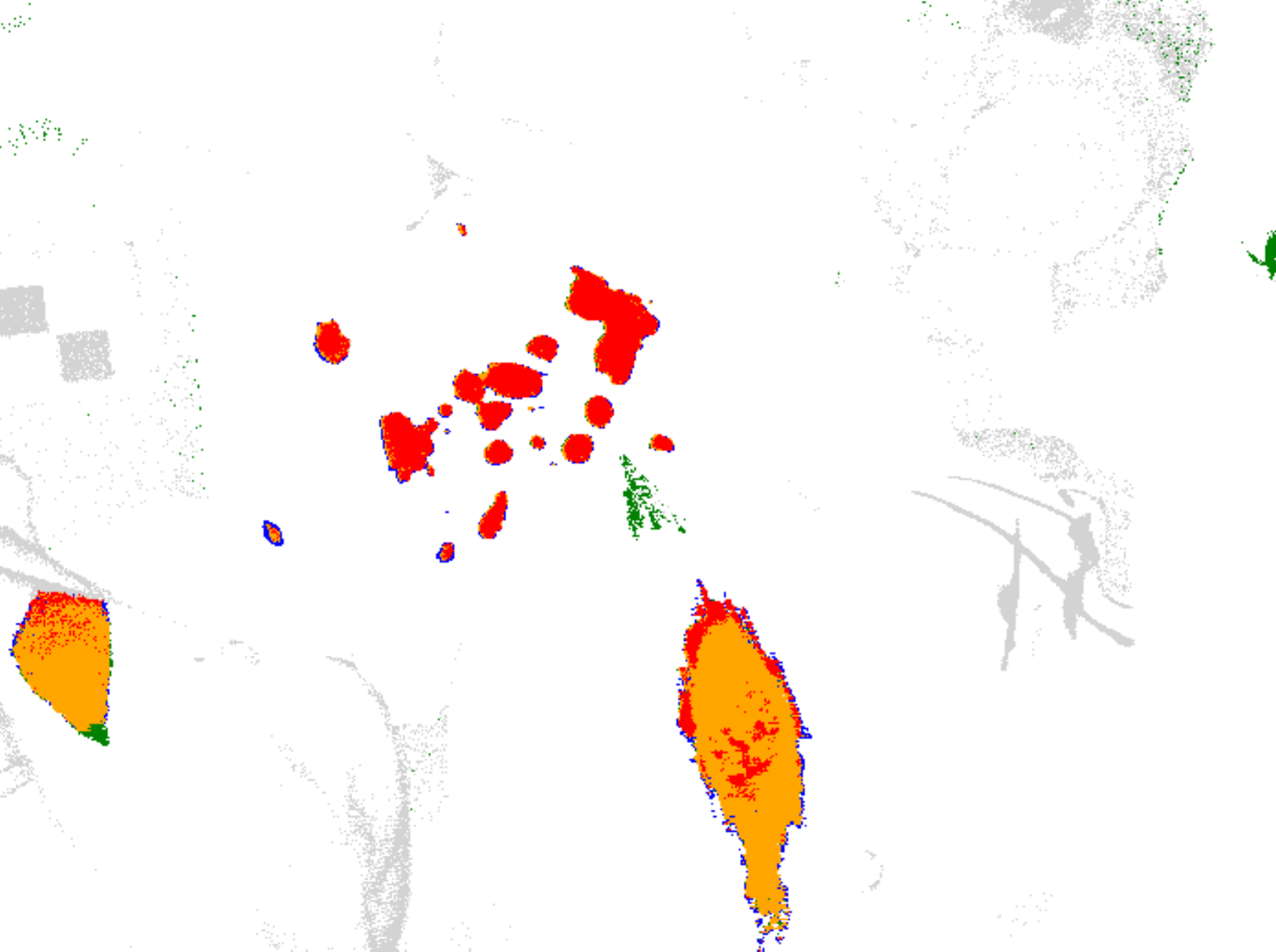}
		\includegraphics[width=0.32\linewidth]{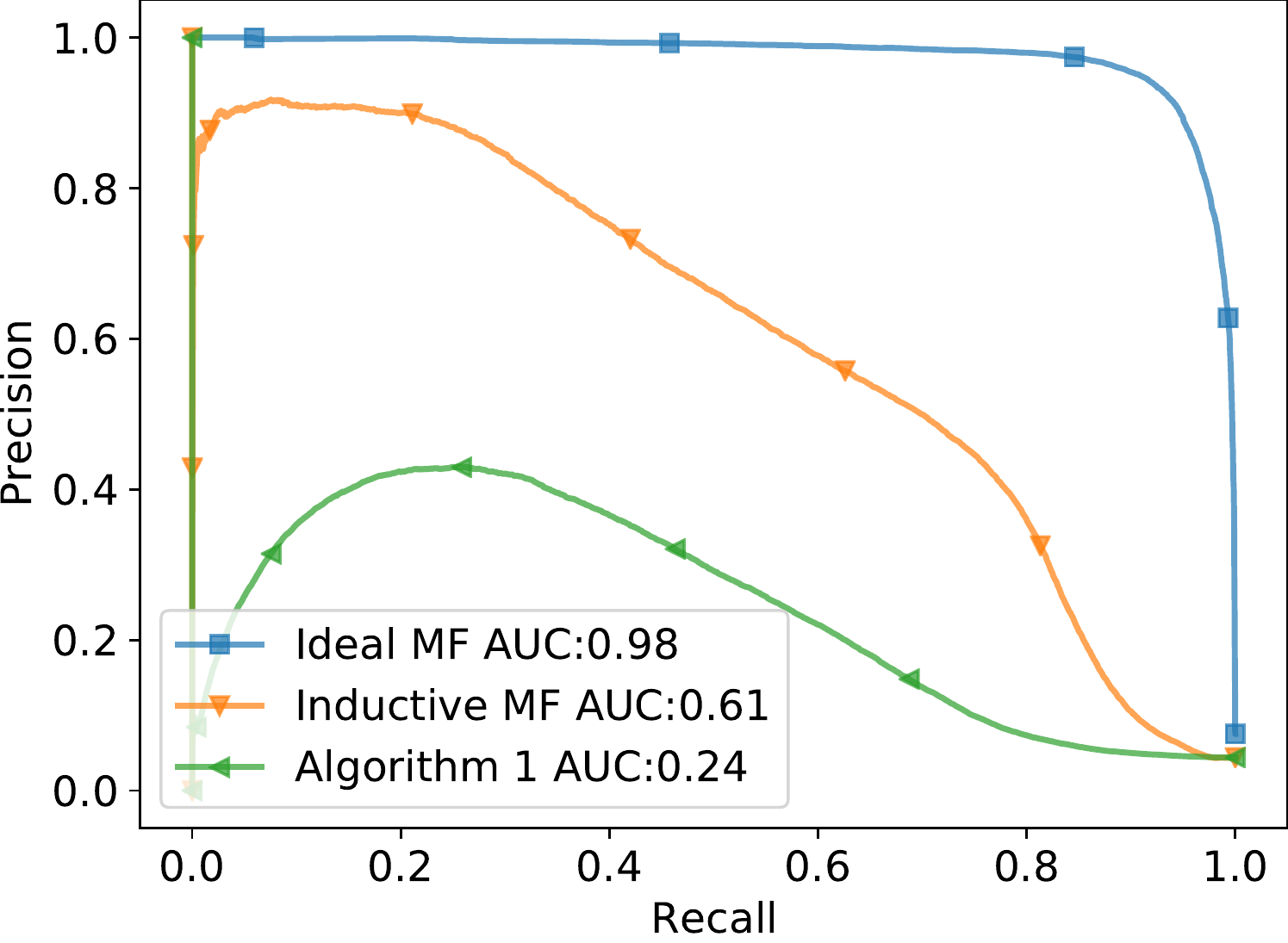}
		\caption{Image \textit{A(1)}}
		\label{fig:res_all_best2:a1}
	\end{subfigure}
	\caption{Performance of the Algorithm~\ref{algorithm:tsmf} for dataset images. Left column presents the ground truth--blood is denoted with red colour. Middle column presents the coloured example output of the detector--correct detections (TP) are coloured red, potential detections (FN of the Algorithm~\ref{algorithm:tsmf} but TP of the \textit{Ideal MF}) are coloured orange, incorrect detections (FP of the Algorithm~\ref{algorithm:tsmf}) are coloured grey, incorrect detections (FP) of the  \textit{Ideal MF} not detected (TN) by the Algorithm~\ref{algorithm:tsmf} are coloured green, detections missed (FN) by both algorithms are coloured blue. The last column presents PR curves for the Algorithm~\ref{algorithm:tsmf}, the \textit{Inductive MF} and the \textit{Ideal MF}.}
	\label{fig:res_all_best2}
\end{figure*}

\begin{figure*}
	\centering
	\begin{subfigure}[b]{1.0\textwidth}
		\includegraphics[width=0.32\linewidth]{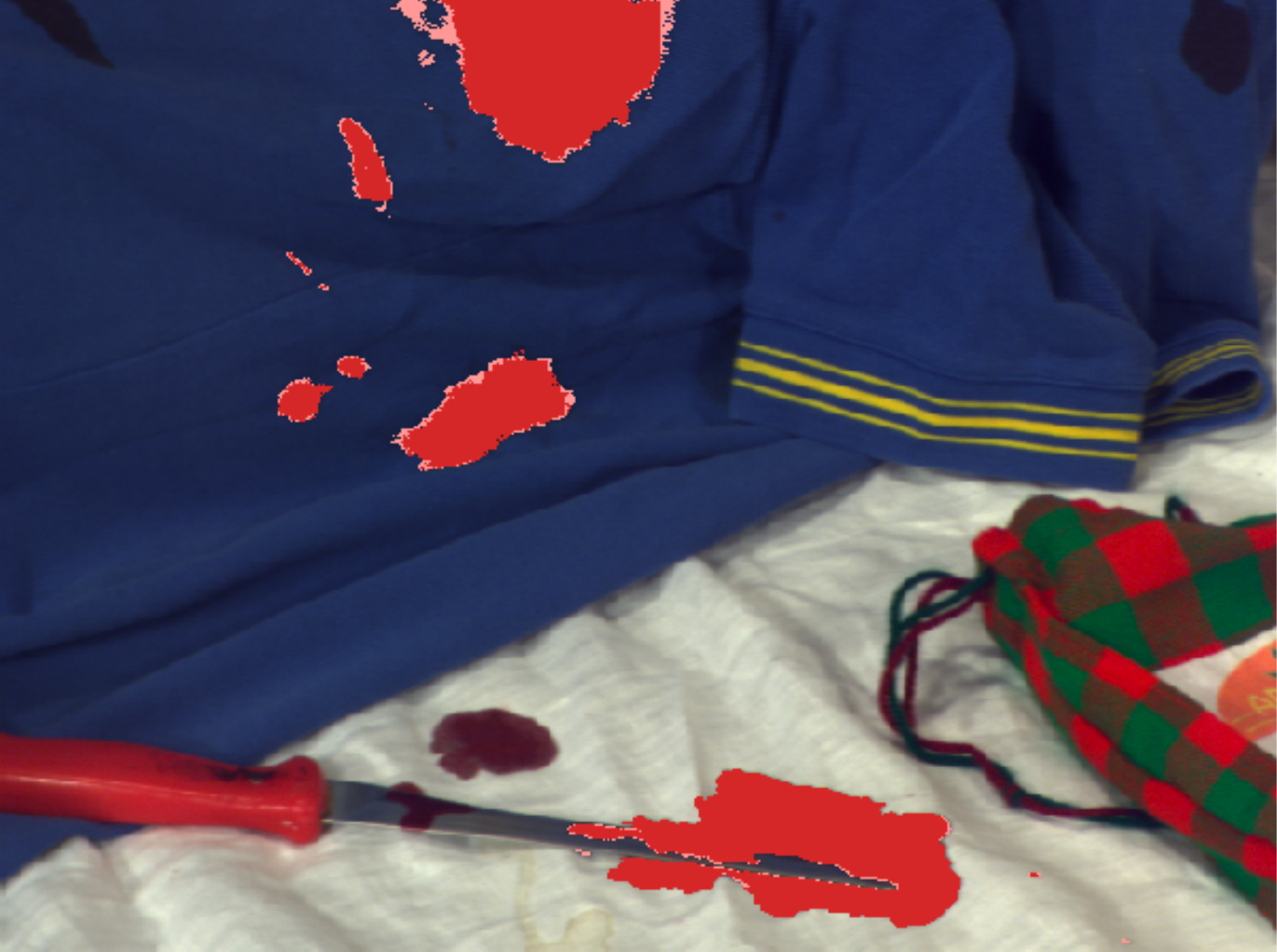}
		\includegraphics[width=0.32\linewidth]{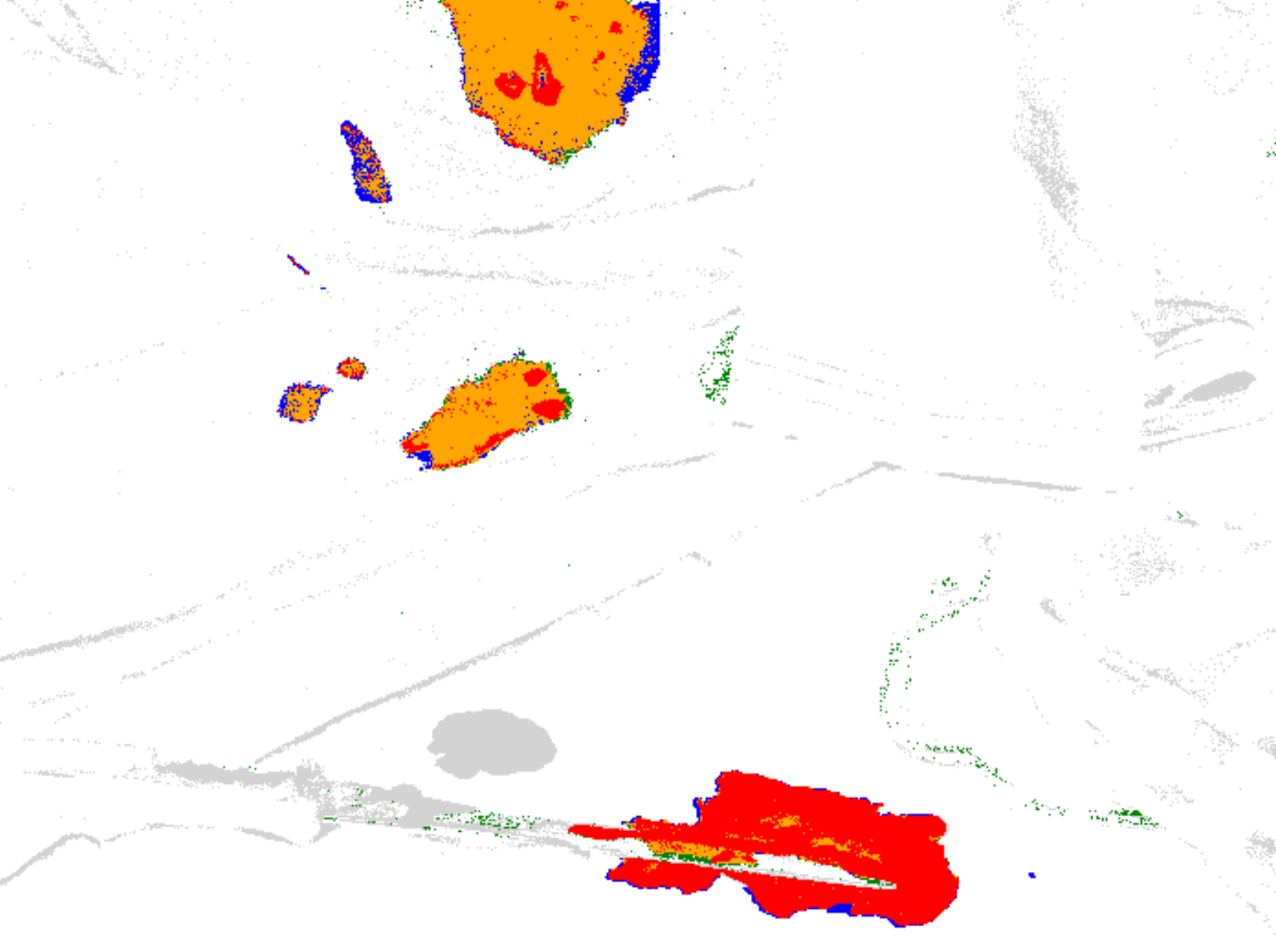}
		\includegraphics[width=0.32\linewidth]{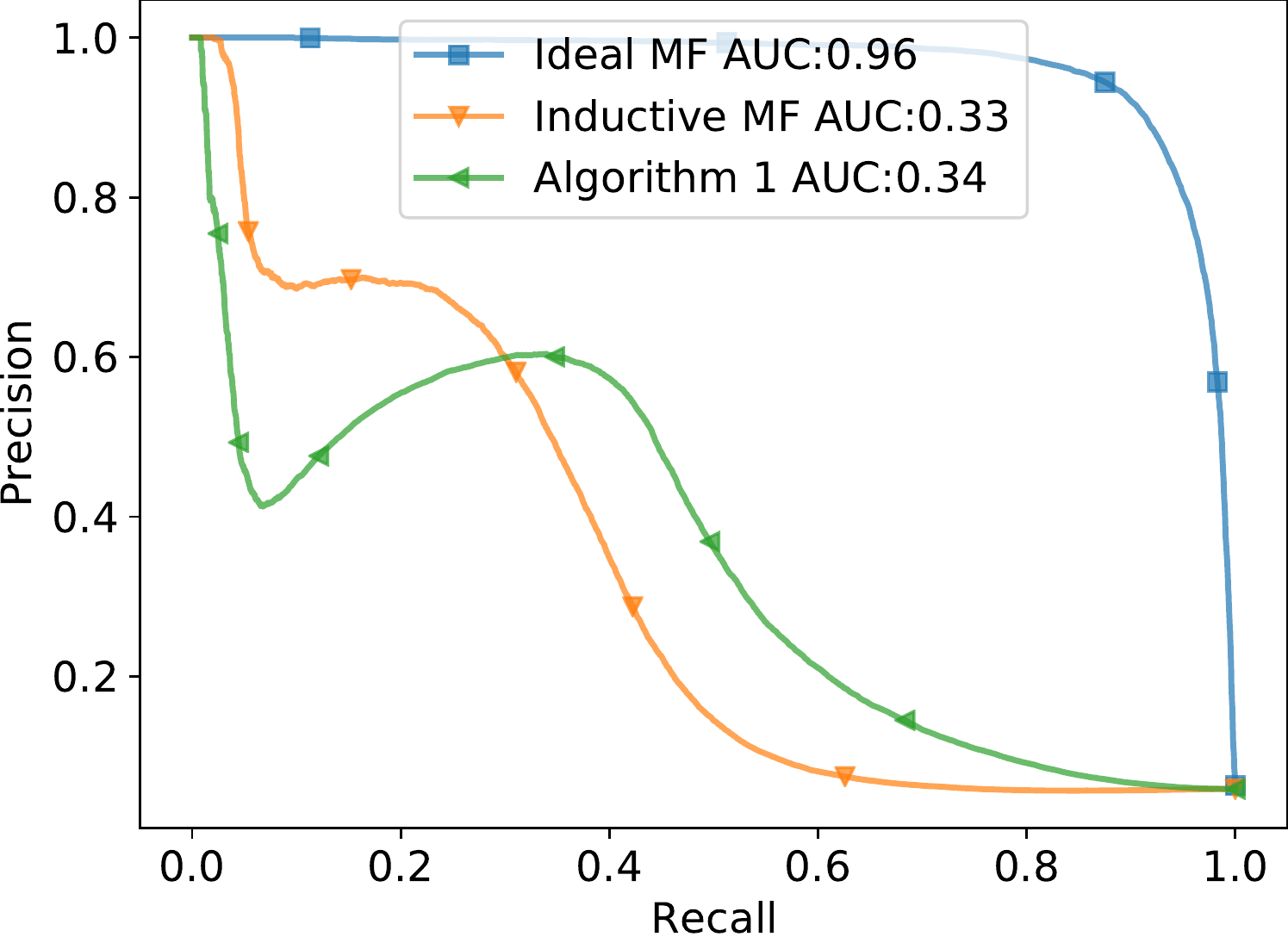}
		\caption{Image \textit{C(1)}}
	\end{subfigure}
	\begin{subfigure}[b]{1.0\textwidth}
		\includegraphics[width=0.32\linewidth]{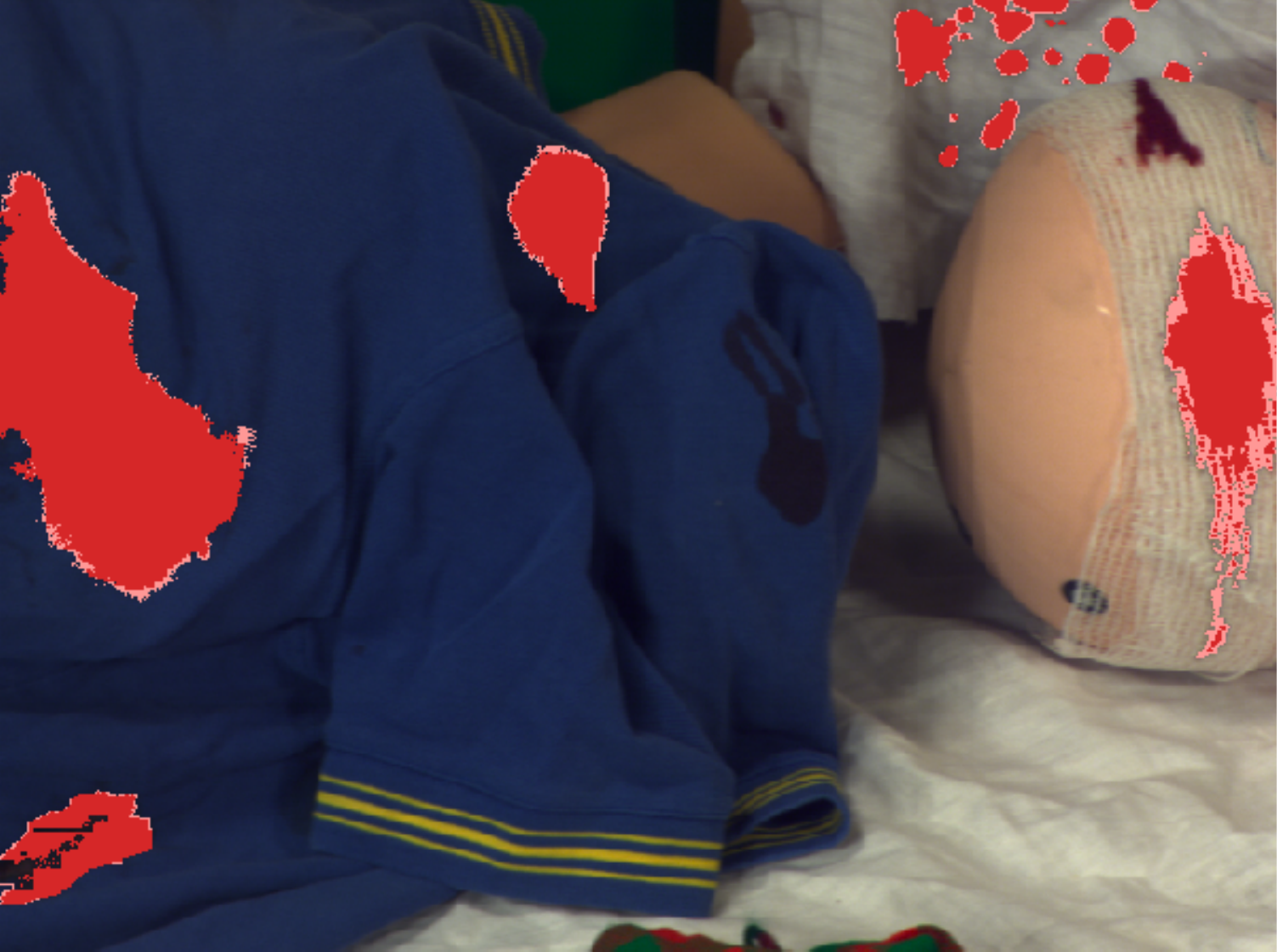}
		\includegraphics[width=0.32\linewidth]{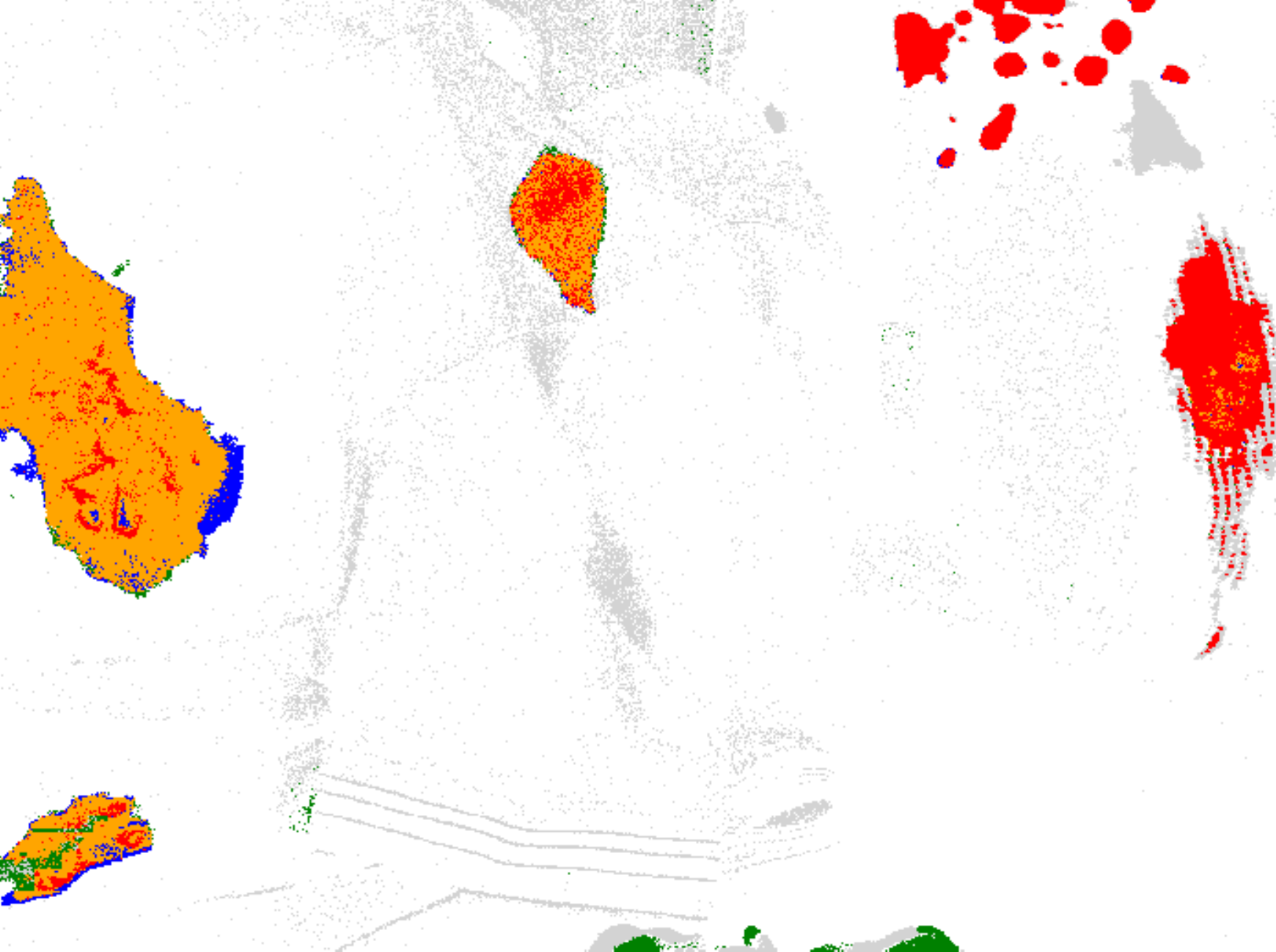}
		\includegraphics[width=0.32\linewidth]{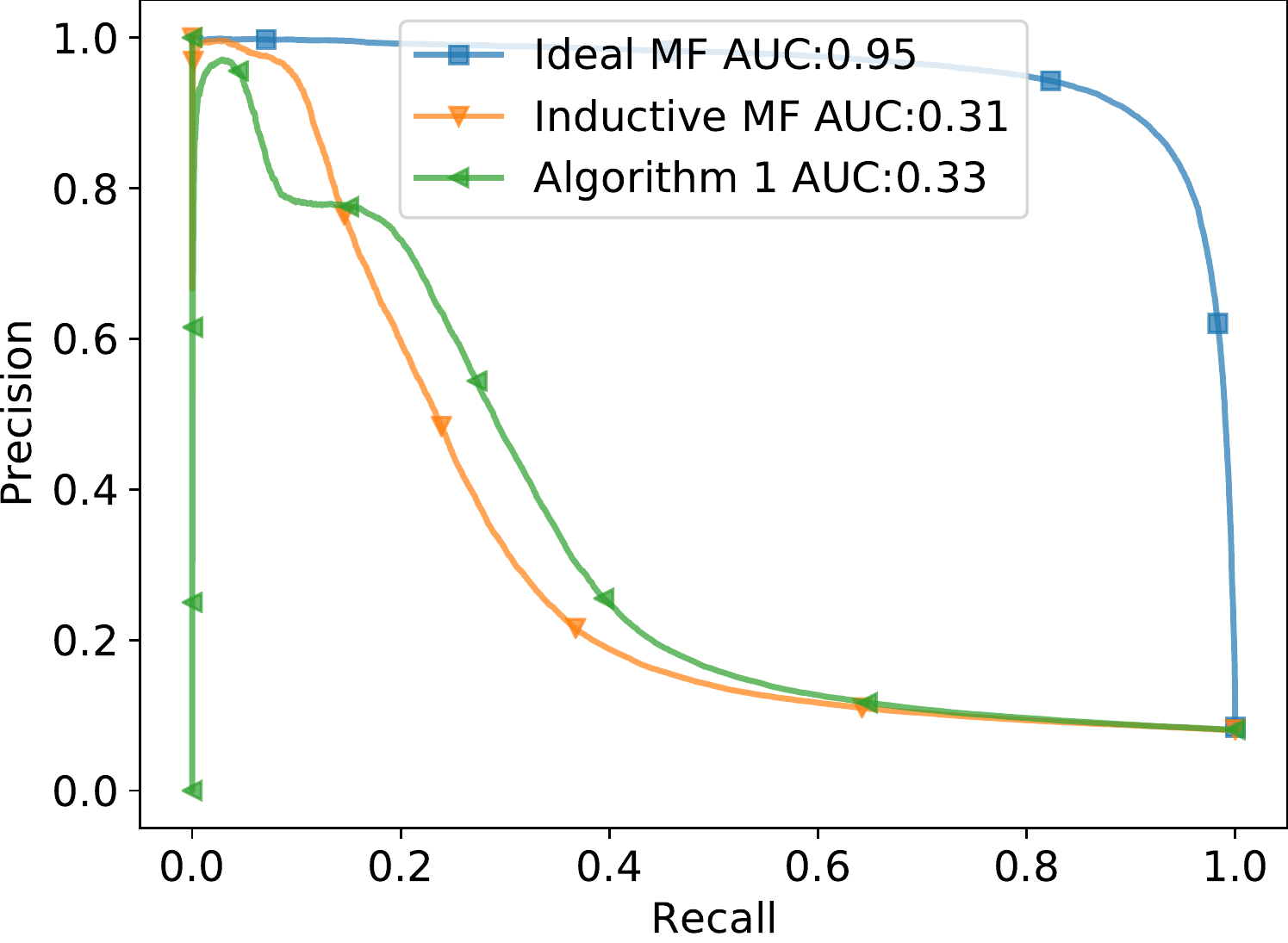}
		\caption{Scene \textit{B(1)}}
	\end{subfigure}
	\begin{subfigure}[b]{1.0\textwidth}
		\includegraphics[width=0.32\linewidth]{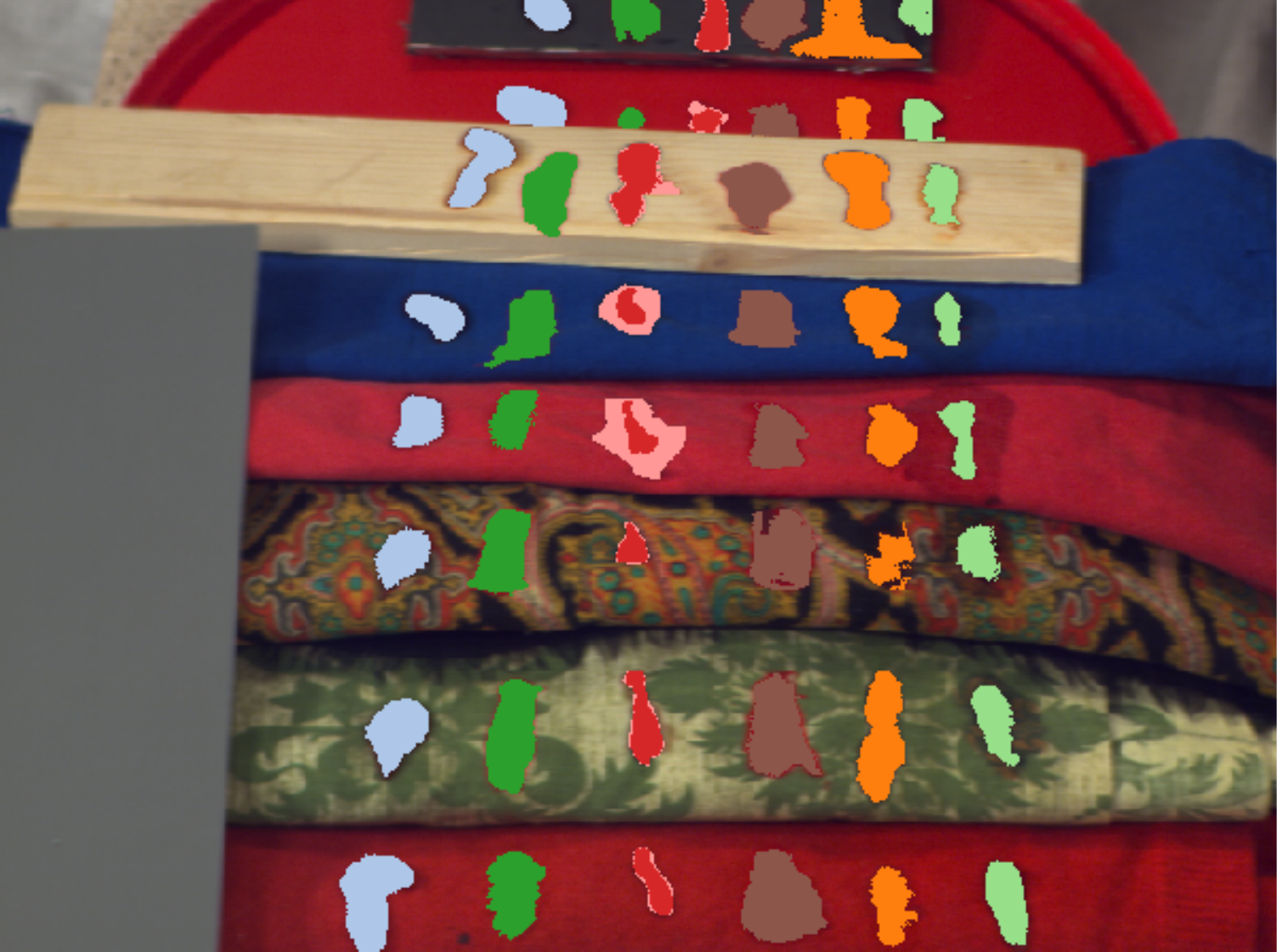}
		\includegraphics[width=0.32\linewidth]{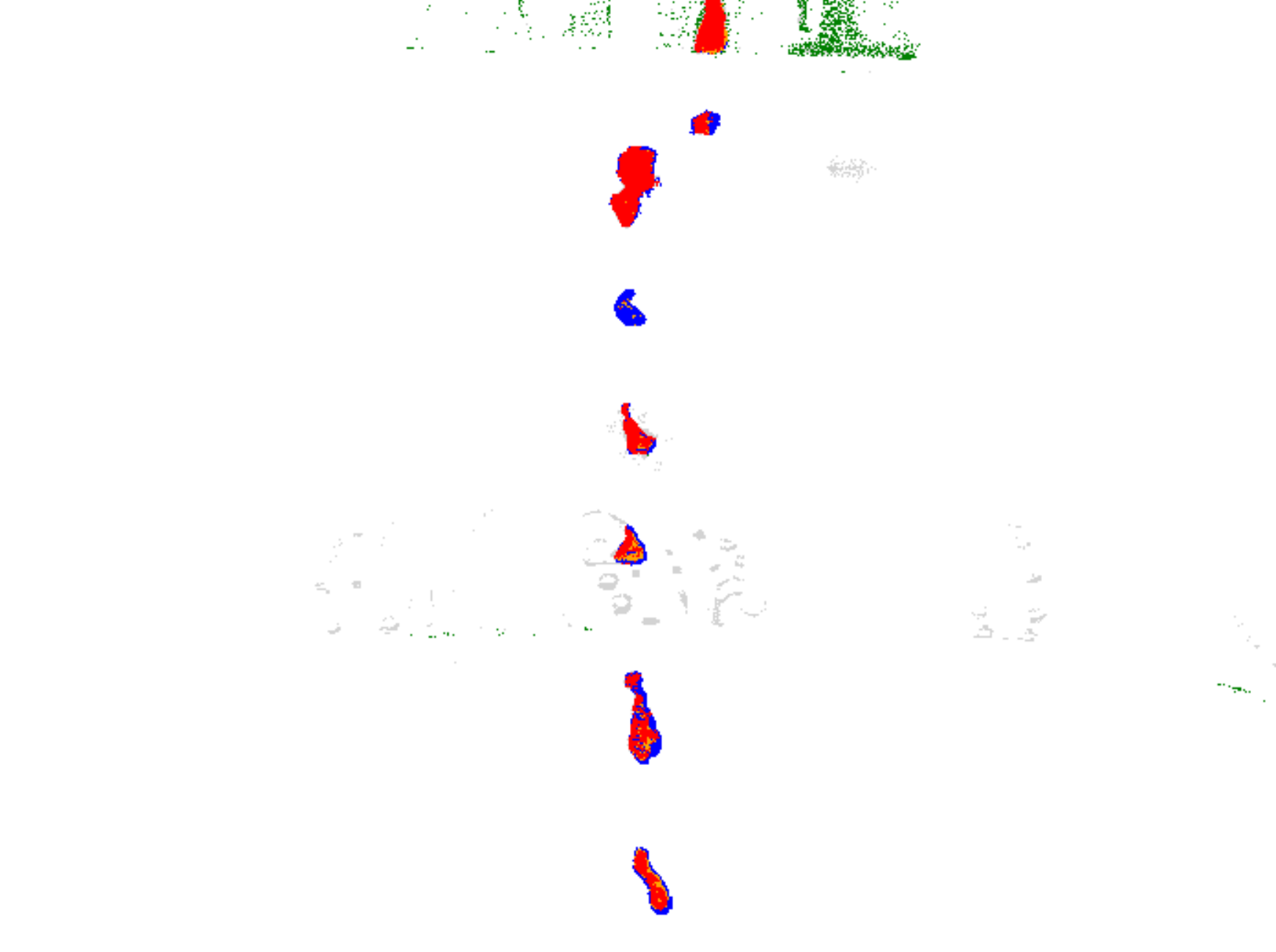}
		\includegraphics[width=0.32\linewidth]{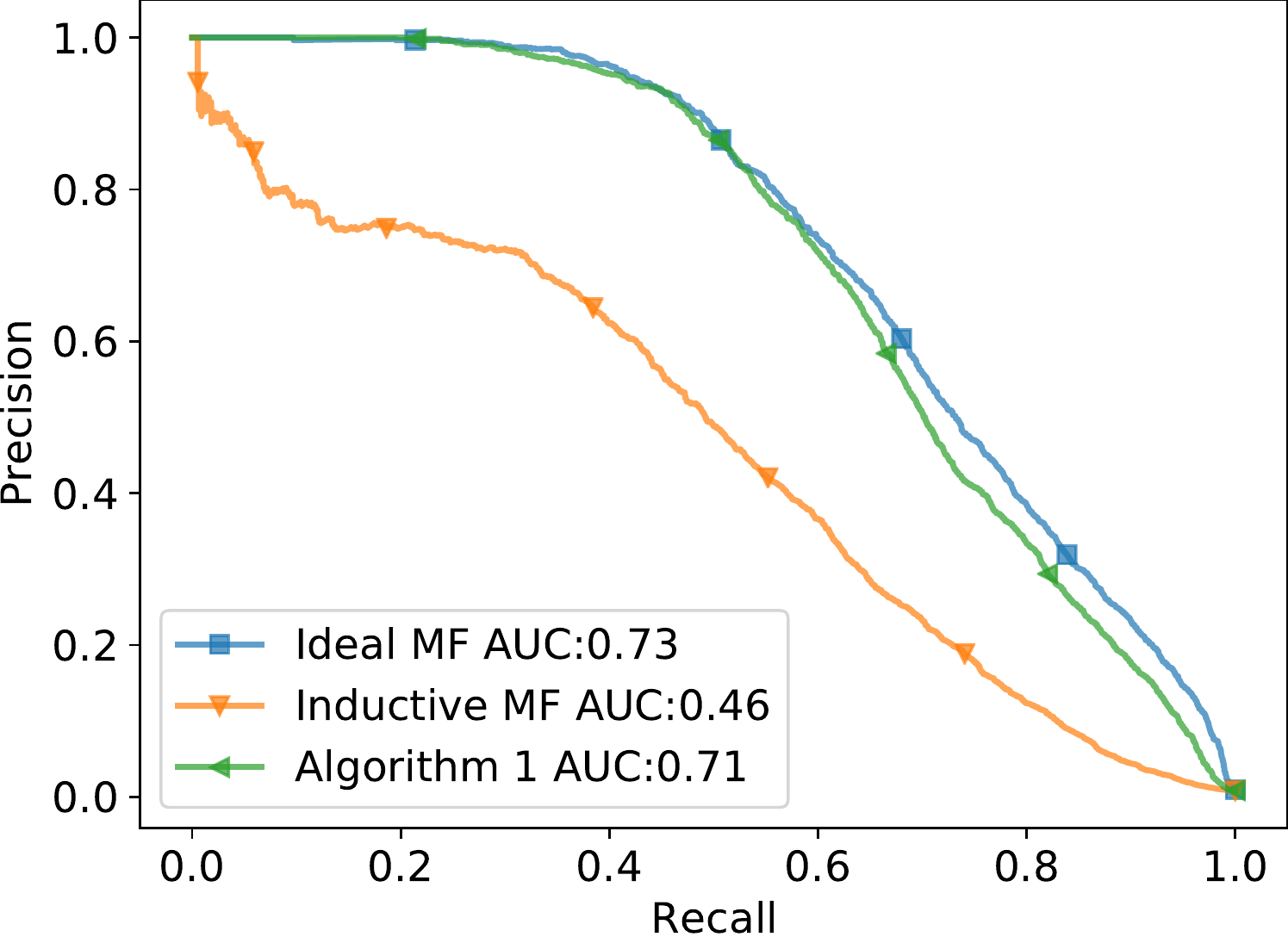}
		\caption{Image \textit{E(1)}}
		\label{fig:res_all_best3:e1}
	\end{subfigure}
	\begin{subfigure}[b]{1.0\textwidth}
		\includegraphics[width=0.32\linewidth]{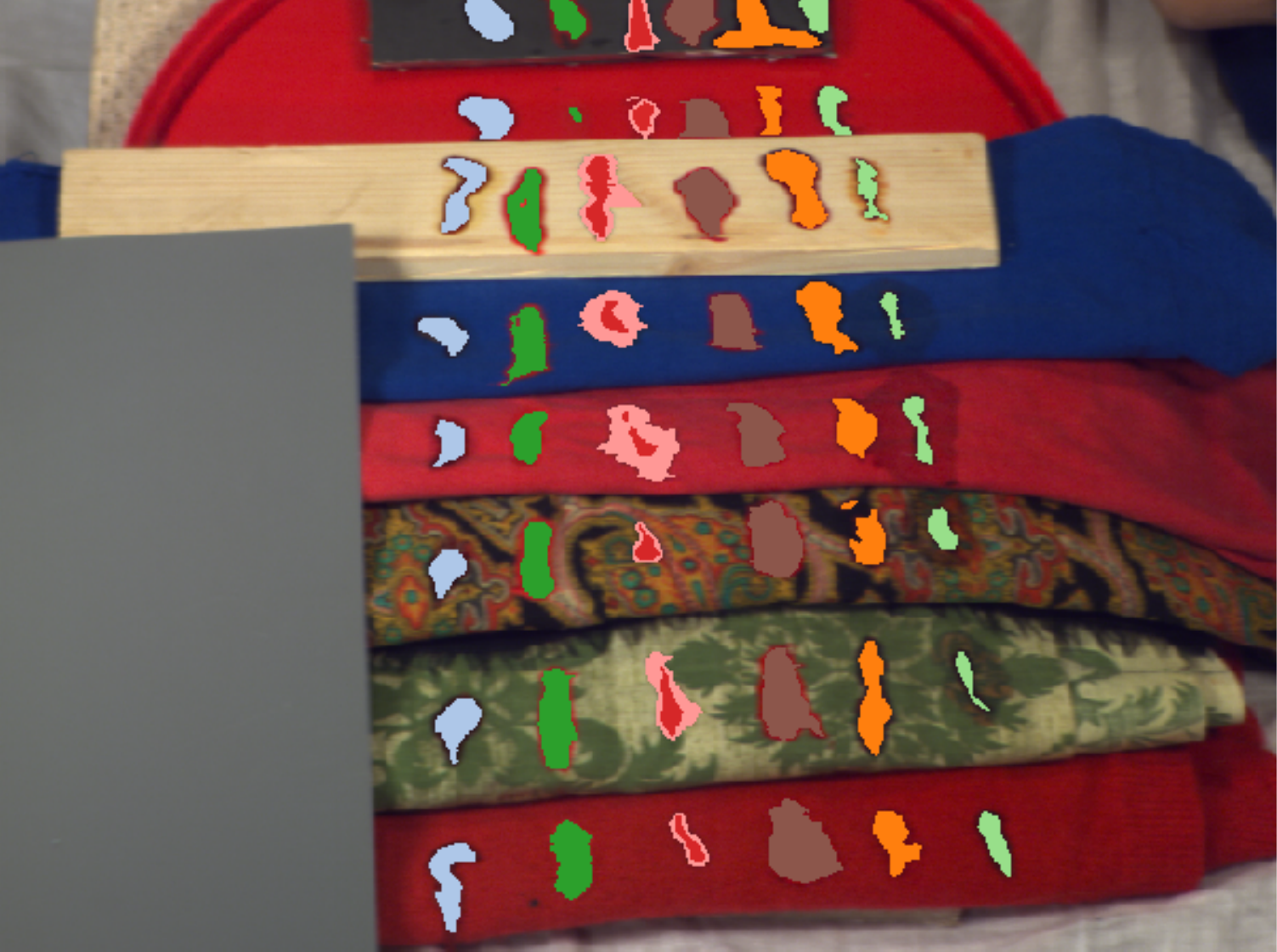}
		\includegraphics[width=0.32\linewidth]{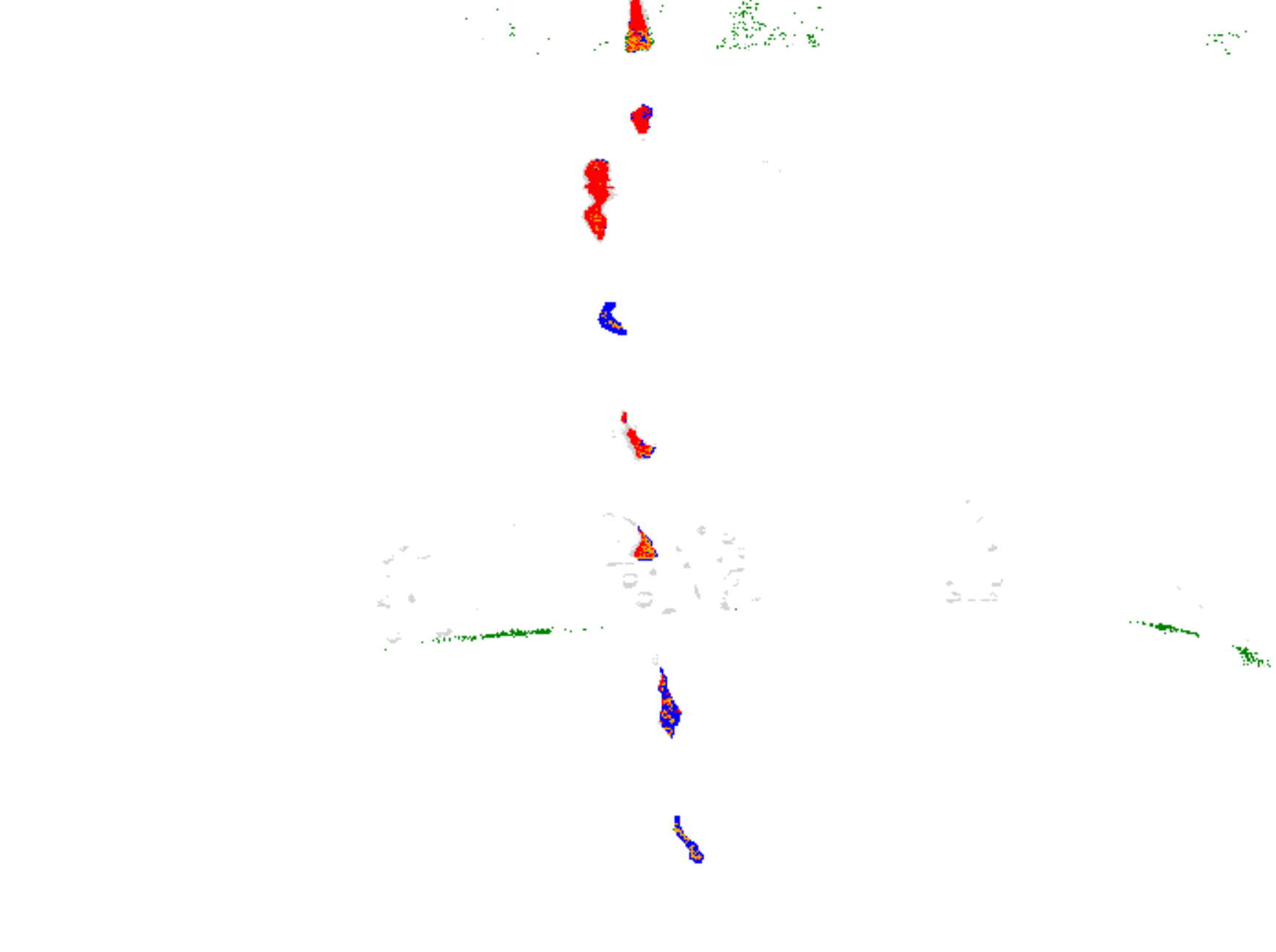}
		\includegraphics[width=0.32\linewidth]{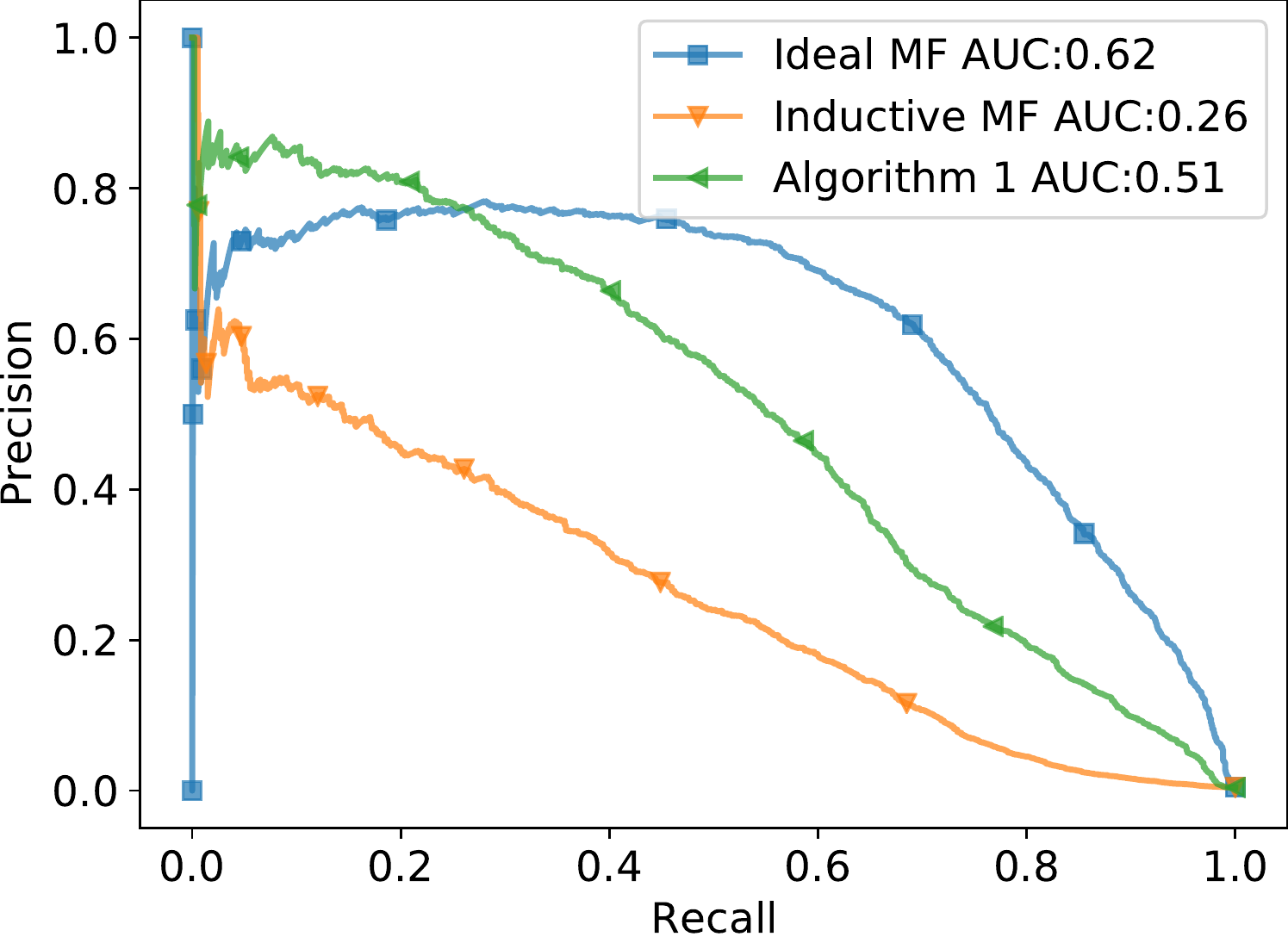}
		\caption{Image \textit{E(7)}}
	\end{subfigure}
	\begin{subfigure}[b]{1.0\textwidth}
		\includegraphics[width=0.32\linewidth]{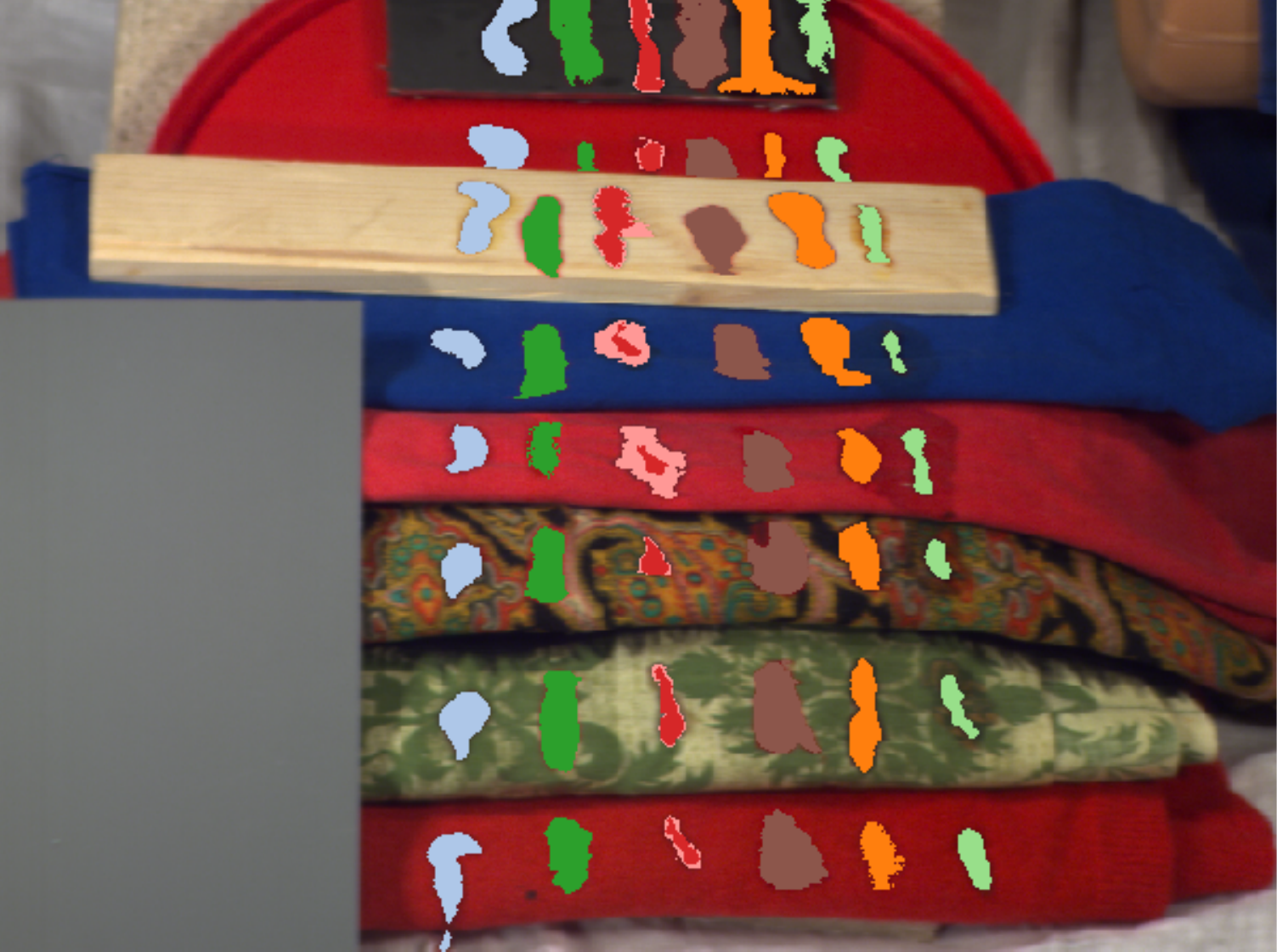}
		\includegraphics[width=0.32\linewidth]{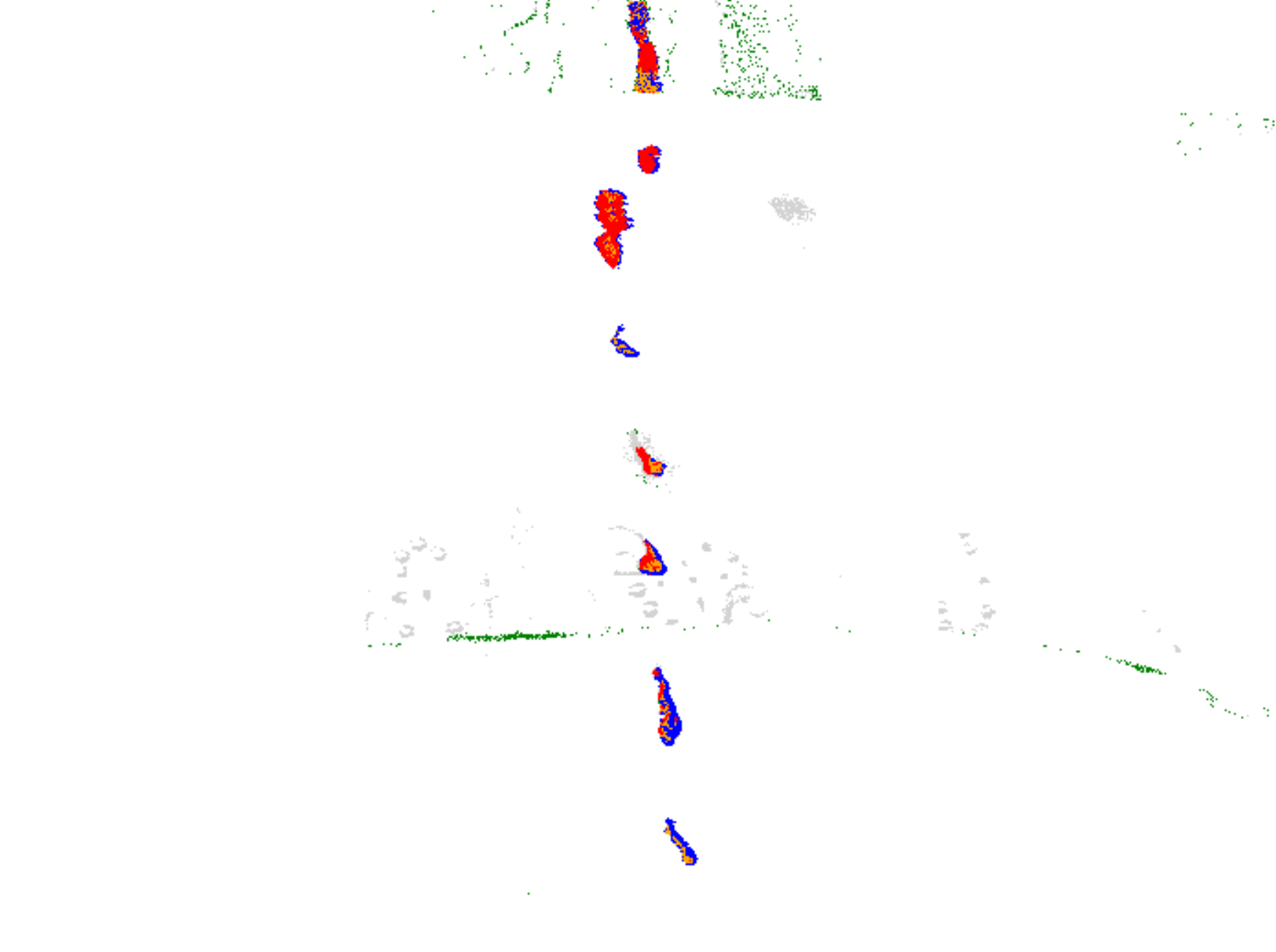}
		\includegraphics[width=0.32\linewidth]{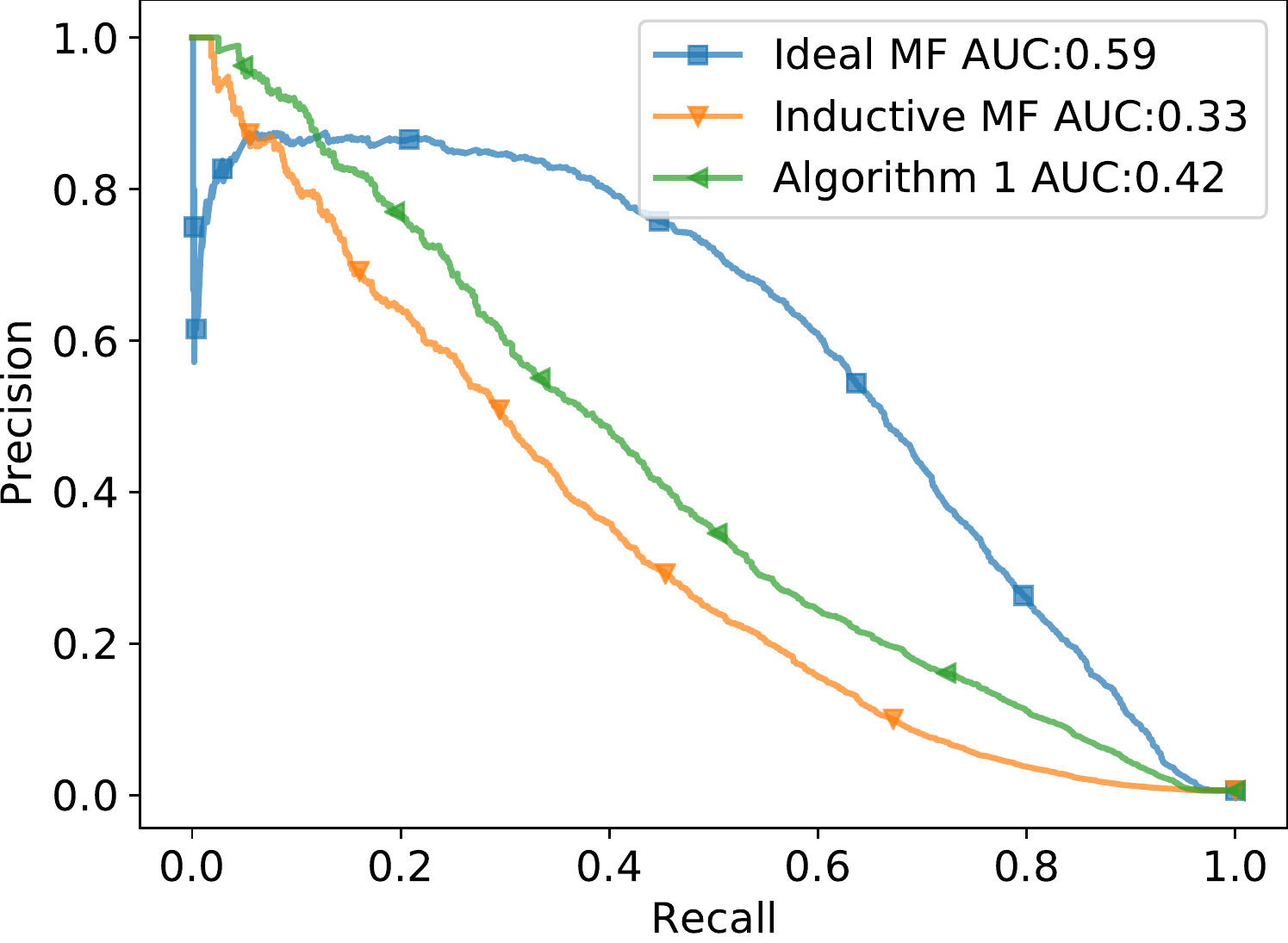}
		\caption{Image \textit{E(21)}}
	\end{subfigure}
	\caption{Performance of the Algorithm~\ref{algorithm:tsmf} for dataset images. Left column presents the ground truth--blood is denoted with red colour. Middle column presents the coloured example output of the detector--correct detections (TP) are coloured red, potential detections (FN of the Algorithm~\ref{algorithm:tsmf} but TP of the \textit{Ideal MF}) are coloured orange, incorrect detections (FP of the Algorithm~\ref{algorithm:tsmf}) are coloured grey, incorrect detections (FP) of the  \textit{Ideal MF} not detected (TN) by the Algorithm~\ref{algorithm:tsmf} are coloured green, detections missed (FN) by both algorithms are coloured blue. The last column presents PR curves for the Algorithm~\ref{algorithm:tsmf}, the \textit{Inductive MF} and the \textit{Ideal MF}.}
	\label{fig:res_all_best3}
\end{figure*}

Visualisation of detection results of the proposed Algorithm~\ref{algorithm:tsmf} on dataset images are presented in Fig.~\ref{fig:res_all_best}, Fig.~\ref{fig:res_all_best2} and Fig.~\ref{fig:res_all_best3}. The third column in these figures presents the PR curves that were used to compute AUC values in Tab.~\ref{tab:new_results}. The second column in figures presents an example of a detection map obtained by setting the threshold $\eta$ (see Sec.~\ref{sec:target_detection}), as previously, to the actual ratio of blood pixels in the image. For such threshold, a perfect detector -- one that would not make errors in pixel scores -- would produce an output perfectly matching the ground truth. The color coding of the maps in the second column is presented in the Tab.~\ref{tab:colorcode}.

\ctable[
caption = {Color-code of Fig.~\ref{fig:res_all_best},~\ref{fig:res_all_best2} and~\ref{fig:res_all_best3}, comparing the proposed algorithm to an idealized detector. Note that this is a subset of the all detection labels, selected for comparision of the \emph{Ideal MF} and Algorithm~\ref{algorithm:tsmf}.},
label   = tab:colorcode,
star
]{llccl}{
	\tnote[a]{Uses blood spectra from the image itself.}
	\tnote[b]{Uses blood spectra from the external library and two-stage detection.}
}{\FL
	Id & Color & \emph{Ideal MF}\tmark[a] & Algorithm~\ref{algorithm:tsmf} (A1)\tmark[b] & Comment\ML
	1 & \textcolor{red}{$\bullet$} Red & -- & TP & Correctly detected by A1\NN
	2 & \textcolor{orange}{$\bullet$} Orange & TP & FN & Not detected by the A1 but detected by \emph{Ideal MF}\NN
	3 & \textcolor{blue}{$\bullet$} Blue & FN & FN & Not detected by both algorithms\NN
	4 & \textcolor{gray}{$\bullet$} Grey & -- & FP & False Positives of A1\NN
	5 & \textcolor{green}{$\bullet$} Green & FP & TN & False Positives of \emph{Ideal MF}, correctly ignored by A1\LL
}

In the Tab~\ref{tab:new_results}, the second column, denoted \emph{Ideal MF}, is the performance of the Matched Filter (MF) detector trained on an average target spectrum of the image itself. We view this result as an approximate of the expected best performance of the MF detector i.e. the `glass ceiling', reachable with a perfectly matching target spectrum. It is thus interesting to compare this detector to the Algorithm~\ref{algorithm:tsmf}, which can be viewed as its realistic counterpart. The comparison include, among others, missed detections of both detectors, correct detections of \emph{Ideal MF} missed by the Algorithm~\ref{algorithm:tsmf}, and False Positives of both detectors:
\begin{itemize}
	\item For some images, e.g. \emph{E(1)}, True Positives of the Algorithm~\ref{algorithm:tsmf} (red colour, first row of Tab.~\ref{tab:colorcode}) correspond quite well to the ground truth, while for other images, such as \emph{F(1)}, only a part of the blood splash is detected. We can observe that for almost every image and every splash/blob visible in this image at least a subset of pixels in the blob is correctly detected. 
	\item True Positives of the \emph{Ideal MF} not detected by the Algorithm~\ref{algorithm:tsmf} (orange colour) could be treated as `potentially detectable' with MF-based algorithm, provided that a proper blood spectrum is available. These pixels are numerous in images where large blood splashes are present, such as \emph{E(1)} or \emph{A(1)}.
	\item False Negatives (FN) according to both the Algorithm~\ref{algorithm:tsmf} and the \textit{Ideal MF} (blue colour) are pixels that can be expected to be difficult to detect with the MF-based algorithm.
	\item Regarding False Positives (FP) of the Algorithm~\ref{algorithm:tsmf} (grey colour), in many images e.g. \emph{F(1)}, the \emph{`artificial blood'} (class~3) is incorrectly detected.
	\item False Positives (FP) of the \emph{Ideal MF} that were not detected by the proposed algorithm (green colour) are often located in less lit or dark areas (e.g.~\emph{E(1)}, top of the image) and regions with the \textit{artificial blood} (class~3).
\end{itemize}


High detection performance of the \emph{Ideal MF} for images such as F(1) or A(1) i.e. AUC(PR) $\approx1.0$ is confirmed by visualizations in Fig.~\ref{fig:res_all_best}-\ref{fig:res_all_best3}, where the union of sets of red and orange pixels corresponds well to the location of blood in the image. In the Tab~\ref{tab:new_results} the \textit{Ideal MF} results are higher that those from other columns, and in particular results from the column \emph{MF}$_{lib}$. The \emph{MF}$_{lib}$ result represents a more `realistic' performance of the detector, obtained by training the MF with spectra from spectral library. This difference illustrates the impact of externally supplied target spectra on the detection performance. Even for spectra obtained in laboratory conditions and with the same age of blood, the performance of the MF is significantly lower than its expected best performance.

The \emph{Inductive MF} spectra were taken from a separate image, but since all images in the dataset were acquired in similar environment and with the same equipment, it can be expected that they will be more similar to the actual target spectra. Indeed, for many images this is the case e.g. for images \emph{F(7)} and \emph{F(21)} the result of AUC(PR) $\approx0.98$ is close to the expected best detection performance. For some images e.g. \emph{F(1a)} the `Inductive' result is lower, but this may be due to the fact, that \emph{F(1a)} and its source image \emph{F(1)} were taken during the first few hours after blood application, when the spectra changes resulting from haemoglobin reactions are largest. The difference in results between \emph{MF}$_{lib}$ and \emph{Inductive MF} columns in Tab~\ref{tab:new_results} may serve as example of the improvement in performance we can obtain by using a different, more similar target spectra.

\begin{figure}
	\centering
	\includegraphics[width=0.9\linewidth]{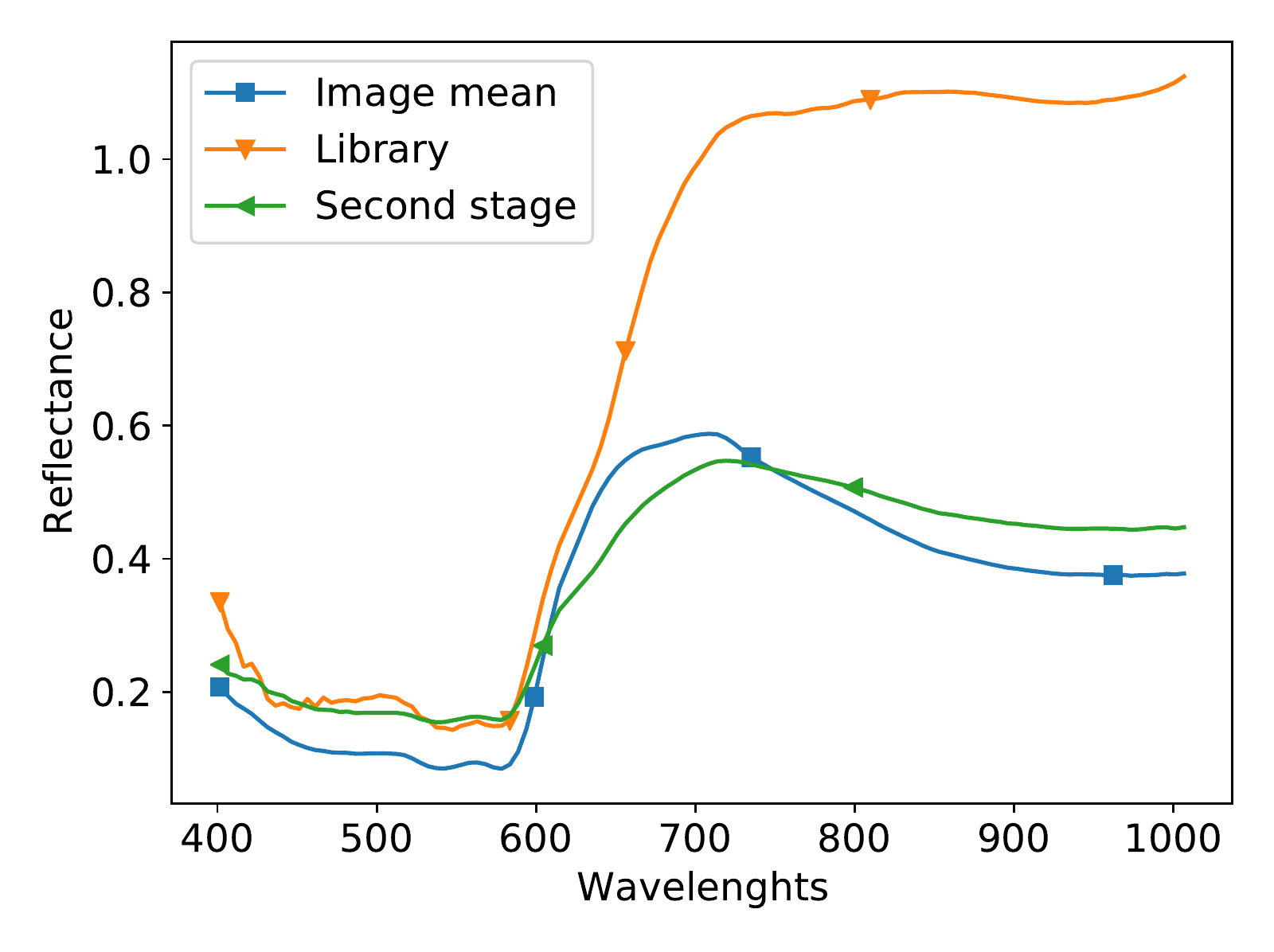}
	\caption{The impact of the second stage of the Algorithm~\ref{algorithm:tsmf} on the target spectrum for the image \emph{F(1)}. The `Image mean' denotes the mean blood spectrum in the image, the `Library' denotes the spectrum from the external library used in our experiments and the `Second stage' denotes the new blood spectrum reference used in the second stage of the algorithm.}
	\label{fig:2stage}
\end{figure}

The results of the Algorithm~\ref{algorithm:tsmf}, represent an improvement from applying `local' spectra obtained by averaging a subset of pixels with the highest score according to the MF detector. It could be expected, that if the initial detector finds blood in the image, the second stage of the algorithm would find a more representative blood spectrum. This approach may be perceived as analogous to Machine Learning task of semi-supervised learning, that extends the initial training set with unlabelled data as e.g. in~\cite{cholewa2018spatial}. The visualisation of difference between target spectra in the first and the second stage of the Algorithm~\ref{algorithm:tsmf} is presented in Fig.~\ref{fig:2stage}.
Results in the Tab~\ref{tab:new_results} support the idea of improving detector's performance with a proposed two-stage approach. In some cases, e.g. for the image \textit{E(1)} the detection result is close to the expected best. 

\section{Discussion}
\label{sec:discussion}

The results of the experiments indicate that Matched Filter detector can achieve high performance in the task of blood detection, provided that suitable reference spectra are available. A mismatch of reference spectra with the image under test may have significant impact on the detection performance. However, heuristic methods, such as the Algorithm~\ref{algorithm:tsmf} can be used to improve the reference quality and final result.

The described dataset was prepared to be consistent with a general approach to hyperspectral non-contact analysis of forensic traces presented e.g. in~\cite{edelman2012hyperspectral,edelman2012hyperspectral2,yang2016comparison,arthur2017image}. The scene layout, lighting, and camera placement were designed to mimic some of the conditions present in a crime scene analysis. Specifically, we take into consideration diverse situations that can appear in forensic practice, e.g. background material interference, in order to provide a way to assess method performance in such cases. The scenarios assumed that the detector is a support for the forensic technician, who is ultimately responsible for the evidence gathering procedure and demonstration of results.

In terms of dataset construction, our approach shares similarities with a number of works published in the literature. 
In~\cite{zhao2019application} the experiment setup is similar to our~\emph{F} frame images; white t-shirt is used with seven blood-like substances.
In~\cite{yang2016spectral} a different setup is used; mechanically defibrinated blood is observed in a liquid form.
The use of coloured cloth background in~\cite{edelman2012identification,li2014application} is very similar to our \emph{E} scene images; our approach complements theirs in terms of using different materials (non-cotton materials, wood, metal) and the non-uniform staining resulting from natural application process.
Simulated crime scene used in~\cite{edelman2012hyperspectral2} was used as a reference in construction of our scene; we've extended the number of objects that were used for blood sample placement.
Our dataset joins together multiple approaches to scene construction in one comprehensive dataset; the usage of hyperspectral images provides additional testing advantage over non-image based spectroscopic data.

Regarding the detection methods, the detectors used in \cite{zhao2019application} and \cite{yang2016spectral} are designed around datasets that are later used for evaluation. Since this approach is influenced by the data used later for verification, we view it as similar to our \emph{Ideal MF} scenario. The high results both in their and in our case (see Tab.~\ref{tab:new_results}) are consistent in conclusion that blood can be reliably detected among visually similar substances, provided that scene elements such as background and lighting are controlled.
A common approach in blood sample identification,  e.g. in~\cite{edelman2012identification,li2014application}, is to use the coefficient of determination. This approach examines goodness of fit expressed as  Pearson correlation coefficient between the measured and reference spectra. Application of this method is preceded by a subtraction or division by the substrate spectra and correction for background absorption. As the proposed Matched Filter approach estimates background properties from the data, it may be simpler to apply in practical conditions.
We also view our results as consistent with~\cite{Pereira2017nir}. We note their experiments are similar to our scenario 2, the \emph{Ideal MF} applied to the frame images (\textit{F(i)}), i.e. the target or train data is provided from the image in question, and the background effect is reduced to minimum. In this case both in our (see Table~\ref{tab:new_results}, second column) and in their case almost no errors are noted. This is additionally verified by another classification study done on our dataset~\cite{Ksiazek2020dhb}, where classification scores for corresponding scenario and images approach $100\%$. However, we explore the problem further by including a more complex scenery (\textit{E(i)} images) and considering scenarios 3 and 4. Our results can be used to asses the expected loss of accuracy in a more challenging case. In such case, our results may provide an initial quantification of a real-life detection scenario, possibly extending many of the results reported in the literature that are confined to highly controlled laboratory conditions, with very high scores reported (see an overview e.g. in~\cite{sharma2018trends}).
Some works, such as~\cite{yang2016comparison} approach the blood detection problem as an anomaly detection (AD) task, e.g. trying to find pixels that are anomalous in the subspace spanned by first PCA components. While authors of~\cite{yang2016comparison} do not express their results using an objective performance measure but rely on visual analysis of detector output, it seems that AD-based approach would not work for our dataset due to the overlap of blood with other pixels, visible in PCA plots presented in Fig.~\ref{fig:blood_properties}.

In the following subsections, we extend the discussion to the influence of individual type of images, impact of target class size and the $N$ parameter of Algorithm~\ref{algorithm:tsmf} and the comparison of Match Filter performance for other (non-blood) classes in our dataset.

\subsection{Sensitivity analysis}
\label{sec:sensitivity_analysis}
Let us compare the performance of the Algorithm~\ref{algorithm:tsmf} for three selected images from the dataset:
\begin{enumerate}
	\item Image \emph{E(1)} (Fig.~\ref{fig:res_all_best3:e1}): correct detections (red areas) of the Algorithm~\ref{algorithm:tsmf} and the \textit{Ideal MF} are similar. The majority of  differences lies in area of False Positives --  the Algorithm~\ref{algorithm:tsmf}  incorrectly detects some pixels (grey colour) belonging to one of the fabrics, a camouflage pattern, while the \emph{Ideal MF} detects some pixels (green colour) belonging to the \emph{`artificial blood'} class. PR curves for both detectors are similar.
	\item Image \emph{F(1)} (Fig.~\ref{fig:res_all_best:f1}): correct detections (red areas) of both algorithms are located mostly on the edges of the blood splash. The rest of the splash is detected only by the \emph{Ideal MF} (orange area), while corresponding detections of the Algorithm~\ref{algorithm:tsmf} are False Positives. 
	\item Image \emph{A(1)} (Fig.~\ref{fig:res_all_best2:a1}): correct detections (red areas) of both algorithms incorporate the majority of blood splashes on the white fabric in the top area of the image, as well as small subsets of pixels located on large splashes in the bottom. The rest of pixels in these two splashes were detected only by the \emph{Ideal MF}. The difference in performance is also visible in PR curves.
\end{enumerate}

The difference in performance of the Algorithm~\ref{algorithm:tsmf} for the presented three images is evident. For the image \emph{E(1)} the algorithm's performance is close to the expected best (\emph{Ideal MF}).  For images \emph{F(1)} and \emph{A(1)} the algorithm detected only a subset of pixels. In addition, as the image \emph{A(1)} contains three distinct areas with blood splashes (splashes in the top, bandage at the bottom, shirt on the left), the majority of pixels detected by the Algorithm~\ref{algorithm:tsmf} were located in the top area.

\begin{figure}
	\centering
	\begin{subfigure}[b]{1.0\linewidth}
		\includegraphics[width=1.0\linewidth]{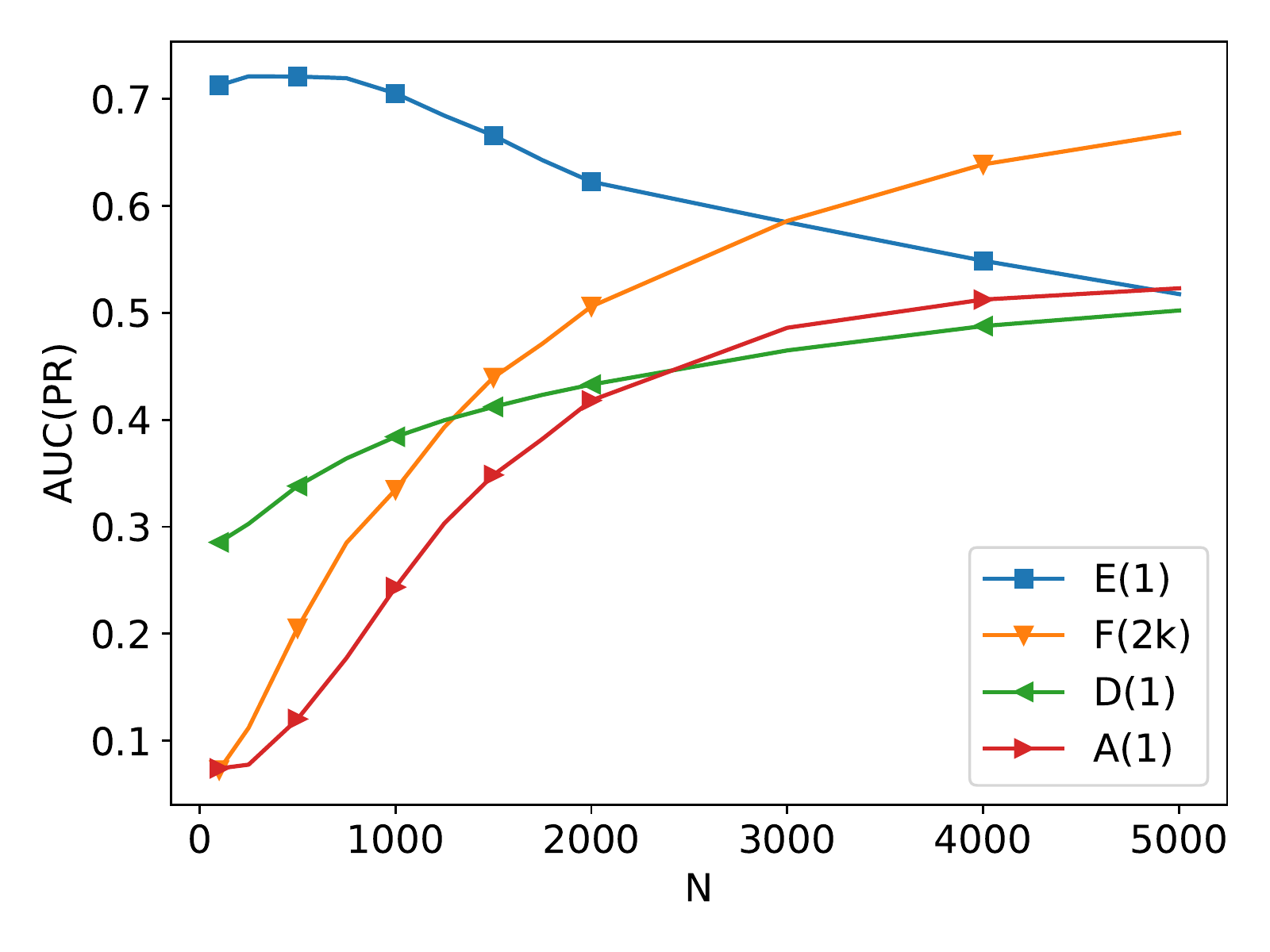}
	\end{subfigure}
	\caption{Sensitivity analysis of the Algorithm~\ref{algorithm:tsmf} for selected scenes: $N$ denotes a number of pixels used for spectra averaging}
	\label{fig:senstitivity}
\end{figure}

These differences can be partly attributed to the effect of the second stage of the Algorithm~\ref{algorithm:tsmf} and in particular its parameter $N$, which denotes the number of pixels averaged to obtain a new target spectrum. If there is a significant number of pixels in the image where target class is visible, averaging a small subset of best scoring pixels can lead to the situation when the new target signature becomes representative only for one particular subclass of the target class, such as the top area in the image \textit{A(1)}. This suggests that increasing the parameter $N$ may increase the detector performance. The impact of this parameter is presented in the Fig.~\ref{fig:senstitivity}. As expected, for images where large blood splashes are present, increasing the value of the parameter increases the algorithm's performance. On the contrary, on images with small number of pixels to detect, small value of the parameter is sufficient. 

This leads us to the question, how to estimate the value of the parameter $N$? If we knew the number of target pixels in the image, we could proportionally set the value of the parameter $N$, in an analogous way to the detection threshold $\eta$ (see Sec.~\ref{sec:target_detection}). Without this knowledge, heuristic estimation of this parameter may be possible, e.g. by analysing histograms of detection output. However, in our opinion such heuristics is beyond the scope of this paper. Therefore, for simplicity, in all our experiments we used a single value of this parameter $N=1000$. As a reference, we provide results for different values of parameter $N$ in Tab.~\ref{tab:new_results_N}.

\ctable[
cap = Performance of Algorithm~\ref{algorithm:tsmf} for different values of N,
caption     =  Peformance of the Algorithm~\ref{algorithm:tsmf} expressed as AUC(PR) for different values of the parameter N. Percentage values are relative to the number of pixels from the target class in the image. ,
label   = tab:new_results_N,
]{llccc}{	
}{     \FL
	Code&750&1000&10\%&20\%\ML
\textit{F(1)}&0.46&0.47&0.49&\textbf{0.51}\NN
\textit{F(1s)}&0.81&0.73&\textbf{0.82}&0.63\NN
\textit{F(1a)}&0.39&0.41&0.43&\textbf{0.43}\NN
\textit{F(2)}&0.69&0.72&0.79&\textbf{0.81}\NN
\textit{F(2k)}&0.29&0.33&0.56&\textbf{0.68}\NN
\textit{F(7)}&0.80&0.83&0.90&\textbf{0.93}\NN
\textit{F(21)}&0.53&0.59&0.77&\textbf{0.82}\NN
\textit{D(1)}&0.36&0.38&0.51&\textbf{0.51}\NN
\textit{A(1)}&0.18&0.24&0.36&0.49\NN
\textit{B(1)}&0.34&0.33&0.33&\textbf{0.34}\NN
\textit{C(1)}&0.33&0.34&0.38&\textbf{0.40}\NN
\textit{E(1)}&0.72&0.71&0.72&\textbf{0.72}\NN
\textit{E(7)}&0.54&0.51&\textbf{0.56}&0.55\NN
\textit{E(21)}&0.44&0.42&0.50&\textbf{0.48}\LL
}

\subsubsection{The impact of target class size}
\label{sec:disc_impact_target__class_size}

Results presented in Fig.~\ref{fig:senstitivity} suggest, that while increasing the parameter $N$ of the Algorithm~\ref{algorithm:tsmf} can increase the detection performance this performance is still lower that the expected best result, represented by the \textit{Ideal MF} scores in Tab.~\ref{tab:new_results}. This indicates that images with large blood splashes are more difficult for the Algorithm~\ref{algorithm:tsmf}. To illustrate this, a complementary detection experiment was performed with the Algorithms~\ref{algorithm:tsmf} applied to a modified image \textit{F(1)}. The modification consisted of randomly removing a subset of pixels belonging to the blood class, leaving the rest of the image intact. The number of blood pixels remaining is denoted as $T$. The parameter $N$ was either constant ($N=1000$) or set as a percentage of the size of a subset of blood pixels in the image. This allows us to examine the effect of the size of blood sample present on the performance of the detector. 

\begin{figure}
	\centering
	\begin{subfigure}[b]{1.0\linewidth}
		\includegraphics[width=1.0\linewidth]{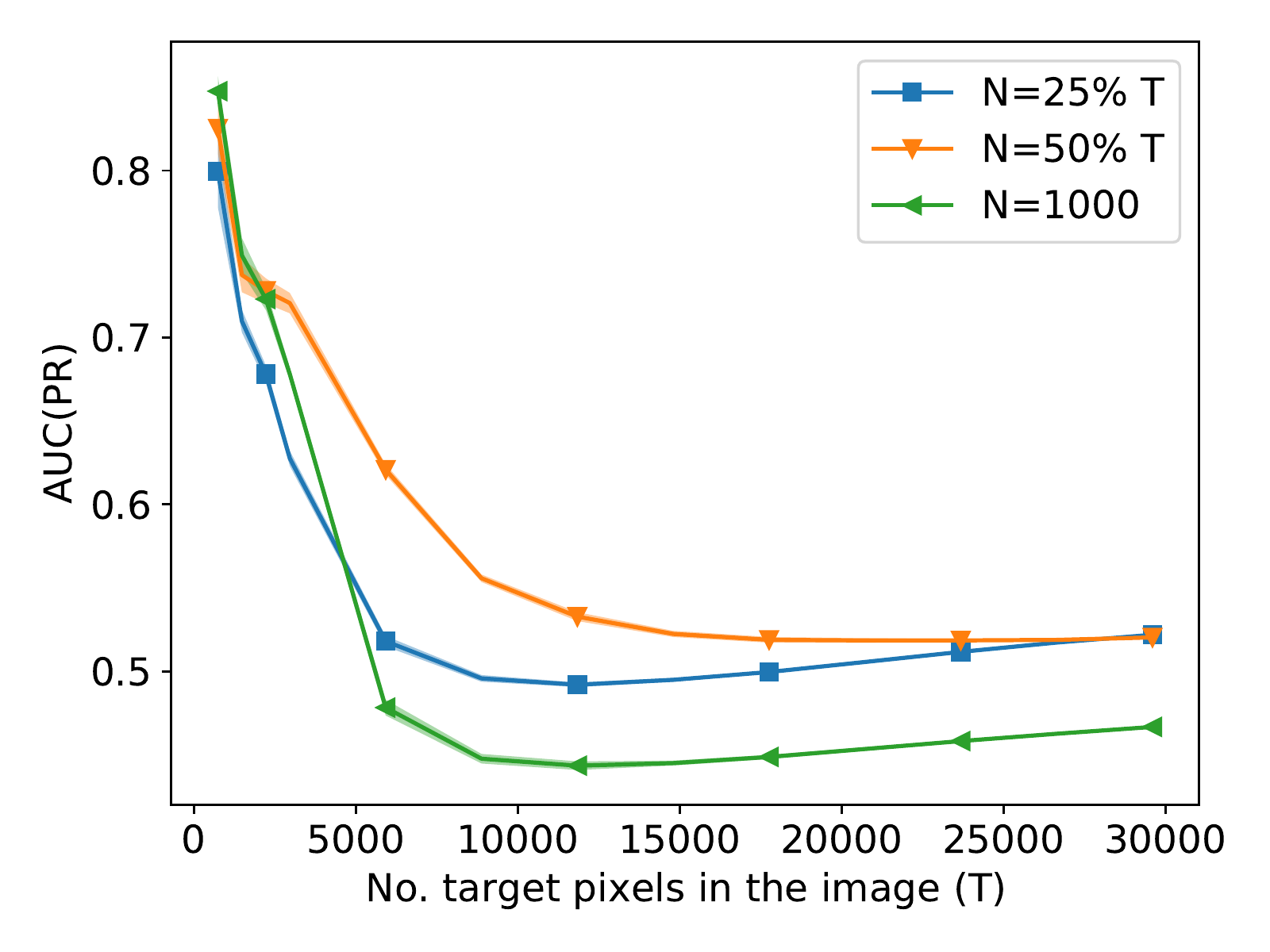}
	\end{subfigure}
	\caption{The impact of the number of blood pixels in the image ($T$) on detection performance. Horizontal axis represents a size of a subset of pixels left in the image \textit{F(1)}. Results are presented for three values of a parameter $N$ where it is either constant (e.g. $N=1000$) or set as a percentage of a value of $T$. Values are averaged over five experiments, a shading denotes standard deviation of the result resulting from random draws of a subset of pixels.}
	\label{fig:senstitivity2}
\end{figure}

Results of the experiment are presented in Fig.~\ref{fig:senstitivity2}. A sharp drop in the performance of the Algorithm~\ref{algorithm:tsmf} for the number of pixels $T<2500$ shows that the performance is inversely proportional to the size of the target class in the image. A detailed examination of the plot reveals a possible combination of two effects; one that produces exponential drop in performance and the second, which provides a linear increase. Additional analysis suggests that the first effect may be related to a `contamination' of the covariance matrix with blood -- target -- pixels. In the ideal case, the background only covariance matrix and data mean is supplied to the algorithm; introducing sizeable target influence can bias the their estimates and in consequence reduce the performance. The second effect is probably due to improved quality of the target pattern; with more blood pixels available the first stage of the Algorithm~\ref{algorithm:tsmf} can produce a more accurate template for the second stage. The interaction between those two effects produces a curve with a minimum for the value for $T\approx 10k$. The number of averaged pixels i.e. the value of the parameter $N$ of the Algorithm~\ref{algorithm:tsmf} visibly affects the result. For the image \emph{F(1)} a `correct' value seems to be about half of the \textit{blood} class size, which is consistent with results in the Tab.~\ref{tab:new_results_N}.

A similar effect was mentioned in~\cite{manolakis2009there}, where such performance degradation was attributed to corruption of a background covariance matrix, estimated as a global sample covariance matrix, by target spectra. 
Note that when we consider the Fig.~\ref{fig:blood_properties}, a significant influence of blood samples on covariance matrix is very likely, as blood and non-blood classes overlap each other. Authors of~\cite{manolakis2009there} suggested that removing best scoring pixels and reapplying the detector may improve its performance.  We performed an initial experiment and verified that this modification indeed allowed to slightly increase the performance of the proposed algorithm. However, since this improvement was minor and would complicate the algorithm e.g. by introducing another parameter related to the number of ignored pixels, we decided that it is beyond the scope of this paper.

The above observations suggest that the size of the target class is of particular interest to the problem of blood detection. In~\cite{manolakis2003hyperspectral}, the authors discuss how the small size of the target class makes it difficult to estimate the parameters of its distribution and to evaluate the result. These considerations are relevant in the case of, for example, hyperspectral photos taken from an aeroplane, where the target class may consist of several pixels. However, they do not seem to be applicable for the images considered in this article, as the size of the target class is significantly larger and ranges between $1.5k$ and $50k$ pixels. On the other hand, a numerous target class causes the problems described above with respect to the properties of the global sample covariance matrix. This suggests that the problem of hyperspectral detection may require a more complex approach and may be of particular interest from the point of view of Machine Learning methods, e.g. classification algorithms.

To summarise, the performance of the Algorithm~\ref{algorithm:tsmf} is higher for images where the size of a target class is small. To increase its performance for images with large number of blood pixels, the value of parameter $N$ should also be increased.

\subsection{Validity of choice of Matched Filters and blood-like classes}

The main motivation behind introducing the dataset is the possibility of digital replication of a case study of forensic trace detection. Therefore, other blood like substances were selected as a set of reasonable adversaries in a detection scenario. This leaves an open question how well the detector would perform for substances other than blood.
To investigate this, a supplementary experiment was performed where the \emph{Ideal MF} and the \emph{Inductive MF} detectors were used to detect classes in the \emph{F} and \emph{E} images. The experiment design was the same as that described in the Sec.~\ref{sec:experiments_and_results}. 

Results of this experiment are presented in the Tab.~\ref{tab:results_all_ideal} for the \emph{Ideal MF} and Tab.~\ref{tab:results_all_inductive} for the \emph{Inductive MF}. Results for the \emph{Ideal MF} indicate that all classes were detected with a similar, high performance in \emph{F} images. For \emph{E} images, detection performance for the \emph{`blood'} class was higher than for \emph{`artificial blood'} and \emph{`tomato concentrate'} but lower than the performance for other classes. For the \emph{Inductive MF} in every image there was a subset of classes that was better detectable than \emph{`blood'}. For \emph{E} images, detection performances of all classes except \emph{`poster paint'} and \textit{`acrylic paint'} were worse than for the \emph{`blood'} class.
For \emph{F} images results were more varied, in particular for \emph{F(1)}, \emph{F(1a)}, \emph{F(2k)} detection results for blood were the worst among all classes. One possible explanation of the this result is the impact of a target class size, discussed in Sec.~\ref{sec:disc_impact_target__class_size}. \emph{`Blood'} class is the most numerous in \emph{F} images which likely affects the background covariance matrix.

The results show significant diversity in the way the classes interact with the Matched Filter method. The \emph{`acrylic paint'} and \emph{`poster paints'} are the easiest do detect, which may be due to low penetrability of light and in consequence lower mixture ratio with the background substances. Others, i.e. \emph{`artificial blood'}, \emph{`tomato concentrate'} and \emph{`ketchup'} are challenging, in particular to inductive approach, probably due to soaking into the background elements. In general, the classes have diverse spectral responses and the \emph{`blood'} class is neither the best nor the worst among them. We view this as an additional argument confining the choice of the adversarial classes in the proposed dataset.

\section{Conclusions}
\label{sec:conclusion}
We presented a diverse dataset aimed to facilitate the development of blood detection algorithms in hyperspectral images. In order to explore this dataset, we approached the task of blood detection from the perspective of hyperspectral target detection. This allowed us to apply well-understood statistical detectors and observe their performance for different images in the dataset. In our opinion this approach is more general than a dedicated algorithm focused e.g. on detection of characteristic bands, while at the same time it provides a logical reference for more specialised methods.

Our results show that blood is spectrally distinct and can be quite reliably detected when a matching set of training examples is available. However, spectra changes induced by ageing of the blood and the fact that spectra are mixed with unknown background can make finding such training set a challenging task. This problem can be approached with heuristic approaches, in a similar way to the detection algorithm presented in our experiment. A possible extension of this approach would be to use spatial-spectral processing in order to make use of the fact that blood pixels can be expected to from spatial groups as e.g. in~\cite{cholewa2018spatial}. Another approach would be to employ mixing models as e.g. in~\cite{skjelvareid2017detection} in order to extract haemoglobin endmembers.

We also believe that the proposed dataset can be an interesting case for the development of classification algorithms. In particular, the development of an `inductive' classifier  trained on one image and tested on others is an open problem.

\ctable[
cap = Detector performance on all classes,
caption     =  Detection performance of the \textit{Ideal MF} for all classes expressed as Area Under Curve (AUC) of the Precision-Recall (PR) curve. The hypen indicates that the class is not present in the image,
label   = tab:results_all_ideal
]{lccccccc}{	
}{     \FL
&\multicolumn{6}{c}{Class number}\NN
Code&\rotatebox[origin=c]{90}{\textit{blood}}&
\rotatebox[origin=c]{90}{\textit{ketchup}}&
\rotatebox[origin=c]{90}{\textit{art. blood}}&
\rotatebox[origin=c]{90}{\textit{beetroot juice}}&
\rotatebox[origin=c]{90}{\textit{poster paint}}&
\rotatebox[origin=c]{90}{\textit{tomato conc.}}&
\rotatebox[origin=c]{90}{\textit{acrylic paint}}\ML
F(1)&1.00&0.99&0.95&0.99&1.00&0.96&0.99\NN
F(1s)&1.00&1.00&0.88&1.00&1.00&0.96&1.00\NN
F(1a)&1.00&0.99&0.96&0.99&1.00&0.98&1.00\NN
F(2)&0.99&0.99&0.98&0.99&0.99&0.94&0.99\NN
F(2k)&1.00&1.00&0.91&0.98&1.00&0.95&1.00\NN
F(7)&1.00&0.97&0.99&0.99&1.00&0.96&1.00\NN
F(21)&1.00&0.98&0.99&0.98&1.00&0.90&0.99\NN
E(1)&0.73&0.92&0.50&-&0.89&0.33&0.90\NN
E(7)&0.62&0.79&0.52&-&0.71&0.36&0.87\NN
E(21)&0.59&0.73&0.43&-&0.81&0.30&0.92\LL
}

\ctable[
cap = Detector performance on all classes,
caption     =  Detection performance of the \textit{Inductive MF} for all classes expressed as Area Under Curve (AUC) of the Precision-Recall (PR) curve. The hypen indicates that the class is not present in the image,
label   = tab:results_all_inductive
]{lccccccc}{	
}{     \FL
	&\multicolumn{6}{c}{Class number}\NN
	Code&\rotatebox[origin=c]{90}{\textit{blood}}&
	\rotatebox[origin=c]{90}{\textit{ketchup}}&
	\rotatebox[origin=c]{90}{\textit{art. blood}}&
	\rotatebox[origin=c]{90}{\textit{beetroot juice}}&
	\rotatebox[origin=c]{90}{\textit{poster paint}}&
	\rotatebox[origin=c]{90}{\textit{tomato conc.}}&
	\rotatebox[origin=c]{90}{\textit{acrylic paint}}\ML
F(1)&0.41&0.82&0.92&0.97&1.00&0.91&0.98\NN
F(1s)&0.92&0.57&0.76&0.63&1.00&0.95&0.99\NN
F(1a)&0.17&0.73&0.93&0.99&1.00&0.69&0.99\NN
F(2)&0.61&0.34&0.02&0.78&0.85&0.47&0.96\NN
F(2k)&0.33&0.40&0.56&0.68&0.98&0.82&0.93\NN
F(7)&0.98&0.96&0.98&0.99&1.00&0.95&1.00\NN
F(21)&0.98&0.96&0.98&0.97&1.00&0.89&0.99\NN
E(1)&0.46&0.40&0.19&-&0.83&0.05&0.82\NN
E(7)&0.26&0.08&0.24&-&0.73&0.07&0.83\NN
E(21)&0.33&0.05&0.20&-&0.64&0.05&0.86\LL
}

\section{Acknowledgements}
Authors would like to thank Paulina Krupska and Joanna Sobczyk for carrying out reference measurements using the hyperspectral imaging system and the spectrometer. In addition, the Laboratory of Analysis and Nondestructive Investigation of Heritage Objects (LANBOZ) at the National Museum in Krakow is acknowledged for allowing the authors to use their SPECIM hyperspectral cameras. Authors would also like to thank Alicja Majda for making available the spectral library described in~\cite{majda2018hyperspectral}.

\bibliographystyle{unsrt}
\bibliography{hyperblood}

\end{document}